\documentclass[twoside,11pt]{article}
\usepackage{jair, theapa, rawfonts}

\usepackage{multirow}
\usepackage[normalem]{ulem}
\useunder{\uline}{\ul}{}

\usepackage{url}
\usepackage{graphicx}

\usepackage{amsmath}
\usepackage{amssymb}
\usepackage{dsfont}

\usepackage{algorithmic}
\usepackage{algorithm}
\usepackage{amsthm}

\usepackage{amsfonts}       
\usepackage{nicefrac}       

\usepackage{wrapfig}
\usepackage{color}
\usepackage{colortbl}

\usepackage[font=small,labelfont=sf,skip=0pt]{caption}
\usepackage[caption=true,font=small,labelfont=sf,textfont=sf,skip=0pt]{subfig}

\newtheorem{prop}{Proposition}[section]
\usepackage{tcolorbox}

\DeclareMathOperator*{\argmin}{arg\,min}
\DeclareMathOperator*{\argmax}{arg\,max}

\jairheading{80}{2024}{127-172}{03/2023}{05/2024}

\ShortHeadings{Effectiveness of Tree-based Ensembles for Anomaly Discovery}{Das, Islam, Jayakodi, \& Doppa}
\firstpageno{1}

\begin{document}

\title{{Effectiveness of Tree-based Ensembles for Anomaly Discovery: Insights, Batch and Streaming Active Learning}
}

\author{\name Shubhomoy Das \email shubhomoy.das@wsu.edu \\
       \name Md Rakibul Islam \email mdrakibul.islam@wsu.edu \\
       \name Nitthilan Kannappan Jayakodi \email n.kannappanjayakodi@wsu.edu \\
       \name Janardhan Rao Doppa \email jana.doppa@wsu.edu \\
			\addr School of EECS, Washington State University, Pullman, WA 99164, USA}

\maketitle

\begin{abstract}
Anomaly detection (AD) task corresponds to identifying the true anomalies among a given set of data instances. AD algorithms score the data instances and produce a ranked list of candidate anomalies. The ranked list of anomalies is then analyzed by a human to discover the true anomalies. Ensemble of tree-based anomaly detectors trained in an unsupervised manner and scoring based on uniform weights for ensembles are shown to work well in practice. However, the manual process of analysis can be laborious for the human analyst when the number of false-positives is very high. Therefore, in many real-world AD applications including computer security and fraud prevention, the anomaly detector must be configurable by the human analyst to minimize the effort on false positives. One important way to configure the detector is by providing true labels (nominal or anomaly) for a few instances. Recent work on active anomaly discovery has shown that greedily querying the top-scoring instance and tuning the weights of ensemble detectors based on label feedback allows us to quickly discover true anomalies.

This paper makes four main contributions to improve the state-of-the-art in anomaly discovery using tree-based ensembles. First, we provide an important insight that explains the practical successes of unsupervised tree-based ensembles and active learning based on greedy query selection strategy. We also present empirical results on real-world data to support our insights and theoretical analysis to support active learning. Second, we develop a novel batch active learning algorithm to improve the diversity of discovered anomalies based on a formalism called compact description to describe the discovered anomalies. Third, we develop a novel active learning algorithm to handle streaming data setting. We present a data drift detection algorithm that not only detects the drift robustly, but also allows us to take corrective actions to adapt the anomaly detector in a principled manner. Fourth, we present extensive experiments to evaluate our insights and our tree-based active anomaly discovery algorithms in both batch and streaming data settings. Our results show that active learning allows us to discover significantly more anomalies than state-of-the-art unsupervised baselines, our batch active learning algorithm discovers diverse anomalies, and our algorithms under the streaming-data setup are competitive with the batch setup.

\end{abstract}

\vspace{-1.0ex}
\section{Introduction}
\label{sec:introduction}

We consider the problem of {\em anomaly detection} (AD), where the goal is to detect unusual but interesting data (referred to as {\em anomalies}) among the regular data (referred to as {\em nominals}). This problem has many real-world applications including credit card transactions, medical diagnostics, computer security, etc., where anomalies point to the presence of phenomena such as fraud, disease, malpractices, or novelty which could have a large impact on the domain, making their timely detection crucial. Anomaly detection poses severe challenges that are not seen in traditional learning problems. First, anomalies are significantly fewer in number than nominal instances. Second, unlike classification problems, there may not be a hard decision boundary to separate anomalies and nominals. 

Typically, anomaly detection algorithms train models to compute scores for all instances, and report instances which receive the highest scores as anomalies. The candidate anomalies in the form of top-ranked data instances are analyzed by a human analyst to identify the true anomalies. Prior work has shown that a class of ensemble anomaly detectors trained in an unsupervised manner and scoring based on uniform weights work very well \cite{das:2016}. Algorithms based on tree-based ensembles such as isolation forest (IFOR) consistently perform well on real-world data \cite{emmott:2015}. However, purely unsupervised algorithms are usually ineffective in the following scenarios: a) The underlying model assumptions are not suited to the real-world task \cite{chandola:09}; b) The data volume is so high that even a detector with low false-positive rate reports a very large absolute number of potential anomalies; c) The nature/type of anomalies is not known beforehand (unknown unknowns) and the user wants to explore the data by directing the detector with feedback in the spirit of collaborative problem solving; and d) the same algorithm is able to identify multiple types of anomalies but only a subset of them are relevant for a particular domain. Consider the following simple example. Let us assume that there are three features (say A, B, and C) and we use a generic unsupervised detector which reports instances which are outside 2-standard deviation along each axis as anomalies. However, if this rule is relevant only for features A and B for a particular domain, then the purely unsupervised algorithm would be less useful for this domain as it would report too many false positives. As a concrete real-world scenario, assume that in a computer network security application, the features A, B, and C correspond to \emph{total bytes transferred over the network}, \emph{no. of unique IP addresses connected to}, and \emph{no. of logins in off-business hours} respectively for a server. With work from home mandates, the last feature might become irrelevant. Unless the detector is adjusted to account for this, the number of false positives could be high.

If the candidate set of anomalies contains many false-positives, it will significantly increase the effort of the human analyst in discovering true anomalies. To overcome the above drawbacks of unsupervised AD algorithms, recent work has explored active learning framework to configure the anomaly detector by the human analyst to minimize the overall effort on false-positives \cite{das:2016,siddiqui:2018}. In this framework, a human analyst provides label feedback (nominal or anomaly) for one or more data instances during each round of interaction. This feedback is used by the anomaly detector to change the scoring mechanism of data instances towards the goal of increasing the true anomalies appearing at the top of the ranked list. Specifically, prior work employs a greedy query selection strategy by selecting the top-scoring unlabeled instance for feedback. Similar to unsupervised anomaly detection ensembles, prior work showed that active learning using tree-based ensembles performs better \cite{das:2016,siddiqui:2018}. 

The effectiveness of tree-based ensembles for anomaly discovery in both unsupervised and active-learning settings raise two fundamental questions: {\bf (Q1)} Why does the average score across ensemble members perform best in most cases \cite{Ensemble:KDD2015} instead of other score combination strategies (e.g., min, max, median etc.)?; and {\bf(Q2)} Why does the greedy query selection strategy for active learning almost always perform best? In this paper, we investigate these two fundamental questions related to tree-based ensembles and provide an important insight that explains their effectiveness for anomaly discovery. We also provide empirical evidence on real-world data to support our intuitive understanding and theoretical analysis to support active anomaly discovery. Prior work on active learning for anomaly discovery has some important shortcomings. First, algorithmic work on enhancing the diversity of discovered anomalies is lacking \cite{gornitz:2013}. Second, most algorithms are designed to handle batch data well, but there are few principled algorithms to handle streaming data setting that arises in many real-world applications. To fill this important gap in active anomaly discovery, we exploit the inherent strengths of tree-based ensembles and propose novel batch and streaming active learning algorithms.

\vspace{1.0ex}

\noindent {\bf Contributions.} The main contribution of this paper is to study and evaluate anomaly discovery algorithms based on tree-based ensembles in both unsupervised and active learning settings. Specific contributions include:
\begin{itemize}
	\item We present an important insight into how tree-based anomaly detector ensembles are naturally suited for active learning, and why the greedy querying strategy of seeking labels for instances with the highest anomaly scores is efficient. We also provide theoretical analysis to support active anomaly discovery using tree-based ensembles.
	\item A novel formalism called compact description (CD) is developed to describe the discovered anomalies using tree-based ensembles. We develop batch active learning algorithms based on CD to improve the diversity of discovered anomalies.
	
	\item To handle streaming data setting, we develop a novel algorithm to robustly detect drift in data streams and design associated algorithms to adapt the anomaly detector on-the-fly in a principled manner.	
	\item We present extensive empirical evidence in support of our insights and algorithms on several benchmark datasets. The results show the efficacy of the proposed active learning algorithms in both batch and streaming data settings. 
	Our results show that in addition to discovering significantly more anomalies than state-of-the-art unsupervised baselines, our learning algorithm under the streaming-data setup is competitive with the batch setup. Our code and data are publicly available\footnote{\url{https://github.com/shubhomoydas/ad_examples}}.
\end{itemize}

\vspace{1.0ex}

\noindent {\bf Outline of the Paper.} The remainder of the paper is organized as follows. In Section~\ref{sec:related-work}, we discuss the prior work related to this paper. We introduce our problem setup and give a high-level overview of the generic human-in-the-loop learning framework for active anomaly discovery in Section~\ref{sec:problem-setup}. In Section~\ref{sec:insights}, we describe the necessary background on tree-based anomaly detector ensembles and provide intuition along with empirical evidence and theoretical analysis to explain their suitability for active anomaly discovery. In Section 5 and 6, we present a series of algorithms to support the framework for batch data and streaming data settings respectively. Section~\ref{sec:experiments} presents our experimental results and finally Section~\ref{sec:conclusions} provides summary and directions for future work.

\section{Related Work}
\label{sec:related-work}

\noindent \textbf{Unsupervised anomaly detection algorithms} are trained without labeled data and have assumptions baked into the model about what defines an anomaly or a nominal \cite{breunig:00,liu:08,pevny:2015,emmott:2015}. They typically cannot change their behavior to correct for the false positives after they have been deployed. Ensembles of unsupervised anomaly detectors \cite{aggarwal:2017} try to guard against the bias induced by a single detector by incorporating decisions from multiple detectors. The potential advantage of ensembles is that when the data is seen from multiple views by more than one detector, the set of anomalies reported by their joint consensus is more reliable and has fewer false positives. Different methods of creating ensembles include the collection of heterogeneous detectors \cite{senator:2013}, feature bagging \cite{lazarevic:05}, varying the parameters of existing detectors such as the number of clusters in a Gaussian Mixture Model \cite{emmott:2015}, sub-sampling, bootstrap aggregation etc. Some unsupervised detectors such as LODA \cite{pevny:2015} are not specifically designed as ensembles but may be treated as such because of their internal structure. Isolation forest (IFOR) \cite{liu:08} is a state-of-the-art tree-based ensemble anomaly detector. Recently, some prior work looked at the problem of ensemble selection by transfer learning \cite{Campos2018AnUB}, ensemble score aggregation methods \cite{dcsozhao2019} by looking at the locality of a data point, approximated neighborhood creation techniques and combine their scores \cite{kirner2017good} as outlier scores, feature selection process in outlier detection \cite{cheng2020outlier} \cite{micenkova2014learning}.
Zero++ \cite{Pang2016ZEROHT} is an unsupervised approach that works only for categorical data. It is similar to IFOR, but there is no statistically significant difference in detection accuracy between Zero++ and IFOR for numerical data that is transformed to categorical data. 

\vspace{1.0ex}

\noindent \textbf{Active learning} corresponds to the setup where the learning algorithm can selectively query a human analyst for labels of input instances to improve its prediction accuracy. The overall goal is to minimize the number of queries to reach the target performance. There is a significant body of work on both theoretical analysis \cite{freund:1997,balcan:2007,balcan:2015,monteleoni:2006,dasgupta:2009,yan:2017,cao2022,cao2023} and applications \cite{settles:2010} of active learning.

\vspace{1.0ex}

\noindent \textbf{Active learning for anomaly detection} has been explored by researchers in several earlier works \cite{almgren:2004,abe:2006,stokes:2008,he:2008,Pichara2011ActiveLA,gornitz:2013}. In this setting, the human analyst provides feedback to the algorithm on true labels (nominal or anomaly). If the algorithm makes wrong predictions, it updates its model parameters to be consistent with the analyst's feedback. Interest in this area has gained prominence in recent years due to significant rise in the volume of data in real-world settings, which made reduction in false-positives much more critical. Most of the popular unsupervised anomaly detection algorithms have been adapted to the active learning setting. Some of these methods are based on ensembles and support streaming data \cite{veeramachaneni:2016,stokes:2008} but maintain separate models for anomalies and nominals internally, and do not exploit the inherent strength of the ensemble in a systematic manner. \cite{nissim:14} incorporates user feedback into SVM, a classification algorithm. \cite{das:2016} and \cite{das:2017}, based on LODA and IFOR respectively, re-weight the components of the ensemble with every user feedback in a manner similar to our work, however, they do not support streaming data and discovery of diverse anomalies. \cite{siddiqui:2018} updates the same model as \cite{das:2017} in an online incremental manner when incorporating user feedback. 

\cite{Ikeda2018HumanAssistedOA} employs an autoencoder as the base anomaly detector and adds a penalty term for the false positives (i.e., normal data that are incorrectly reported as anomalies). The user examines the anomalies reported by the algorithm and labels the false-positives. The model is then retrained with the new labeled data in an online manner. \cite{Pimentel_2020} adds a classifier layer on top of any unsupervised deep network based anomaly detector and updates the parameters of the deep network based on human feedback. \cite{trittenbach2019one} extends SVDD, a one-class classifier model for outlier detection, to multiple sub-spaces. The multiple sub-spaces are intended to help with interpretability of the outliers and support better query strategies for active learning. \cite{kim2023} learns an adaptive boundary using a deep SVDD model by searching for locations having a similar number of normal and abnormal samples. This boundary is iteratively adjusted by getting labeling feedback from users with an uncertainty based query strategy. Deep semi-supervised anomaly detection techniques have also been used to incorporate feedback from users \cite{Ruff2018DeepOC,Ruff2020DeepSA,li2023-soel}. The performance gain for such techniques on classical tabular datasets can be quite low in comparison to performance on complex raw data such as images. SOEL \cite{li2023-soel} employs a loss function that has opposing effects on normal and anomalous data points. Using this loss, SOEL infers latent labels \{\textit{anomaly}, \textit{normal}\} for each training example using a semi-supervised EM-style algorithm. In this work, we present a comparison of our algorithm with SOEL in Section \ref{subsec:limited-label-baseline-comparison}.

Our proposed algorithmic instantiation of HiLAD for batch data setting (HiLAD-Batch) and recent work on feedback-guided AD via online optimization (\texttt{Feedback-guided Online}) \cite{siddiqui:2018} both build upon the same tree-based model proposed by \cite{das:2017}. Therefore, there is no fundamental difference in their performance (\ref{subsec:semisupervised-baseline-comparison}). The uniform initialization of weights employed by both HiLAD-Batch and \texttt{Feedback-guided Online} plays a critical role in their overall effectiveness. \texttt{Feedback-guided Online} also adopts the greedy query selection strategy mainly because it works well in practice. In this work, we present the fundamental reason why greedy query selection strategy is label-efficient for human-in-the-loop learning with tree-based anomaly detection ensembles.

Anomaly detection for streaming data setting has many real-life applications. Under this setup, data is sent continuously as a stream to the anomaly detector. \cite{Gupta2014OutlierDF} present a broad overview of this setup and categorize the outlier detection techniques into three sub-categories: (a) evolving prediction models, (b) distance based outlier detection for sliding windows, (c) methods for high-dimensional data streams. Our proposed approach falls under the class of distance based outlier detection in a sliding window. Existing approaches can detect either local anomalies \cite{breunig:00,Salehi2016FastME,Na2018DILOFEA} or global anomalies \cite{Angiulli2007DetectingDO,Yang2009NeighborbasedPD,Kontaki2011ContinuousMO}. \cite{guha:2016} proposed a tree-based streaming anomaly detector. The key idea is to create a sketch (or summary) of the data using a random cut forest and to report a data point as anomaly when the displacement score is high. However, unlike our work, none of these prior streaming anomaly detection methods incorporate human feedback. 

\section{Problem Setup and Human-in-the-Loop Learning Framework}
\label{sec:problem-setup}

We are given a dataset ${\bf D} = \{{\bf x}_1, ..., {\bf x}_n\}$, where ${\bf x}_i \in \mathds{R}^d$ is a data instance that is associated with a hidden label $y_i \in \{-1, +1\}$. Instances labeled $+1$ represent the \textit{anomaly} class and are at most a small fraction $\tau$ of all instances. The label $-1$ represents the \textit{nominal} class. We also assume the availability of an ensemble $\mathcal{E}$ of $m$ anomaly detectors which assigns scores ${\bf z} = \{z^1, ..., z^m\}$ to each instance ${\bf x}$ such that instances labeled $+1$ tend to have scores higher than instances labeled $-1$. Our framework is applicable for any ensemble of detectors, but we specifically study the instantiation for tree-based ensembles because of their beneficial properties. We denote the ensemble score matrix for all unlabeled instances by ${\mathbf H}$. The score matrix for the set of instances labeled $+1$ is denoted by ${\mathbf H}_+$, and the matrix for those labeled $-1$ is denoted by ${\mathbf H}_-$. 

We setup a scoring function \texttt{Score}(${\bf x}$) to score data instances, which is parameterized by ${\mathbf w}$ and ensemble of detectors $\mathcal{E}$. For example, with tree-based ensembles, we consider a linear model with weights ${\mathbf w} \in \mathds{R}^m$ that will be used to combine the scores of $m$ anomaly detectors as follows: \texttt{Score}(${\bf x}$) = ${\mathbf w} \cdot {\bf z}$, where ${\bf z} \in \mathds{R}^m$ corresponds to the scores from anomaly detectors for instance ${\bf x}$. This linear hyperplane separates anomalies from nominals. We will denote the optimal weight vector by ${\bf w}^*$. 

\begin{table}[]
\centering
\caption{Notational convention}
\label{tab:list-of-symbols}

\begin{tabular}{cl}
\hline
Symbol & \textbf{Description} \\ \hline
D & Dataset \\
$x_i$ & A data instance \\
$y_i$ & A label (hidden) \\
$\mathcal{E}$ & Ensemble \\
$z_m$ & Anomaly score from $m^{th}$ ensemble member \\
${\mathbf H}$ & Score matrix of all unlabeled data instances \\
${\mathbf H}_+$ & Score matrix of all positively labeled data instances \\
${\mathbf H}_-$ & Score matrix of all negatively labeled data instances \\
${\mathbf w}$ & Weight vector \\
B & Query budget \\
$\mathcal{L}$ & Labeled dataset \\
$\tau$ & Initial guess on the fraction of anomalous data instances \\
$q$ & Score for $\tau^{th}$ ranked anomaly \\ \hline
\end{tabular}
\end{table}

\begin{figure}[h]
	\centering
	\includegraphics[width=6in]{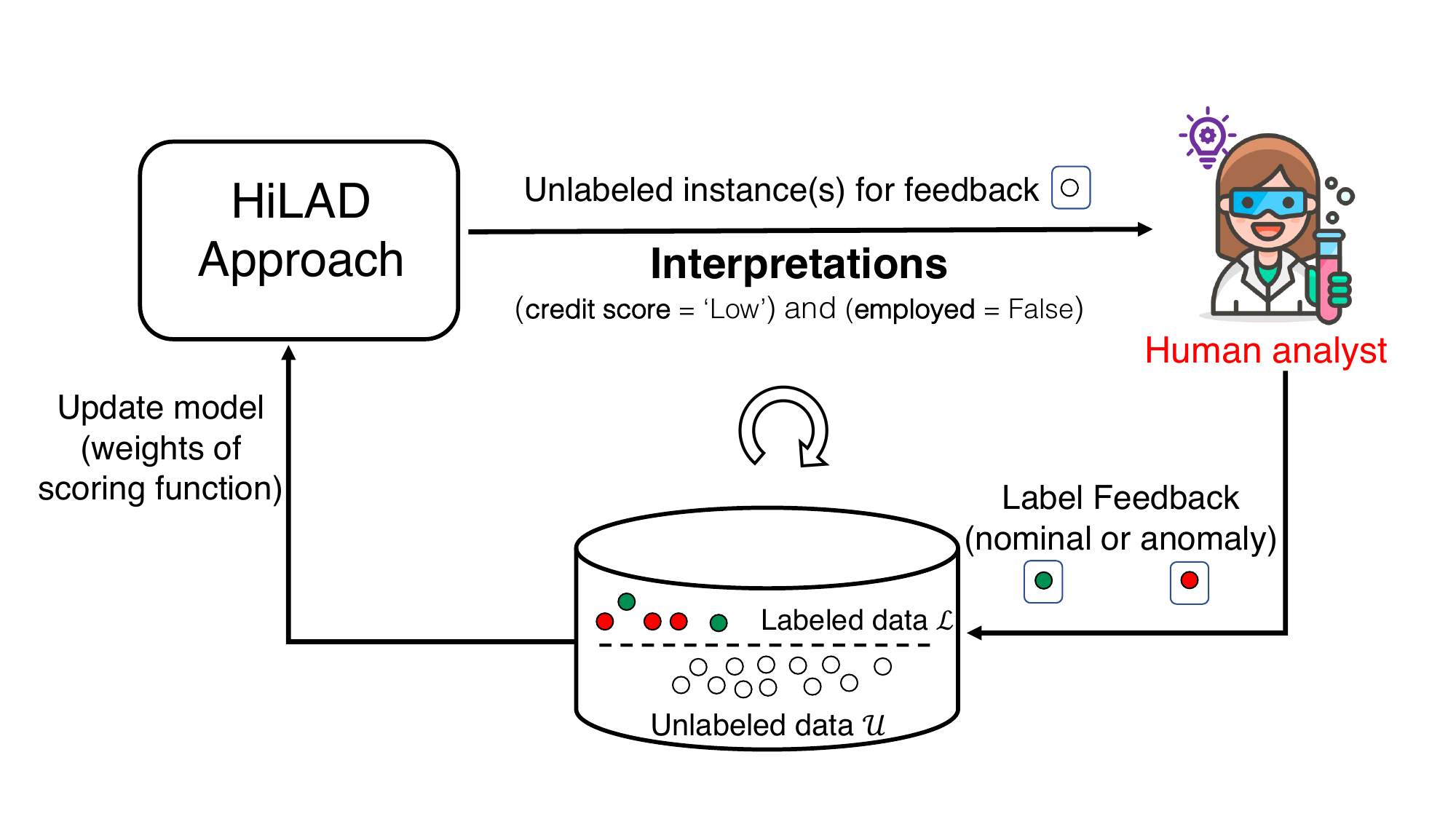}
	\caption{High-level overview of the human-in-the-loop learning framework for anomaly detection. Our goal is to maximize the number of true anomalies presented to the analyst.
}
	\label{fig:active_loop}
\end{figure}

The generic human-in-the-loop learning framework HiLAD assumes the availability of an analyst who can provide the true label for any instance in an interactive loop as shown in Figure~\ref{fig:active_loop}. In each iteration of the learning loop, we perform the following steps: 1) Select one or more unlabeled instances from the input dataset ${\bf D}$ according to a query selection strategy; 2) Query the human analyst for labels of selected instances; and 3) Update the weights of the scoring function based on the aggregate set of labeled and unlabeled instances. The overall goal of the framework is to learn optimal weights for maximizing the number of true anomalies shown to the human analyst.

\begin{algorithm}
	\caption{\texttt{HiLAD Framework}}
	\label{alg:hil-framework}
	\begin{algorithmic}
		\STATE \textbf{Input:} Unlabeled dataset ${\mathcal U}$, Feedback budget $B$, and Anomaly detection ensembles $\mathcal{E}$ of size $m$
		\STATE Initialize weights ${\mathbf w} \in \mathds{R}^m$ of the linear scoring function \texttt{Score}(${\bf x}$)
		\STATE Initialize ${\mathcal L} \leftarrow \emptyset$ // Labeled dataset
		\WHILE{$|{\mathcal L}| < B$}
		\STATE ${\mathbf X}_{query} \leftarrow $ one or more instances selected from ${\mathcal U}$ using query selection strategy
		\STATE Get label feedback for instances in ${\mathbf X}_{query}$ from the human analyst by showing interpretations of anomalous instances
		\STATE Add newly labeled examples to ${\mathcal L}$
		\STATE Update weights ${\mathbf w}$ to be consistent with aggregate labeled data ${\mathcal L}$
		\STATE Update unlabeled set: ${\mathcal U} \leftarrow {\mathcal U} \setminus {\mathbf X}_{query}$
		\ENDWHILE
		\RETURN discovered anomalous instances from ${\mathcal L}$
	\end{algorithmic}
\end{algorithm}

\vspace{2.0ex}

\noindent We provide algorithmic solutions for key elements of this human-in-the-loop framework: 

\begin{itemize}

\item Initializing the parameters ${\bf w}$ of the scoring function \texttt{Score}(${\bf x}$) based on a key insight for tree-based anomaly detection ensembles.

\item Query selection strategies to improve the label-efficiency of learning.

\item Updating the weights of scoring function based on label feedback. 

\item Updating ensemble members as needed to support the streaming data setting.

\end{itemize}

\vspace{-2.0ex}

\section{Suitability of Tree-based Ensembles for Human-in-the-Loop Learning}
\label{sec:insights}

In this section, we first provide the background on tree-based anomaly detector ensembles and describe {\em Isolation Forest (IFOR)} in more depth as we employ it for our experimental evaluation. Subsequently, we provide intuition on how tree-based ensembles are naturally suited for human-in-the-loop learning and show empirical evidence using isolation forest.

\subsection{Tree-based Anomaly Detection Ensembles}
\label{sec:ensembles}

Ensemble of tree-based anomaly detectors have several attractive properties that make them an ideal candidate for human-in-the-loop learning:
\begin{itemize}
	\setlength\itemsep{0em}
	\item They can be employed to construct large ensembles inexpensively.
	\item Treating the nodes of tree as ensemble members allows us to both focus our feedback on fine-grained subspaces\footnote{In this article, subspace stands for location encompassed by a leaf node of the tree, i.e., it is a rectangle for 2D space. All the leaf nodes for a tree cover the complete data space. This not to be confused with subspaces from Linear Algebra. } as well as increase the \emph{capacity} of the model in terms of separating anomalies (positives) from nominals (negatives) using complex class boundaries.
	\item Since some of the tree-based models such as \textit{Isolation Forest} (IFOR) \cite{liu:08}, \textit{HS Trees} (HST) \cite{tan:2011}, and \textit{RS Forest} (RSF) \cite{wu:2014} are state-of-the-art unsupervised detectors \cite{emmott:2015,domingues:2018}, it is a significant gain if their performance can be improved with minimal label feedback.
\end{itemize}

In this work, we will focus mainly on IFOR because it performed best across all datasets \cite{emmott:2015}. However, we also present results on HST and RSF wherever applicable.

\begin{figure}[h]
	\centering
	\includegraphics[width=4.0in]{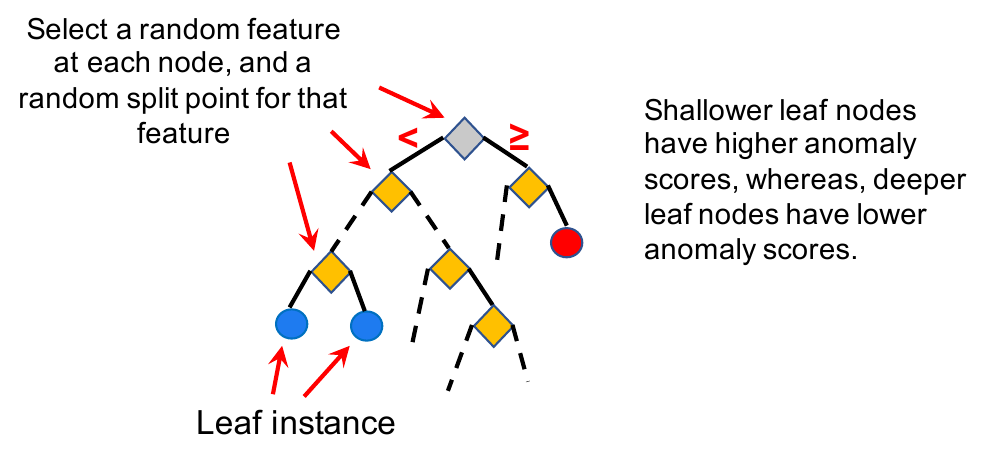}
	\caption{Illustration of Isolation tree from \cite{das:2017}.}
	\label{fig:isolation_tree}
\end{figure}

\begin{figure}[t]
	\centering
	\subfloat[Toy dataset]{\includegraphics[width=.3\linewidth]{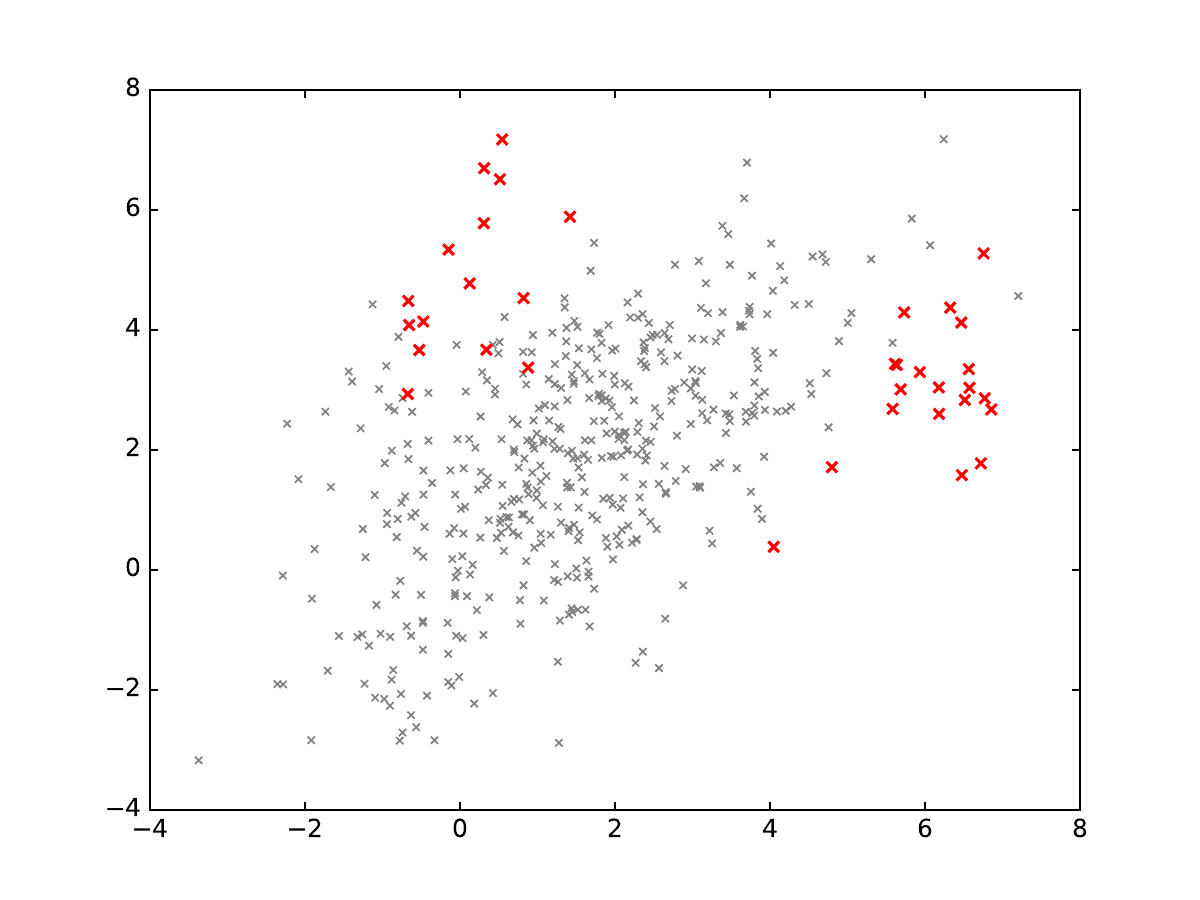} \label{fig:synthetic_dataset}}
	\subfloat[One isolation tree]{\includegraphics[width=.3\linewidth]{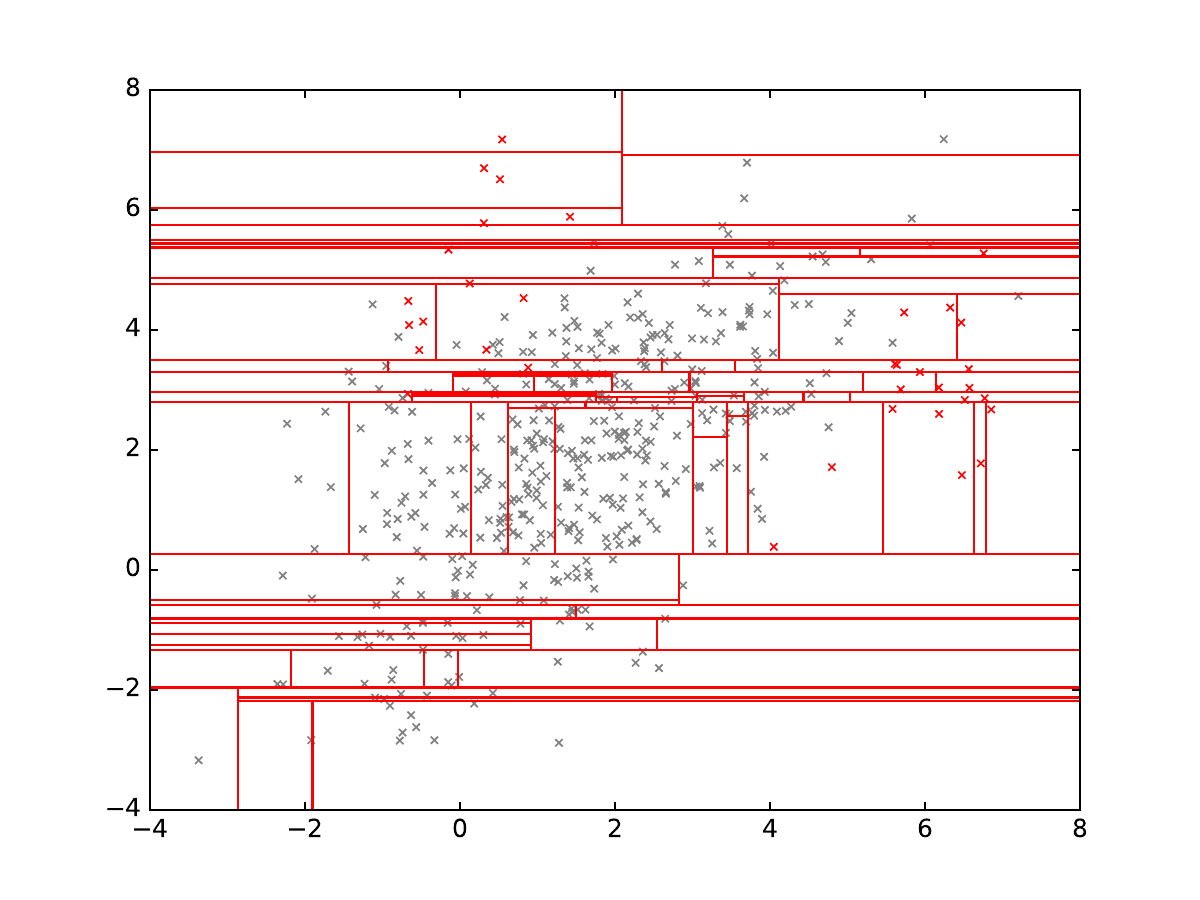} \label{fig:one_tree}}
	\subfloat[Scores from one tree]{\includegraphics[width=.3\linewidth]{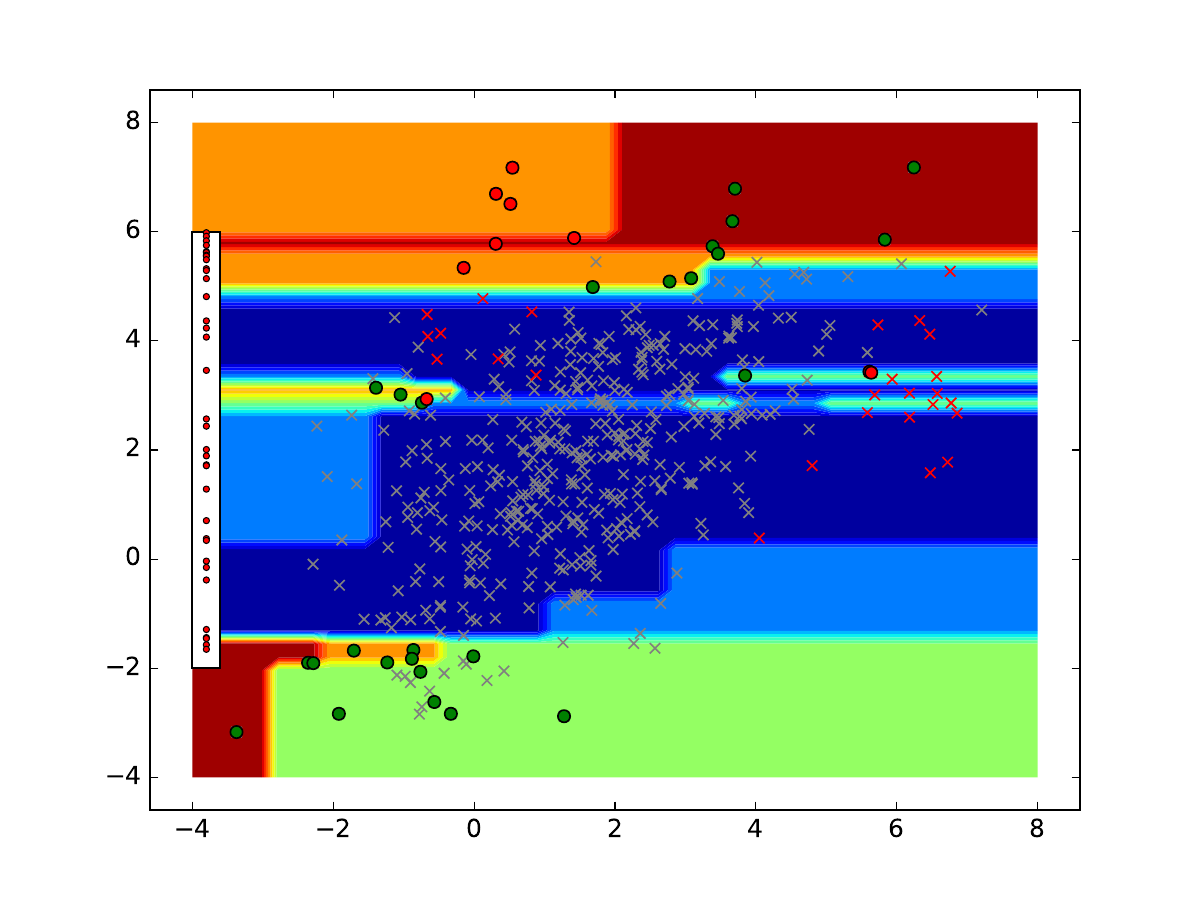} \label{fig:contours_one_tree}}
	\caption{Illustration of Isolation Tree on simple data. {\bf (a)} Toy dataset \cite{das:2017}. {\bf (b)} A single isolation tree for the Toy dataset. {\bf (c)} Regions having deeper red belong to leaf nodes which have shorter path lengths from the root and correspondingly, higher anomaly scores. Regions having deeper blue correspond to longer path lengths and lower anomaly scores.}
	\label{fig:isolation_tree_illustration}
\end{figure}

\begin{figure}[h]
	\centering
	\subfloat[IFOR]{\includegraphics[width=0.33\textwidth]{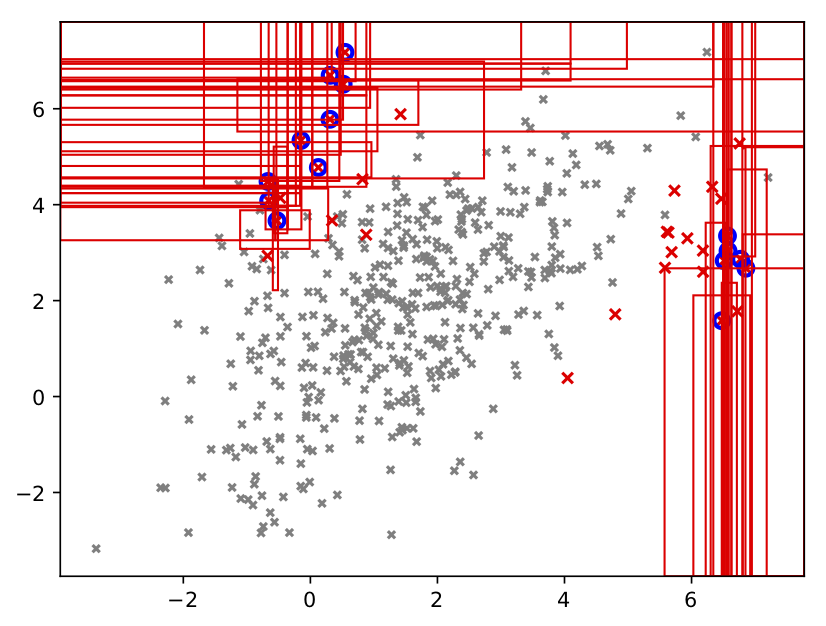}%
		\label{fig:iforest_regions}}
	\subfloat[HST (depth=$15$)]{\includegraphics[width=0.33\textwidth]{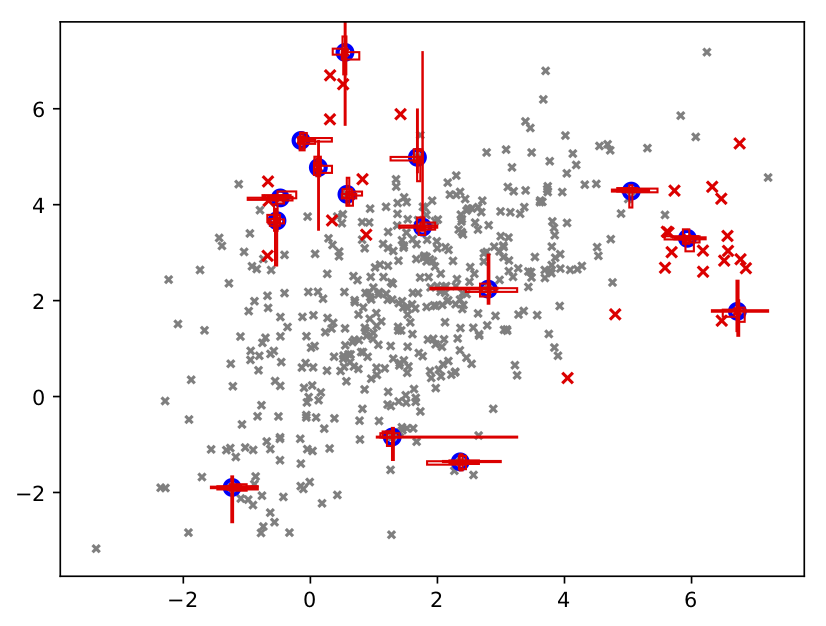}%
		\label{fig:hstrees_regions_15}}
	\subfloat[HST (depth=$8$)]{\includegraphics[width=0.33\textwidth]{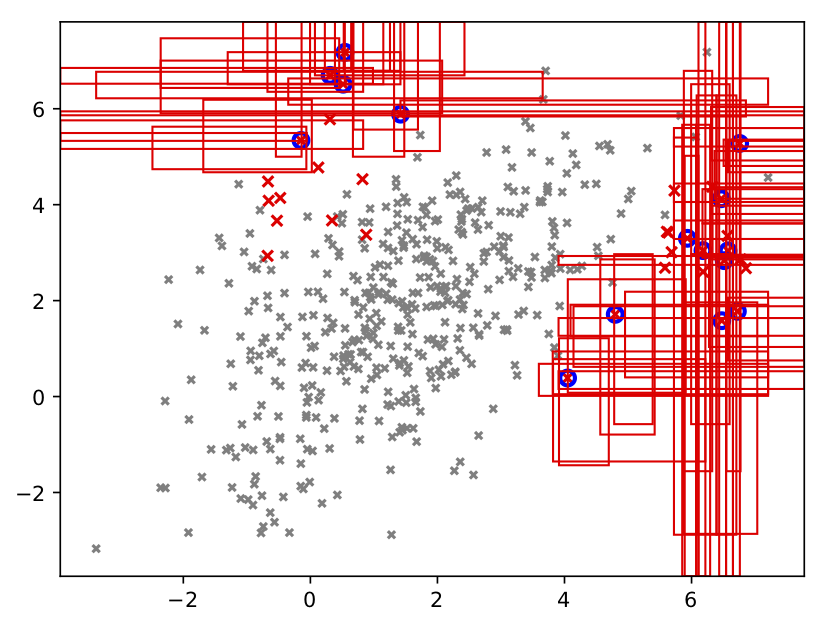}%
		\label{fig:hstrees_regions_8}}
	\caption{Illustration of differences among different tree-based ensembles. The \textcolor{red}{red} rectangles show the union of the $5$ most anomalous subspaces across each of the $15$ most anomalous instances (\textcolor{blue}{blue}). These subspaces have the highest influence in propagating feedback across instances through gradient-based learning under our model. HST has fixed depth which needs to be high for accuracy (recommended $15$ \cite{tan:2011}). IFOR has adaptive height and most anomalous subspaces are shallow. Higher depths are associated with smaller subspaces which are shared by fewer instances. As a result, feedback on any individual instance gets passed on to many other instances in IFOR, but to fewer instances in HST. RSF has similar behavior as HST. We set the depth for HST (and RSF \cite{wu:2014}) to $8$ (Figure~\ref{fig:hstrees_regions_8}) in our experiments in order to balance accuracy and feedback efficiency.}
	\label{fig:tree_differences}
\end{figure}

\noindent \textbf{Isolation Forest (IFOR)} comprises of an ensemble of \textit{isolation} trees. Each tree partitions the original feature space at random by recursively splitting an unlabeled dataset. At every tree-node, first a feature is selected at random, and then a split point for that feature is sampled uniformly at random (Figure~\ref{fig:isolation_tree}). This partitioning operation is carried out until every instance reaches its own leaf node. The key idea is that anomalous instances, which are generally isolated in the feature space, reach the leaf nodes faster by this partitioning strategy than nominal instances which belong to denser regions (Figure~\ref{fig:isolation_tree_illustration}). Hence, the path from the root node is shorter to the leaves of anomalous instances when compared to the leaves of nominal instances. This path length is assigned as the unnormalized score for an instance by an isolation tree. After training an IFOR with $T$ trees, we extract the leaf nodes as the members of the ensemble. Such members could number in the thousands (typically $4000-7000$ when $T=100$). Assume that a leaf is at depth $l$ from the root. If an instance belongs to the partition defined by the leaf, it gets assigned a score $-l$ by the leaf, else $0$. As a result, anomalous instances receive higher scores on average than nominal instances. Since every instance belongs to only a few leaf nodes (equal to $T$), the score vectors are \textit{sparse} resulting in low memory and computational costs.

\textbf{HST} and \textbf{RSF} apply different node splitting criteria than IFOR, and compute the anomaly scores on the basis of the sample counts and densities at the nodes. We apply log-transform to the leaf-level scores so that their unsupervised performance remains similar to the original model and yet improves with feedback. The trees in HST and RSF have a fixed depth which needs to be larger in order to improve the accuracy. In contrast, trees in IFOR have adaptive depth and most anomalous subspaces are shallow. Larger depths are associated with smaller subspaces, which are shared by fewer instances. As a result, feedback on any individual instance gets passed on to very few instances in HST and RSF, but to much more number of instances in IFOR. Therefore, it is more efficient to incorporate feedback in IFOR than it is in HST or RSF (see Figure~\ref{fig:tree_differences}).

\subsection{Beneficial Properties of Tree-based Anomaly Detector Ensembles}

\label{subsec:properties--for-treebased-ad}
In this section, we provide intuition for the effectiveness of unsupervised tree-based anomaly detector ensembles and list their beneficial properties for human-in-the-loop learning framework. Our arguments are motivated by the active learning theory for standard classification.

\begin{figure}[h]
	\centering
	\subfloat[\textbf{C1}: Common case]{\includegraphics[width=1.3in,height=1.3in]{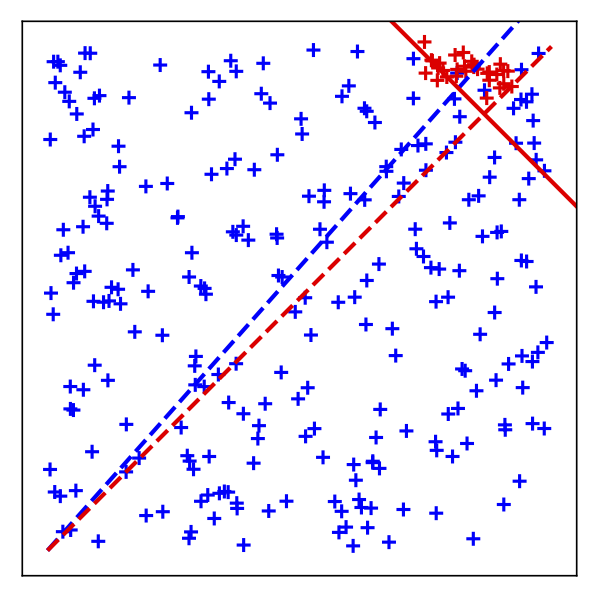} \label{fig:idealized_rect}}
	\subfloat[\textbf{C2}: Similar to Active Learning theory]{\includegraphics[width=2.7in,height=1.3in]{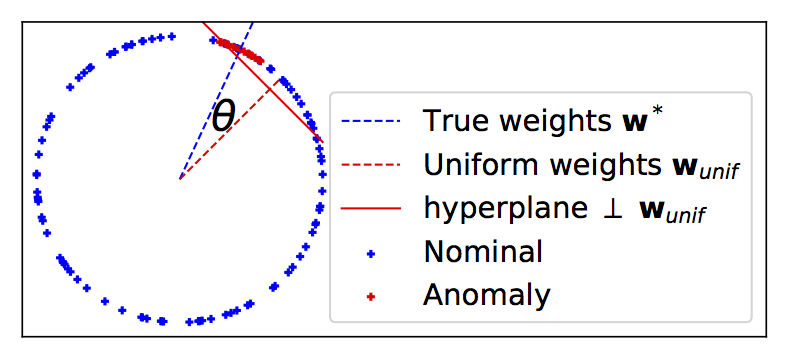} \label{fig:idealized_circ}}
	\subfloat[\textbf{C3}: IFOR case]{\includegraphics[width=1.3in,height=1.3in]{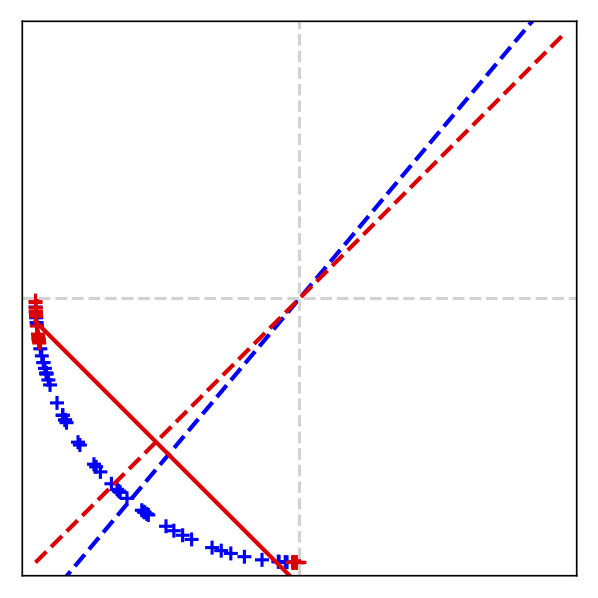} \label{fig:idealized_ifor}}
	\caption{Illustration of candidate score distributions from an ensemble in 2D. The two axes represent two different ensemble members. (a) \textbf{C1} represents the common case where both anomaly detectors want to score anomalous data points higher, (b) \textbf{C2} illustrates how active learning helps the model to learn the slight angle deviation, $\theta$, (c) \textbf{C3} is specifically for IFOR case where the anomalous data points has smaller path length. So, anomalous data points will be located at two extremes where the path lengths are smallest and can be separated by the non-homogeneous hyperplane.}
	\label{fig:idealized}
\end{figure}

Without loss of generality, we assume that all scores from the members of tree-based ensemble of anomaly detectors are normalized (i.e., they lie in $[-1, 1]$ or $[0, 1]$), with higher scores implying more anomalous. For the following discussion, ${\mathbf w}_{unif} \in \mathds{R}^m$ represents a vector of equal values, and $||{\mathbf w}_{unif}|| = 1$. Figure~\ref{fig:idealized} illustrates a few possible distributions of normalized scores from the ensemble members in 2D. In each figure, there are two ensemble members and each member tends to assign higher scores to anomalies than to nominals such that, on \emph{average}, anomalies tend to have higher scores than nominals. Let ${\bf z}$ represent the vector of anomaly scores for an instance and let $m$ be the number of ensemble members. Then, the average score for the same instance can be computed as \texttt{Score}=$\frac{1}{m}\sum_{i=1}^{m}{z_i} = \frac{1}{\sqrt{m}}({\bf w}_{unif}\cdot{\bf z})\propto{\bf w}_{unif}\cdot{\bf z}$. We observe that under this setup, points in the score space closest to the top right region of the score space have the highest average scores since they are most aligned with the direction of ${\bf w}_{unif}$. When ensemble members are ``good'', they assign higher scores to anomalies and push them to this region of the score space. The most general case is illustrated in case {\bf C1} in Figure~\ref{fig:idealized_rect}. This behavior makes it easier to separate anomalies from nominals by a hyperplane. Moreover, it also sets a prior on the position and alignment of the separating hyperplane (as shown by the red solid line). Having priors on the hyperplane helps determine the region of label uncertainty (anomaly vs. nominal) even without any supervision, and this, in theory, strongly motivates active learning using the human analyst in the loop.

\begin{figure}[t]
	\centering
	\subfloat[Toy dataset]{\includegraphics[width=.33\linewidth]{figures/synth_dataset.pdf} \label{fig:synthetic_dataset}}
	\subfloat[Anomaly scores]{\includegraphics[width=.31\linewidth]{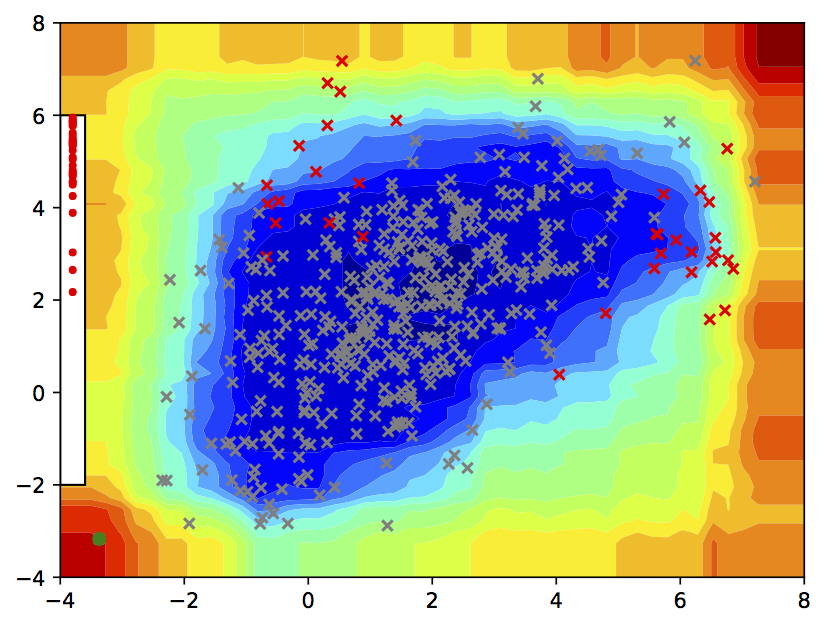} \label{fig:initial_scores}}
	\subfloat[Angles]{\includegraphics[width=.3\linewidth]{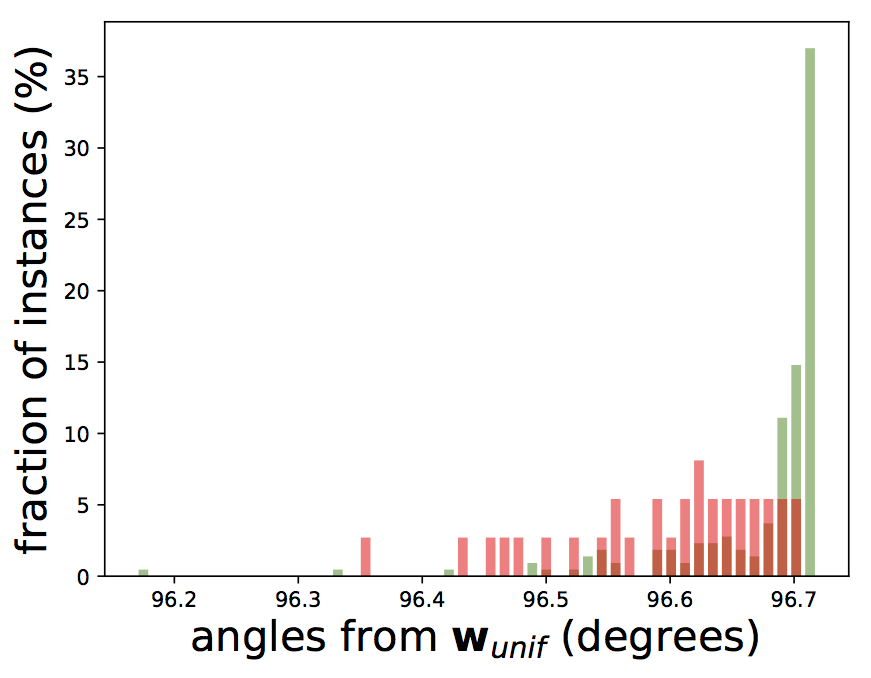} \label{fig:angles_toy2}}
	\caption{Isolation Tree. {\bf (a)} Toy dataset \cite{das:2017} which will be used as the running example through out the text to illustrate the ideas. Red points are anomalies and black points are nominals. {\bf (b)} Anomaly scores assigned by IFOR to the Toy dataset. {\bf (c)} Histogram distribution of the angles between score vectors from IFOR and ${\mathbf w}_{unif}$. The red and green histograms show the angle distributions for anomalies and nominals respectively. Since the red histograms are closer to the left, anomalies are aligned closer to ${\mathbf w}_{unif}$.}
	\label{fig:synthetic_dataset_angle}
\end{figure}

Most theoretical research on active learning for \emph{classification} \cite{kearns:1998,balcan:2007,kalai:2008,dasgupta:2009,balcan:2015} makes simplifying assumptions such as uniform data distribution over a unit sphere and with homogeneous (i.e., passing through the origin) hyperplanes. However, for \emph{anomaly detection}, arguably, the idealized setup is closer to case {\bf C2} (Figure~\ref{fig:idealized_circ}), where \emph{non-homogeneous} decision boundaries are more important. More importantly, in order for the theory to be relevant for anomaly detection, the assumptions behind it should be realizable with multi-dimensional real-world data and a competitive anomaly detector. We present empirical evidence (Section~\ref{sec:experiments}) which shows that scores from the state-of-the-art \textit{Isolation Forest} (IFOR) detector are distributed in a similar manner as case {\bf C3} (Figure~\ref{fig:idealized_ifor}). {\bf C3} and {\bf C2} are similar in theory (for active learning) because both involve searching for the optimum non-homogeneous decision boundary. In all cases, the common theme is that when the ensemble members are ideal, then \emph{the scores of true anomalies tend to lie in the farthest possible location in the positive direction of the uniform weight vector ${\bf w}_{unif}$ \textbf{by design}}. Consequently, the average score for an instance across all ensemble members works well for anomaly detection. However, not all ensemble members are ideal in practice, and the true weight vector (${\bf w}^*$) is displaced by an angle $\theta$ from ${\bf w}_{unif}$. Figure~\ref{fig:angles_toy2} shows an illustration of this scenario on a {\em Toy} dataset. In large datasets, even a small misalignment between ${\bf w}_{unif}$ and ${\bf w}^*$ results in many false positives. While the performance of ensemble on the basis of the AUC metric may be high, the detector could still be impractical for use by analysts.

The property, that the misalignment is usually small, can be leveraged by active learning to learn the optimal weights efficiently using a small amount of label feedback. To understand this, observe that the top-ranked instances are close to the decision boundary and are therefore, in the uncertainty region. The key idea is to design a hyperplane that passes through the uncertainty region which then allows us to select query instances by uncertainty sampling. Selecting instances on which the model is uncertain for labeling is efficient for active learning \cite{cohn:1994,balcan:2007,cao2023}. Specifically, greedily selecting instances with the highest scores is first of all more likely to reveal anomalies (i.e., true positives), and even if the selected instance is nominal (i.e., false positive), it still helps in learning the decision boundary efficiently. This is an important insight and has significant practical implications. We present the label complexity of this greedy (but efficient) active learning algorithm for anomaly detection below.

\noindent\textbf{Label complexity of active learning with fixed weights:} Our trained ensemble of two anomaly detectors assign scores uniformly distributed over a unit sphere (Fig~\ref{fig:idealized}). Since we have assumed that $\tau$ fraction of the data is anomalous, marginal density of anomalies is then $\frac{\tau}{2\pi}$. We also have a pool of unlabeled data. For simplicity, we assume that anomalies have a Gaussian distribution as a function of the angle $\omega \in [-\pi, \pi]$ from ${\mathbf u}$ on the unit sphere: $f(y=+1|\omega) \sim \mathcal{N}(0, \sigma^2)$. We query an analyst for the labels of only the top $\beta$ fraction of instances sorted on scores, i.e., $\{{\mathbf z} : \arccos({\mathbf w} \cdot {\mathbf z}) \leq \beta\pi\}$. Let the angle between ${\mathbf u}$ and ${\mathbf w}_{unif}$ be $\theta$. Let $\beta_{l}$ and $\beta_{r}$ be the minimum and maximum of the values $\{\theta-\beta\pi, \theta+\beta\pi\}$ respectively. As per our query selection strategy, the probability of labeling a true anomaly is then $p_{\theta}=\int_{\beta_l}^{\beta_r}f(y=+1|\omega)d\omega = (\Phi(\frac{\beta_{r}}{\sigma}) - \Phi(\frac{\beta_{l}}{\sigma}))$, where $\Phi(\cdot)$ is the c.d.f of the Gaussian distribution. If $|\theta|$ is large, i.e., ${\mathbf w}_{unif}$ and ${\mathbf u}$ are not ``close'', then $p_{\theta}$ will be small.
\begin{prop}
	\label{prop:fixed_wts}
	Let $\delta\in[0, 1]$. For the 2D case, the number of labels needed to learn the decision boundary with probability $(1-\delta)$ with pool-based active learning is $T = O(\log(\frac{1}{\sigma})\frac{1}{p_{\theta}}\log(\frac{1}{\delta}))$.
\end{prop}
\noindent\textit{Proof:} Essentially, we need to locate the mean of the Gaussian distribution. This can be done through binary search because the density is highest at the mean and decreases continuously as we move away from it. For the search, we estimate the density of anomalies at $O(\log(\frac{1}{\sigma}))$ locations (for accuracy to one s.d.). For any estimate, at least one anomaly must be sampled. When the angle is $\theta$ (i.e., ${\bf w}$=${\bf w}_{unif}$) at the initial location, then the number of samples required is, with probability $(1-\delta)$: $(1-p_{\theta})^{l} \leq \delta \Rightarrow l \geq \frac{1}{p_{\theta}}\log(\frac{1}{\delta})$. The number of samples at the initial location sets the baseline estimate; therefore, the total number of samples is $T=O(\log(\frac{1}{\sigma})\frac{1}{p_{\theta}}\log(\frac{1}{\delta}))$. \qedsymbol\\

\noindent {\bf Remark.} If we ignore the initial location of search, the number of labels required would be $O(\log(\frac{1}{\sigma})\frac{2\pi}{\tau}\log(\frac{1}{\delta}))$. Thus, initializing at ${\bf w}_{unif}$ helps when $p_{\theta}$ is larger than $\frac{\tau}{2\pi}$.\\

\noindent\textbf{Label complexity of active learning with varying weights:} In this case, we select one data instance from the top ranked $\beta$ fraction of instances at each time step for querying, and then adjust the weights ${\mathbf w}$ in response to the label received from the analyst. We denote the weight vector and its angular displacement from ${\mathbf u}$ at each time step $t$ by ${\mathbf w}_t$ and $\theta_t$ respectively, starting with ${\bf w}_1={\bf w}_{unif}$ and $\theta_1=\theta$.

\begin{prop}
	\label{prop:vary_wts}
	\textit{Let the algorithm $\mathcal{A}$ update the weights with each new label at time $t$ while learning the decision boundary such that $\theta_{t+1}\leq\theta_{t}$. If the label complexity of algorithm $\mathcal{A}$ is $T'$, then $T' \leq T$ (where T is defined in Proposition \ref{prop:fixed_wts}).}
\end{prop}

\noindent\textit{Proof:} As in the previous case, we sample at $O(\log(\frac{1}{\sigma}))$ locations for the binary search. Let $l'$ be the number of samples required at the initial location of the binary search for density estimation, and set $\theta_1=\theta$. Now, $l'$ sets the baseline for the number of samples in each round of the search. Since, by assumption, $\theta_{t+1}\leq\theta_{t}$ at each label iteration $t$, $p_{\theta_{t+1}} \geq p_{\theta_{t}}$. For any estimate, at least one anomaly needs to be selected with probability $(1-\delta)$; i.e., $(1-p_{\theta_1})..(1-p_{\theta_{l'}}) \leq \delta$. Since,  $(1-p_{\theta})^{l'} \geq (1-p_{\theta_1})..(1-p_{\theta_{l'}})$, $\exists l \geq l'$ s.t. $(1-p_{\theta})^{l} \leq \delta$. Therefore, $T' = O(\log(\frac{1}{\sigma})l') \leq O(\log(\frac{1}{\sigma})l) = T$. \qedsymbol

When the number of members in the ensemble is $m$, then $p_{\theta}$ will be a function of $m$ and the number of locations for binary search will be $O(m\log(\frac{1}{\sigma}))$. 

\vspace{2.0ex}

\noindent {\bf Summary.} Tree-based anomaly detector ensembles have several beneficial properties for human-in-the-loop learning for anomaly discovery:
\begin{itemize}
	\item Uniform prior over weights helps in improving the label-efficiency of discovering true anomalies. Indeed, in our experiments, we demonstrate the effectiveness of uniform weights over random weights for initialization in discovering anomalies. We also show the histogram distribution of the angles between score vectors from IFOR and uniform weights ${\mathbf w}_{unif}$ to demonstrate that anomalies are aligned closer to ${\mathbf w}_{unif}$.
	\item With uniform prior over weights, the greedy strategy of querying the labels for top ranked instances is efficient, and is therefore a good yardstick for evaluating the performance of other querying strategies as well. This point will be particularly significant when we evaluate a different querying strategy to enhance the diversity of discovered anomalies as part of this work.
	\item Learning the weights of anomaly detector ensembles with human-in-the-loop learning that generalizes to unseen data helps in limited-memory or streaming data settings. Indeed, our experimental evaluation corroborates this hypothesis.
\end{itemize}

\section{Algorithmic Instantiation of HiLAD for Batch Data Setting}
\label{sec:technical}

In this section, we describe a series of algorithms to instantiate the generic human-in-the-loop learning framework HiLAD for batch data setting (HiLAD-Batch). First, we present a novel formalism called {\em compact description} that describes groups of instances compactly using a tree-based model, and apply it to improve the diversity of instances selected for labeling (Section~\ref{sec:descriptions}). Second, we describe an algorithm to update the weights of the scoring function based on label feedback in the \textit{batch} setting, where the entire data is available at the outset (Section~\ref{sec:weight-update}). 

\subsection{Compact Description for Diversified Querying and Interpretability}
\label{sec:descriptions}

In this section, we first describe the compact description formalism to describe a group of instances. Subsequently, we propose algorithms for selecting diverse instances for querying and to generate succinct interpretable rules using compact description.

\vspace{1.0ex}

\begin{figure}[h]
	\centering
	\subfloat[Baseline]{
		\includegraphics[width=0.3\textwidth]{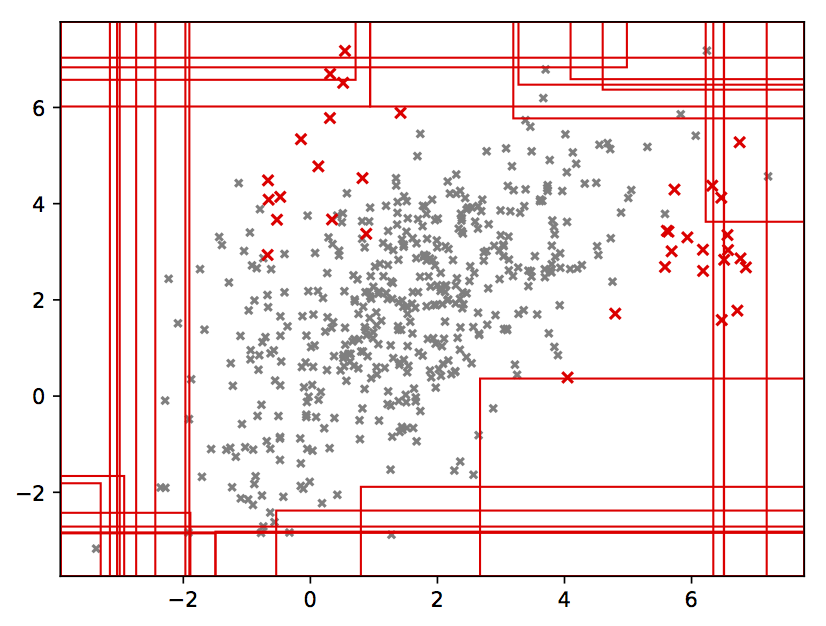}
		\label{fig:baseline_rects}}
	\subfloat[Active Anomaly Detection]{
		\includegraphics[width=0.3\textwidth]{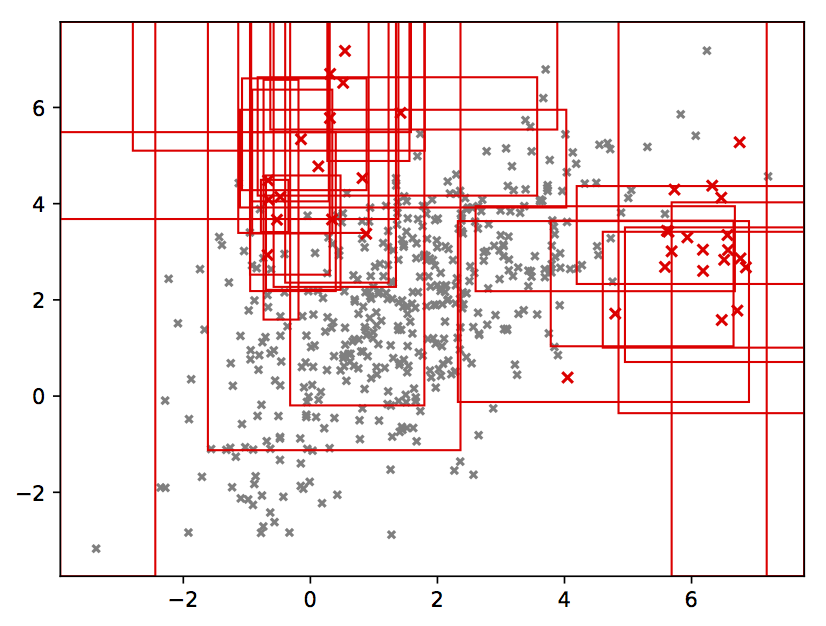}
		\label{fig:aad_rects}}
	\subfloat[Compact Description]{
		\includegraphics[width=0.3\textwidth]{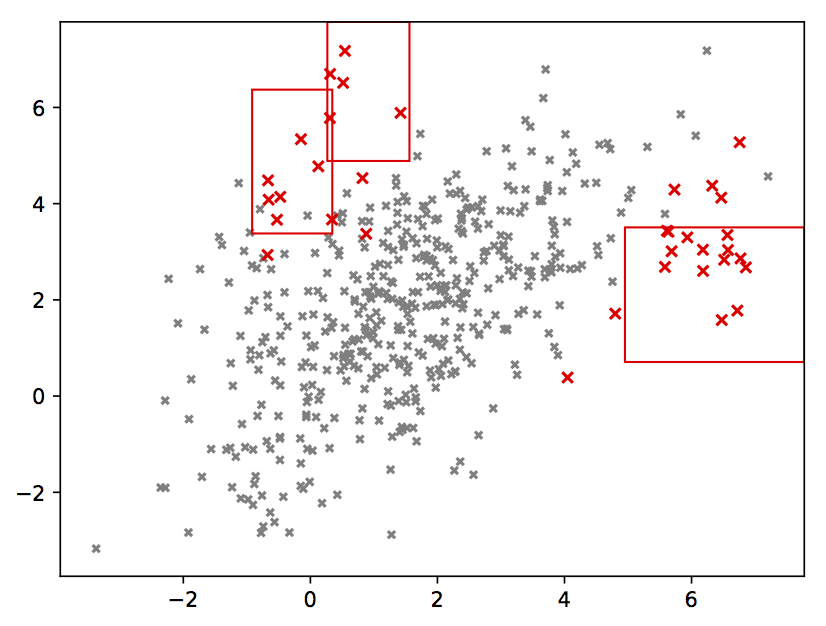}
		\label{fig:compact_rects}}
	\caption{Top $30$ subspaces ranked by ${\bf w}\circ{\bf d}$ (relevance). Red points are anomalies. {\bf (a)} shows the top $30$ most relevant subspaces (w.r.t their \textit{anomalousness}) without any feedback. We can see that initially these subspaces simply correspond to the exterior regions of the dataset. Our human-in-the-loop anomaly detection approach \textbf{learns the true relevance of subspaces} via label feedback. {\bf (b)} shows that after incorporating the labels of $35$ instances, the subspaces around the labeled anomalies have emerged as the most relevant. {\bf (c)} shows the set of \textbf{important} subspaces which compactly cover all labeled anomalies. These were computed by solving Equation~\ref{eqn:compact}. Note that the compact subspaces only cover anomalies that were discovered in the $35$ feedback iterations. Anomalies which were not detected are likely to fall outside these compact subspaces.} \label{fig:rects}
\end{figure}

\noindent {\bf Compact Description (CD).} The tree-based model assigns a weight and an anomaly score to each leaf (i.e., subspace). We denote the vector of leaf-level anomaly scores by ${\bf d}$, and the overall anomaly scores of the subspaces (corresponding to the leaf-nodes) by ${\bf a} = \left[a_1, ..., a_m\right] = {\bf w}\circ{\bf d}$, where $\circ$ denotes element-wise product operation. The score $a_i$ provides a good measure of the \emph{relevance} of the $i$-th subspace. This relevance for each subspace is determined automatically through the label feedback. Our goal is to select a small subset of the most relevant and ``compact'' (by volume) subspaces which together contain all the instances in a group that we want to describe. We treat this problem as a specific instance of the \emph{set covering} problem. We illustrate this idea on a synthetic dataset in Figure~\ref{fig:rects}. This approach can be potentially interpreted as a form of non-parametric clustering.

Let $\mathcal{Z}$ be the set of instances that we want to describe, where $|\mathcal{Z}|=p$. For example, $\mathcal{Z}$ could correspond to the set of anomalous instances discovered by our human-in-the-loop learning approach. Let ${\bf s}_i$ be the $\delta$ most relevant subspaces (i.e., leaf nodes) which contain ${\bf z}_i \in \mathcal{Z}, i = {1, ..., p}$. Let $\mathcal{S}=s_1 \cup ... \cup s_p$ and $|\mathcal{S}|=k$. Denote the \textit{volumes} of the subspaces in $\mathcal{S}$ by the vector ${\bf v} \in \mathbb{R}^k$. Suppose ${\bf x} \in \{0, 1\}^k$ is a binary vector which contains $1$ in locations corresponding to the subspaces in $\mathcal{S}$ which are included in the covering set, and $0$ otherwise. Let ${\bf u}_i \in \{0, 1\}^k$ denote a vector for each instance ${\bf z}_i \in \mathcal{Z}$ which contains $1$ in all locations corresponding to subspaces in $s_i$. Let ${\bf U} = [{\bf u}_1^T, ..., {\bf u}_n^T]^T$. A compact set of subspaces ${\bf S}^*$ which contains (i.e., describes) all the candidate instances can be computed using the optimization formulation in Equation~\ref{eqn:compact}. We employ an off-the-shelf ILP solver (CVX-OPT) to solve this problem.

\begin{align}
{\bf S}^* &= \argmin_{{\bf x} \in \{0, 1\}^k} \; {\bf x} \cdot {\bf v} \label{eqn:compact} \\
\text{s.t. \;} & {\bf U} \cdot {\bf x} \geq {\bf 1} \text{ (where ${\bf 1}$ is a column vector of $p$ 1's)} \nonumber
\end{align}

\noindent {\it Applications of Compact Description.} Compact descriptions have multiple uses including:
\begin{itemize}
	\setlength\itemsep{0em}
	\item Discovery of diverse classes of anomalies very quickly by querying instances from different subspaces of the description.
	\item Improved interpretability of anomalous instances. We assume that in a practical setting, the analyst(s) will be presented with instances along with their corresponding description(s). Additional information can be derived from the description and shown to the analyst (e.g., number of instances in each compact subspace), which can help prioritize the analysis.
\end{itemize}

In this work, we present empirical results on improving query diversity and also compare with another state-of-the-art algorithm that extracts interpretations \cite{wang:2016}.

\begin{algorithm}
	\caption{\texttt{Select-Diverse} (${\mathbf X}$, $b$, $n$)}
	\label{alg:diverse}
	\begin{algorithmic}
		\STATE \textbf{Input:} Unlabeled dataset ${\mathbf X}$, \# instances to select $b$, \# candidate instances $n$ ($n \geq b$)
		\STATE Let $\mathcal{Z}$ = $n$ top-ranked instances as candidates $\subseteq {\mathbf X}$ (blue points in Figure~\ref{fig:query_regions})
		\STATE Let ${\bf S}^*$ = subspaces with Equation~\ref{eqn:compact} that contain $\mathcal{Z}$ (rectangles in Figures \ref{fig:query_baseline} and \ref{fig:query_diverse})
		\STATE Set ${\bf Q} = \emptyset$
		\WHILE{$|{\bf Q}| < b$}
		\STATE Let ${\bf x}$ = instance with highest anomaly score $\in \mathcal{Z}$ s.t. ${\bf x}$ has minimal
		\STATE \hspace{0.54in} overlapping regions in ${\bf S}^*$ with instances in ${\bf Q}$
		\STATE Set ${\bf Q} = {\bf Q} \cup \{{\bf x}\}$ (green circles in Figure~\ref{fig:query_diverse})
		\STATE Set $\mathcal{Z} = \mathcal{Z} \setminus \{{\bf x}\}$
		\ENDWHILE
		\RETURN ${\bf Q}$
	\end{algorithmic}
\end{algorithm}

\begin{figure}[h]
	\centering
	\subfloat[Candidate Subspaces]{\includegraphics[width=0.3\textwidth]{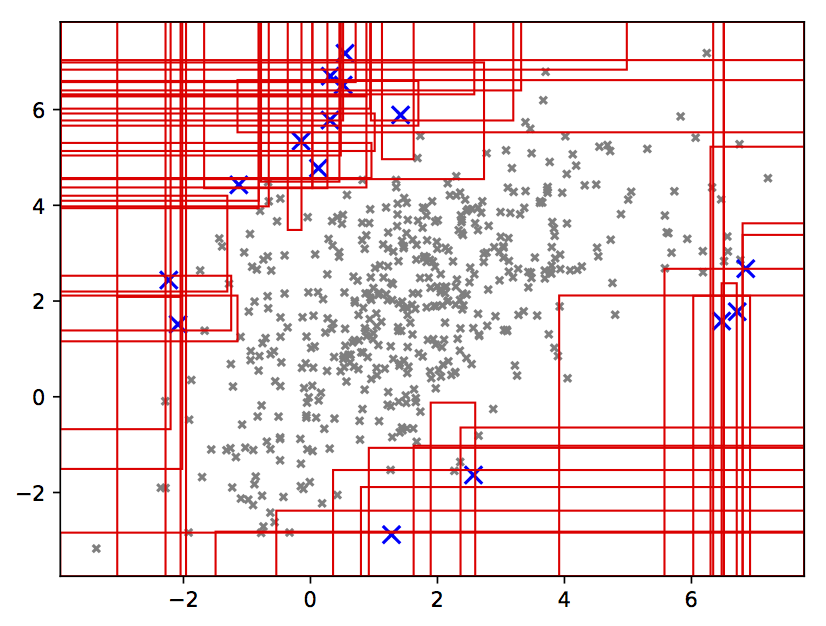}%
		\label{fig:query_regions}}
	\subfloat[Select-Top]{\includegraphics[width=0.3\textwidth]{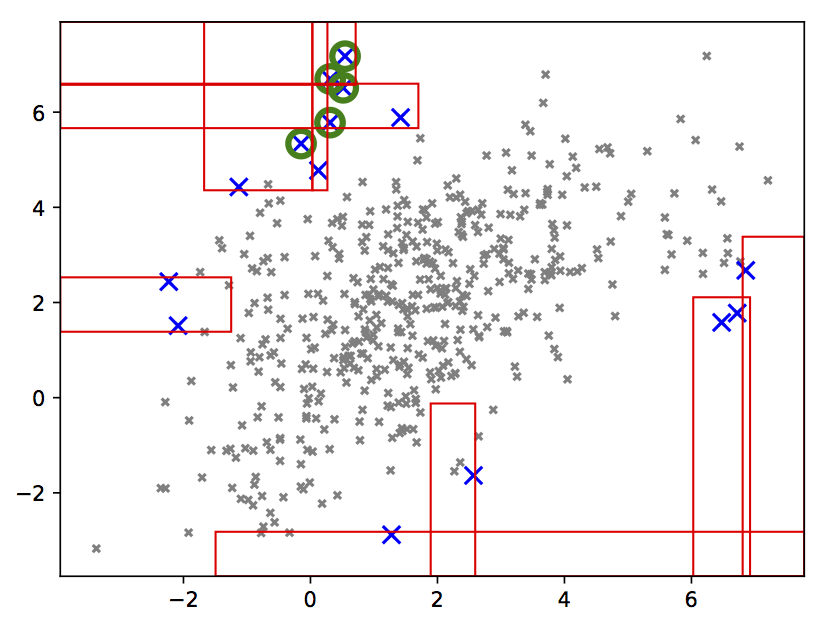}%
		\label{fig:query_baseline}}
	\subfloat[Select-Diverse]{\includegraphics[width=0.3\textwidth]{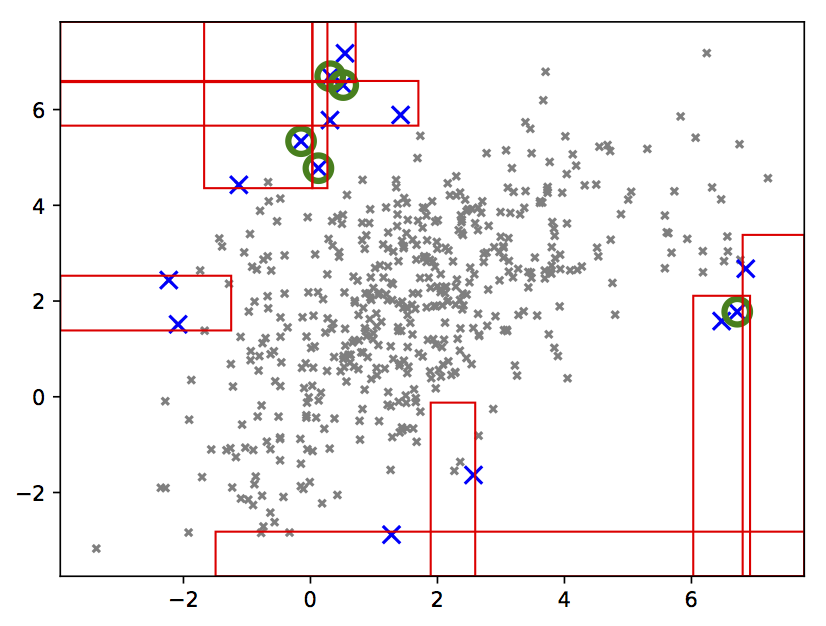}%
		\label{fig:query_diverse}}
	\caption{Illustration of compact description and diversity of selected instances for labeling using IFOR. Most anomalous $15$ instances (blue checks) are selected as the query candidates. The red rectangles in {\bf (a)} form the union of the $\delta$ ($= 5$ works well in practice) most \emph{relevant} subspaces across each of the query candidates. {\bf (b)} and {\bf (c)} show the most ``compact'' set of subspaces which together cover all the query candidates. {\bf (b)} shows the most anomalous $5$ instances (green circles) selected by the greedy \texttt{Select-Top} strategy. {\bf (c)} shows the $5$ ``diverse'' instances (green circles) selected by \texttt{Select-Diverse}.}
	\label{fig:query}
\end{figure}

\noindent {\bf Diversity-based Query Selection Strategy.} In Section~\ref{sec:insights}, we reasoned that the greedy strategy of selecting top-scored instances (referred as \texttt{Select-Top}) for labeling is efficient. However, this strategy might lack diversity in the types of instances presented to the human analyst. It is likely that different types of instances belong to different subspaces in the original feature space. Our proposed strategy (\texttt{Select-Diverse}), which is described next, is intended to increase the diversity by employing tree-based ensembles and compact description to select groups of instances from subspaces that have minimum overlap.

Assume that the human analyst can label a batch of $b$ instances, where $b > 1$, in each feedback iteration. Algorithm~\ref{alg:diverse} employs the compact description to achieve this diversity. The algorithm first selects $n$ ($> b$) top-ranked anomalous instances $\mathcal{Z}$ and computes the corresponding compact description (small set of subspaces ${\bf S}^*$). Subsequently, performs an iterative selection of $b$ instances from $\mathcal{Z}$ by minimizing the overlap in the corresponding subspaces from ${\bf S}^*$. Figure~\ref{fig:query} provides an illustration comparing \texttt{Select-Top} and \texttt{Select-Diverse} query selection strategies.

\subsection{Algorithmic Approach to Update Weights of Scoring Function}
\label{sec:weight-update}

In this section, we provide an algorithm to update weights of the scoring function for HiLAD-Batch instantiation for batch setting: the entire data ${\bf D}$ is available at the outset. 

\begin{algorithm}[h]
	\caption{\texttt{HiLAD-Batch} ($B$, ${\mathbf w}^{(0)}$, ${\mathbf H}$, ${\mathbf H}_+$, ${\mathbf H}_-$)}
	\label{alg:batch}
	\begin{algorithmic}[14]
		\STATE \textbf{Input:} Query budget $B$, initial weights ${\mathbf w}^{(0)}$, unlabeled instances ${\mathbf H}$, 
		\STATE \hspace{0.1in} labeled instances ${\mathbf H}_+$ and ${\mathbf H}_-$
		\STATE Set $t=0$
		\WHILE{$t \leq B$}
		\STATE Set $t = t + 1$
		\STATE Set ${\mathbf a} = {\mathbf H} \cdot {\mathbf w}$ (i.e., ${\mathbf a}$ is the vector of anomaly scores)
		\STATE Let ${\bf q} = {\mathbf z}_i$, where $i = \argmax_{i}(a_i)$
		\STATE Get $y_i \in \{-1,+1\}$ for ${\mathbf q}$ from analyst
		\IF{$y_i = +1$}
		\STATE Set ${\mathbf H}_+ = \{{\bf z}_i\} \cup {\mathbf H}_+$
		\ELSE
		\STATE Set ${\mathbf H}_- = \{{\bf z}_i\} \cup {\mathbf H}_-$
		\ENDIF
		\STATE Set ${\mathbf H} = {\mathbf H} \setminus {\mathbf z}_i$
		\STATE ${\mathbf w}^{(t)}$ = learn new weights; normalize $\|{\mathbf w}^{(t)}\|=1$ \label{alg:batch:weightupdate_}
		\ENDWHILE
		\RETURN ${\mathbf w}^{(t)}$, ${\mathbf H}$, ${\mathbf H}_+$, ${\mathbf H}_-$
	\end{algorithmic}
\end{algorithm}
Recall that our scoring function is of the following form: \texttt{Score}(${\bf x}$) = ${\mathbf w} \cdot {\bf z}$, where ${\bf z} \in \mathds{R}^m$ corresponds to the scores from anomaly detectors for instance ${\bf x}$. We extend the AAD approach (based on LODA projections) \cite{das:2016} to update the weights for tree-based models. AAD makes the following assumptions: (1) $\tau$ fraction of instances (i.e., $n\tau$) are anomalous, and (2) Anomalies should lie above the optimal hyperplane while nominals should lie below. AAD tries to satisfy these assumptions by enforcing constraints on the labeled examples while learning the weights of the hyperplane. If the anomalies are rare and we set $\tau$ to a small value, then the two assumptions make it more likely that the hyperplane will pass through the region of uncertainty. Our previous discussion then suggests that the optimal hyperplane can now be learned efficiently by greedily asking the analyst to label the most anomalous instance in each feedback iteration. We simplify the AAD formulation with a more scalable unconstrained optimization objective. Crucially, the ensemble weights are updated with an intent to maintain the hyperplane in the region of uncertainty through the entire budget $B$. The HiLAD-Batch learning approach is presented in Algorithm~\ref{alg:batch} and depends on only one hyper-parameter $\tau$.

\begin{figure}[htb]
	\centering
	\subfloat[Initial scores]{\includegraphics[width=.3\linewidth]{figures/iter_00} \label{fig:iter_00}}
	\subfloat[8 iterations]{\includegraphics[width=.3\linewidth]{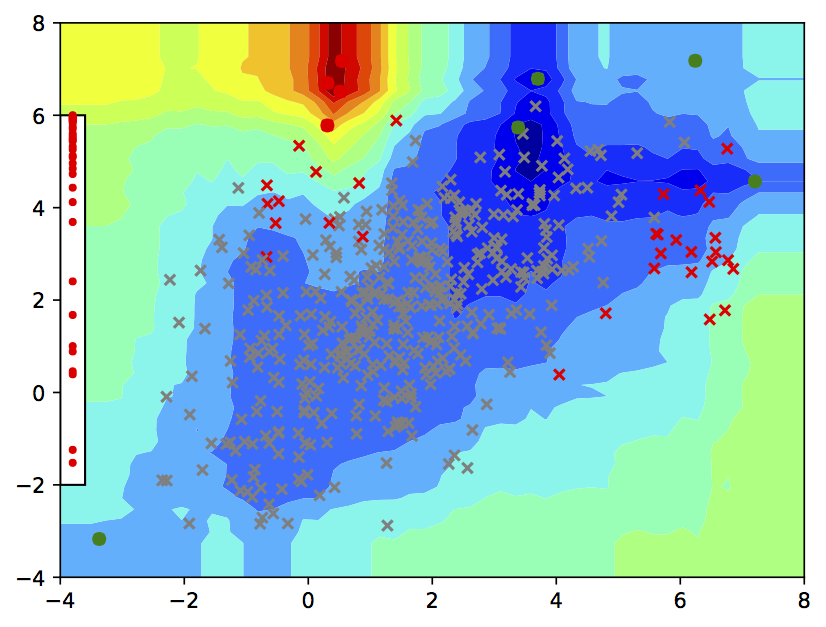} \label{fig:iter_08}}
	\subfloat[16 iterations]{\includegraphics[width=.3\linewidth]{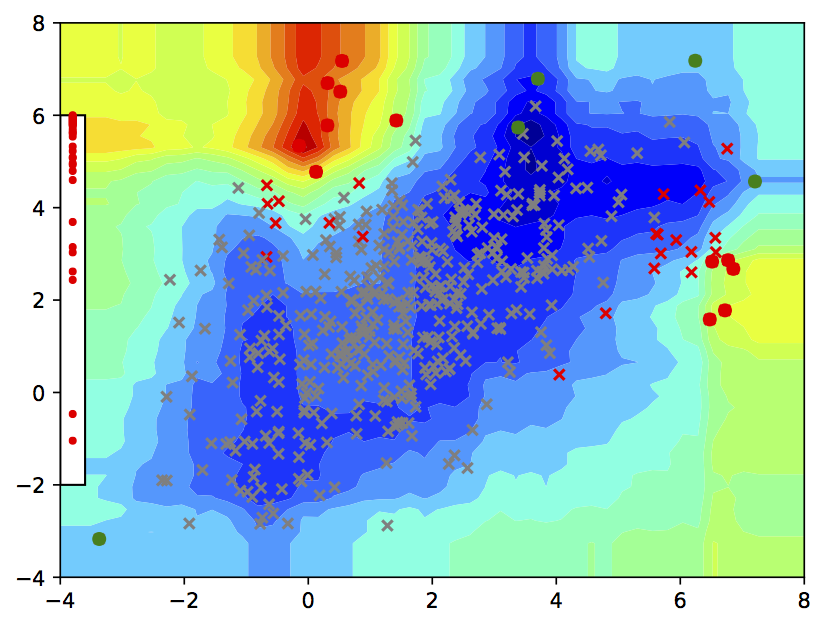} \label{fig:iter_16}} \\
	\subfloat[24 iterations]{\includegraphics[width=.3\linewidth]{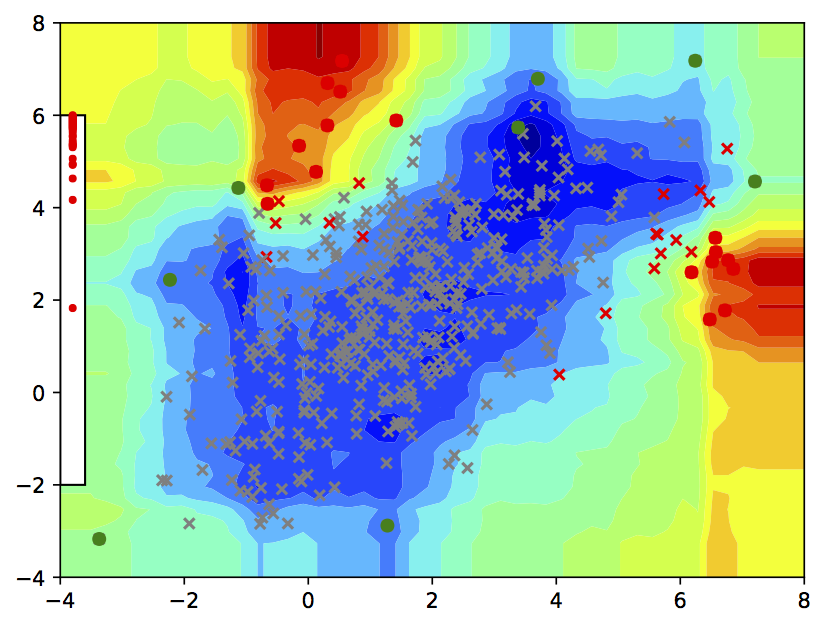} \label{fig:iter_24}}
	\subfloat[32 iterations]{\includegraphics[width=.3\linewidth]{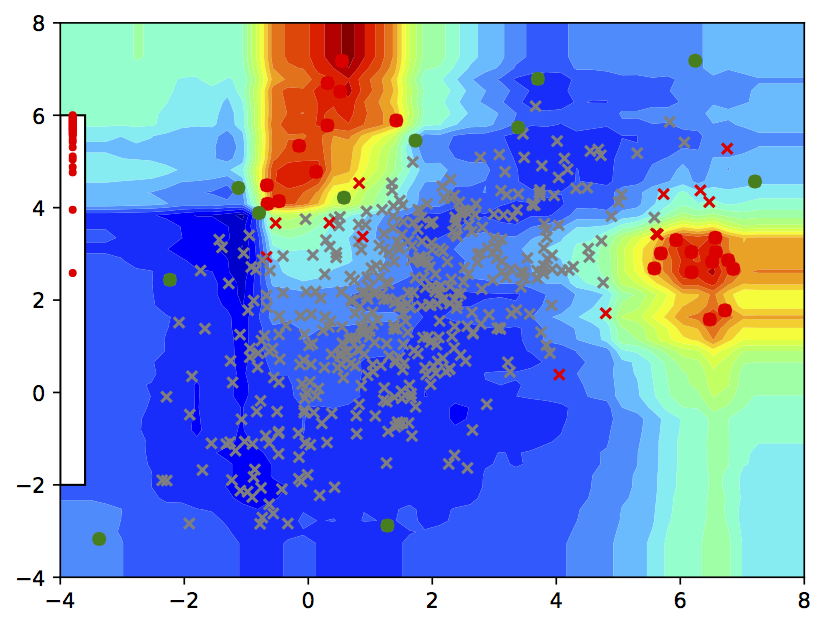} \label{fig:iter_32}}
	\subfloat[34 iterations]{\includegraphics[width=.3\linewidth]{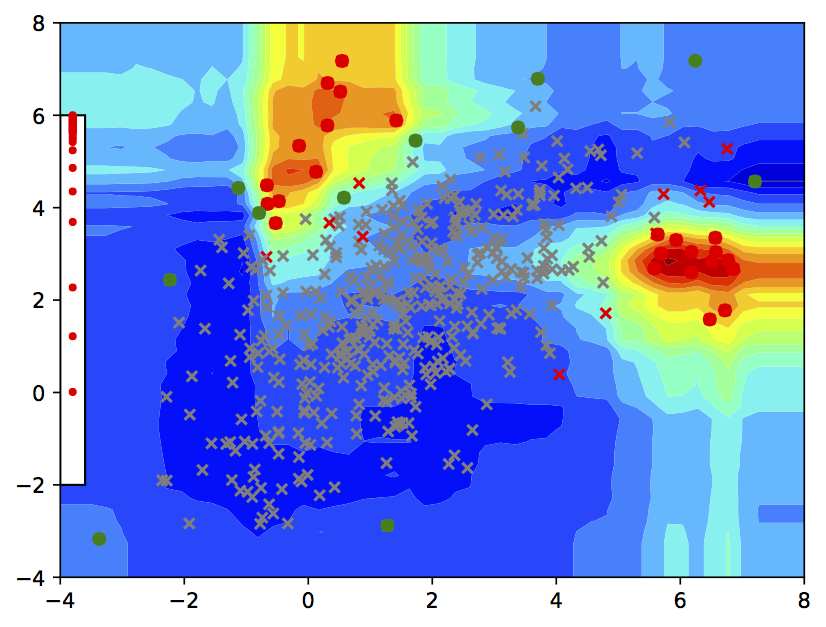} \label{fig:iter_34}}
	\caption{Score contours across $34$ \texttt{HiLAD-Batch} feedback iterations on the Toy dataset (Figure~\ref{fig:synthetic_dataset}).}
	\label{fig:bal_iters}
\end{figure}

We first define the hinge loss $\ell(q, {\mathbf w}; ({\mathbf z}_i, y_i))$ in Equation~\ref{eqn:aatploss} that penalizes the model when anomalies are assigned scores lower than $q$ and nominals higher. Equation~\ref{eqn:preflearn_aatp} then formulates the optimization problem for learning the optimal weights in Line~14 of the batch algorithm (Algorithm~\ref{alg:batch}). Figure~\ref{fig:bal_iters} illustrates how HiLAD-Batch changes the anomaly score contours across feedback iterations on the synthetic dataset using an isolation forest with $100$ trees.

\begin{align}
& \ell(q, {\mathbf w}; ({\mathbf z}_i, y_i)) = \nonumber \\ 
& \left \{ 
\begin{array}{lr}
0 & {\mathbf w}\cdot {\mathbf z}_i \ge q \text{ and $y_i=+1$} \\
0 & {\mathbf w}\cdot {\mathbf z}_i < q \text{ and $y_i=-1$} \\
(q - {\mathbf w}\cdot {\mathbf z}_i) & {\mathbf w}\cdot {\mathbf z}_i < q \text{ and $y_i=+1$} \\
({\mathbf w}\cdot {\mathbf z}_i - q) & {\mathbf w}\cdot {\mathbf z}_i \ge q \text{ and $y_i=-1$}
\end{array} 
\right. \label{eqn:aatploss}
\end{align}

$\lambda^{(t)}$ determines the influence of the prior. For the batch setup, we set $\lambda^{(t)}=\frac{0.5}{|{\bf H}_+|+|{\bf H}_-|}$ such that the prior becomes less important as more instances are labeled. When there are no labeled instances, $\lambda^{(t)}$ is set to $\frac{1}{2}$. The third and fourth terms of Equation~\ref{eqn:preflearn_aatp} encourage the scores of anomalies in $\mathbf{H}_+$ to be higher than that of ${\mathbf z}_{\tau}^{(t-1)}$ (the $n\tau$-th ranked instance from the previous iteration), and the scores of nominals in $\mathbf{H}_-$ to be lower than that of ${\mathbf z}_{\tau}^{(t-1)}$. We employ gradient descent to learn the optimal weights ${\mathbf w}$ in Equation~\ref{eqn:preflearn_aatp}. Our prior knowledge that ${\bf w}_{unif}$ is a good prior provides a good initialization for gradient descent. We later show empirically that ${\bf w}_{unif}$ is a better starting point than random weights.

\begin{align}
{\mathbf w}^{(t)} &= \argmin_{{\mathbf w}} \sum_{s \in \{-, +\}} \left( \frac{1}{|{\mathbf H}_s|} \sum_{{\mathbf z}_i \in {\mathbf H}_s}\ell(\hat{q}_{\tau}( {{\mathbf w}^{(t-1)}}), {\mathbf w}; ({\mathbf z}_i, y_i)) \right. \nonumber \\
& \left. \qquad \qquad + \frac{1}{|{\mathbf H}_s|} \sum_{{\mathbf z}_i \in {\mathbf H}_s}\ell({\mathbf z}_{\tau}^{(t-1)} \cdot {\mathbf w}, {\mathbf w}; ({\mathbf z}_i, y_i)) \right) \nonumber \\
& \qquad \qquad + \lambda^{(t)}\|{\mathbf w} - {\mathbf w}_{unif}\|^2 \label{eqn:preflearn_aatp} \\
& \text{where, } \mathbf{w}_{unif} = [\frac{1}{\sqrt{m}},\ldots,\frac{1}{\sqrt{m}}]^T, \text{ and,} \nonumber \\ 
&{\mathbf z}_{\tau}^{(t-1)} \text{ and } \hat{q}_{\tau}({{\mathbf w}^{(t-1)}}) \text{ are computed by ranking} \nonumber \\
&\text{anomaly scores with ${\mathbf w} = {\mathbf w}^{(t-1)}$} \nonumber
\end{align}

\section{Algorithmic Instantiation of HiLAD for Streaming Data Setting}
\label{sec:streaming} 

\begin{algorithm}[]
	\caption{\texttt{HiLAD-Stream} ($K$, $B$, $Q$, $\mathcal{E}^{(0)}$, ${\mathbf X}_0$, ${\mathbf w}^{(0)}$, $\alpha_{KL}$)}
	\label{alg:stream}
	\begin{algorithmic}
		\STATE \textbf{Input:} Stream window size $K$, total query budget $B$, 
		\STATE \hspace{0.1in} queries per window $Q$, anomaly detector ensemble $\mathcal{E}^{(0)}$, 
		\STATE \hspace{0.1in} initial instances ${\mathbf X}_0$ (used to create $\mathcal{E}^{(0)}$), initial weights ${\mathbf w}^{(0)}$,
		\STATE \hspace{0.1in} significance level $\alpha_{KL}$
		\STATE
		\STATE Set ${\mathbf H} = {\mathbf H}_+ = {\mathbf H}_- = \emptyset$
		\STATE // initialize KL-divergence baselines
		\STATE Set $q_{KL}^{(0)}$ = \texttt{Get-KL-Threshold}(${\mathbf X}_0$, $\mathcal{E}^{(0)}$, $\alpha_{KL}$, $10$)
		\STATE Set $\mathcal{P}^{(0)}$ = \texttt{Get-Ensemble-Distribution}(${\mathbf X}_0$, $\mathcal{E}^{(0)}$)
		\STATE
		\STATE Set $t=0$
		\WHILE{$<$stream is not empty$>$}
		\STATE Set $t = t + 1$
		\STATE Set ${\mathbf X}_t$ = $K$ data instances from stream
		\STATE Set ${\mathbf H}_t$ = transform ${\mathbf X}_t$ to ensemble features
		\STATE 
		\STATE // \texttt{Update-Model} either updates node counts (e.g., for HST and RSF), 
		\STATE // or replaces a fraction of the oldest trees in $\mathcal{E}$ with new
		\STATE // ones constructed using ${\mathbf X}_t$ (e.g., for IFOR)
		\STATE Set $\mathcal{E}^{(t)}$, $q_{KL}^{(t)}$, $\mathcal{P}^{(t)}$ = \texttt{Update-Model}(${\mathbf X}_t$, $\mathcal{E}^{(t-1)}$, $q_{KL}^{(t-1)}$, $\mathcal{P}^{(t-1)}$, $\alpha_{KL}$)
		\STATE 
		\STATE // \texttt{Merge-and-Retain}(${\mathbf w}$, ${\mathbf H}$, $K$) retains $K$ most anomalous instances in ${\mathbf H}$
		\STATE Set ${\mathbf H} =$ \texttt{Merge-and-Retain}(${\mathbf w}^{(t-1)}$, $\{{\mathbf H} \cup {\mathbf H}_t\}$, $K$)
		\STATE Set ${\mathbf w}^{(t)}$, ${\mathbf H}$, ${\mathbf H}_+$, ${\mathbf H}_-$ = \texttt{HiLAD-Batch}($Q$, ${\mathbf w}^{(t-1)}$, ${\mathbf H}$, ${\mathbf H}_+$, ${\mathbf H}_-$)
		\ENDWHILE
	\end{algorithmic}
\end{algorithm}
In this section, we describe algorithms to support human-in-the-loop anomaly detection using tree-based ensembles in the \textit{streaming} data setting (HiLAD-Stream), where the data comes as a continuous stream.

In the streaming setting, we assume that the data is input to the algorithm continuously in \textit{windows} of size $K$ and is potentially unlimited. The HiLAD-Stream instantiation is shown in Algorithm~\ref{alg:stream}. Initially, we train all the members of the ensemble with the first window of data. When a new window of data arrives, the underlying tree model is updated as follows: in case the model is an HST or RSF, only the node counts are updated while keeping the tree structures and weights unchanged; whereas, if the model is an IFOR, a subset of the current set of trees is replaced as shown in \texttt{Update-Model} (Algorithm~\ref{alg:stream}). The updated model is then employed to determine which unlabeled instances to retain in memory, and which to ``forget''. This step, referred to as \texttt{Merge-and-Retain}, applies the simple strategy of retaining only the most anomalous instances among those in the memory and in the current window, and discarding the rest. Next, the weights are fine-tuned with analyst feedback through human-in-the-loop learning loop similar to the batch setting with a small budget $Q$. Finally, the next window of data is read, and the process is repeated until the stream is empty or the total budget $B$ is exhausted. {\em In the rest of this section, we will assume that the underlying tree model is IFOR.}
\begin{algorithm}[H]
	\caption{\texttt{Update-Model} (${\mathbf X}$, $\mathcal{E}$, $q_{KL}$, $\mathcal{P}$, $\alpha_{KL}$)}
	\label{alg:update_model}
	\begin{algorithmic}
		\STATE \textbf{Input:} Instances ${\mathbf X}$, anomaly detector ensemble $\mathcal{E}$,
		\STATE \hspace{0.1in} current KL threshold $q_{KL}$, baseline distributions $\mathcal{P}$, 
		\STATE \hspace{0.1in} significance level $\alpha_{KL}$
		\STATE Set $T$ = number of trees in $\mathcal{E}$
		\STATE Set $\mathcal{Q} $ = \texttt{Get-Ensemble-Distribution}(${\mathbf X}, \mathcal{E}$)
		\STATE Initialize $\mathbf{KL} = {\mathbf 0} \in \mathbb{R}^T$
		\FOR{$t \in 1..T$}
		\STATE Set $KL_t = D_{KL}(\mathcal{P}_t || \mathcal{Q}_t)$
		\ENDFOR
		\STATE Set $S = \{t : KL_t > q_{KL}\}$
		\IF{$|S| < 2\alpha_{KL}T$}
		\STATE // the number of trees with divergence is not significant
		\RETURN $\mathcal{E}, q_{KL}, \mathcal{P}$
		\ENDIF
		\STATE Set $\mathcal{E}'$ = replace trees in $\mathcal{E}$ whose indexes are in $S$, with new trees trained using $\mathbf{X}$
		\STATE // Recompute threshold and baseline distributions
		\STATE Set $q_{KL}'$ = \texttt{Get-KL-Threshold}(${\mathbf X}, \mathcal{E}'$, $\alpha_{KL}$, $10$)
		\STATE Set $\mathcal{P}'$ = \texttt{Get-Ensemble-Distribution}(${\mathbf X}, \mathcal{E}'$)
		\RETURN $\mathcal{E}', q_{KL}', \mathcal{P}'$
	\end{algorithmic}
\end{algorithm}
When we replace a tree in \texttt{Update-Model}, its leaf nodes and corresponding weights get discarded. On the other hand, adding a new tree implies adding all its leaf nodes with weights initialized to a default value $v$. We first set $v = \frac{1}{\sqrt{m'}}$ where $m'$ is the total number of leaves in the new model, and then re-normalize the updated ${\bf w}$ to unit length. The HiLAD instantiation for streaming data is presented in Algorithm~\ref{alg:stream}. In all the HiLAD-Stream experiments, we set the number of queries per window $Q=20$, and $\lambda^{(t)}=\frac{1}{2}$.

HiLAD-Stream approach can be employed in two different situations: (1) \textbf{limited memory with no concept drift}, and (2) \textbf{streaming data with concept drift}. The type of situation determines how fast the model needs to be updated in \texttt{Update-Model}. If there is no concept drift, we need not update the model at all. If there is a large change in the distribution of data from one window to the next, then a large fraction of members need to be replaced. When we replace a member tree in our tree-based model, all its corresponding nodes along with their learned weights have to be discarded. Thus, some of the ``knowledge'' is lost with the model update. In general, it is hard to determine the true rate of drift in the data. One approach is to replace, in the \texttt{Update-Model} step, a reasonable number (e.g. 20\%) of older ensemble members with new members trained on new data. Although this ad hoc approach often works well in practice, a more principled approach is preferable.

\begin{algorithm}[H]
	\caption{\texttt{Get-KL-Threshold} (${\mathbf X}$, $\mathcal{E}$, $\alpha_{KL}$, $n$)}
	\label{alg:get_KL_threshold}
	\begin{algorithmic}
		\STATE \textbf{Input:} Instances ${\mathbf X}$, anomaly detector ensemble $\mathcal{E}$, 
		\STATE \hspace{0.1in} significance level $\alpha_{KL}$,
		\STATE \hspace{0.1in} repetitions of KL-divergence computations $n$
		\STATE Set $T$ = number of trees in $\mathcal{E}$
		\STATE Initialize ${\mathbf {KL}} = {\mathbf 0} \in \mathbb{R}^T$ // mean KL-divergence for each tree
		\FOR{$i$ in $1 \cdots n$}
		\STATE Partition ${\mathbf X}$ = $\{{\mathbf A}, {\mathbf B}\}$ at random s.t. ${\mathbf X} = {\mathbf A} \cup {\mathbf B}$ and $|{\mathbf A}| \approx |{\mathbf B}|$
		\FOR{$t \in 1 \cdots T$}
		\STATE Let $\mathcal{T}_t = t\text{-th tree in } \mathcal{E}$
		\STATE Set ${\mathbf p}_A$ = \texttt{Get-Tree-Distribution}(${\mathbf A}, \mathcal{T}_t$)
		\STATE Set ${\mathbf p}_B$ = \texttt{Get-Tree-Distribution}(${\mathbf B}, \mathcal{T}_t$)
		\STATE Set $KL_{t} = KL_{t} + D_{KL}({\mathbf p}_A || {\mathbf p}_B)$
		\ENDFOR
		\ENDFOR
		\STATE Set ${\mathbf {KL}} = \frac{{\mathbf {KL}}}{n}$ \hspace{0.25in}// average the values
		\STATE Set $q_{KL} = (1-\alpha_{KL})\times100 \text{ quantile value in } {\mathbf {KL}}$
		\RETURN $q_{KL}$
	\end{algorithmic}
\end{algorithm}

\noindent {\bf Drift Detection Algorithm.}
Algorithm~\ref{alg:update_model} presents a principled methodology that employs KL-divergence (denoted by $D_{KL}$) to determine which trees should be replaced. The set of all leaf nodes in a tree are treated as a set of histogram bins which are then used to estimate the data distribution. We denote the total number of trees in the model by $T$, and the $t$-th tree by $\mathcal{T}_t$. When $\mathcal{T}_t$ is initially created with the first window of data, the data from the same window is also used to initialize the {\em baseline} distribution for $\mathcal{T}_t$, denoted by ${\bf p}_t$ (\texttt{Get-Ensemble-Distribution} in Algorithm~\ref{alg:get_ensemble_distribution}). After computing the baseline distributions for each tree, we estimate the $D_{KL}$ threshold $q_{KL}$ at the $\alpha_{KL}$ (typically $0.05$) significance level by sub-sampling (\texttt{Get-KL-Threshold} in Algorithm~\ref{alg:get_KL_threshold}). When a new window is read, we first use it to compute the new distribution ${\bf q}_t$ (for $\mathcal{T}_t$). Next, if ${\bf q}_t$ differs {\em significantly} from ${\bf p}_t$ (i.e., $D_{KL}({\bf p}_t||{\bf q}_t) > q_{KL}$) for at least $2\alpha_{KL}T$ trees, then we replace all such trees with new ones created using the data from the new window. Finally, if any tree in the forest is replaced, then the baseline densities for all trees are recomputed with the data in the new window.

\begin{algorithm}[h]

	\caption{\texttt{Get-Ensemble-Distribution} (${\mathbf X}$, $\mathcal{E}$)}
	\label{alg:get_ensemble_distribution}
	\begin{algorithmic}
		\STATE \textbf{Input:} Instances ${\mathbf X}$, ensemble $\mathcal{E}$
		\STATE Set $T$ = number of trees in $\mathcal{E}$
		\STATE Set ${\mathcal{P}} = \emptyset$
		\FOR{$t \in 1 \cdots T$}
		\STATE Let $\mathcal{T}_t = t\text{-th tree in } \mathcal{E}$
		\STATE Set ${\mathbf p}_t $ = \texttt{Get-Tree-Distribution}(${\mathbf X}, \mathcal{T}_t$)
		\STATE Set $\mathcal{P} = \mathcal{P} \cup {\mathbf p}_t$
		\ENDFOR
		\RETURN $\mathcal{P}$
	\end{algorithmic}
\end{algorithm}

\begin{algorithm}[H]

	\caption{\texttt{Get-Tree-Distribution} (${\mathbf X}$, $\mathcal{T}$)}
	\label{alg:get_tree_distribution}
	\begin{algorithmic}
		\STATE \textbf{Input:} Instances ${\mathbf X}$, tree $\mathcal{T}$
		\STATE Set ${\mathbf p} = $ distribution of instances in ${\mathbf X}$ at the leaves of $\mathcal{T}$
		\RETURN ${\mathbf p}$
	\end{algorithmic}
\end{algorithm}

\section{Experiments and Results}
\label{sec:experiments}

\noindent \textbf{Datasets.} We evaluate our human-in-the-loop learning framework on ten publicly available benchmark datasets (\cite{woods:1993},\cite{ditzler:2013},\cite{harries:1999}, UCI\cite{uci}) listed in Table~\ref{tab:datasets_full}. The anomaly classes in \textit{Electricity} and \textit{Weather} were down-sampled to be 5\% of the total.

\begin{table}[h]
\centering
\caption{Description of benchmark datasets used in our experiments.}
\resizebox{\textwidth}{!}{%
\begin{tabular}{lccrrr}
\hline
\multicolumn{1}{c}{\textbf{Dataset}} & \textbf{Nominal Class} & \textbf{Anomaly Class} & \textbf{Total} & \textbf{Dims} & \textbf{Anomalies(\%)} \\ \hline
Abalone          & 8, 9, 10               & 3, 21                   & 1920   & 9  & 29 (1.5\%)    \\
ANN-Thyroid-1v3  & 3                      & 1                       & 3251   & 21 & 73 (2.25\%)   \\
Cardiotocography & 1 (Normal)             & 3 (Pathological)        & 1700   & 22 & 45 (2.65\%)   \\
KDD-Cup-99       & \textit{`normal'}      & \textit{`u2r', `probe'} & 63009  & 91 & 2416 (3.83\%) \\
Mammography      & -1                     & +1                      & 11183  & 6  & 260 (2.32\%)  \\
Shuttle          & 1                      & 2, 3, 5, 6, 7           & 12345  & 9  & 867 (7.02\%)  \\
Yeast            & \textit{CYT, NUC, MIT} & \textit{ERL, POX, VAC}  & 1191   & 8  & 55 (4.6\%)    \\ \hline
Covtype          & 2                      & 4                       & 286048 & 54 & 2747 (0.9\%)  \\
Electricity      & \textit{DOWN}                   & \textit{UP}                      & 27447  & 13 & 1372 (5\%)    \\
Weather          & \textit{No Rain}                & \textit{Rain}                    & 13117  & 8  & 656 (5\%)     \\ \hline
\end{tabular}%
}
\label{tab:datasets_full}
\end{table}

\noindent {\bf Evaluation Methodology.} For each variant of the human-in-the-loop framework HiLAD, we plot the percentage of the total number of anomalies shown to the analyst versus the number of instances queried; this is the most relevant metric for an analyst in any real-world application. A higher plot means the algorithmic instantiation of the framework is better in terms of discovering anomalies. All results presented are averaged over 10 different runs and the error-bars represent $95\%$ confidence intervals.

\begin{figure*}[h]
	\centering
	\subfloat[Abalone]{\includegraphics[width=0.24\textwidth]{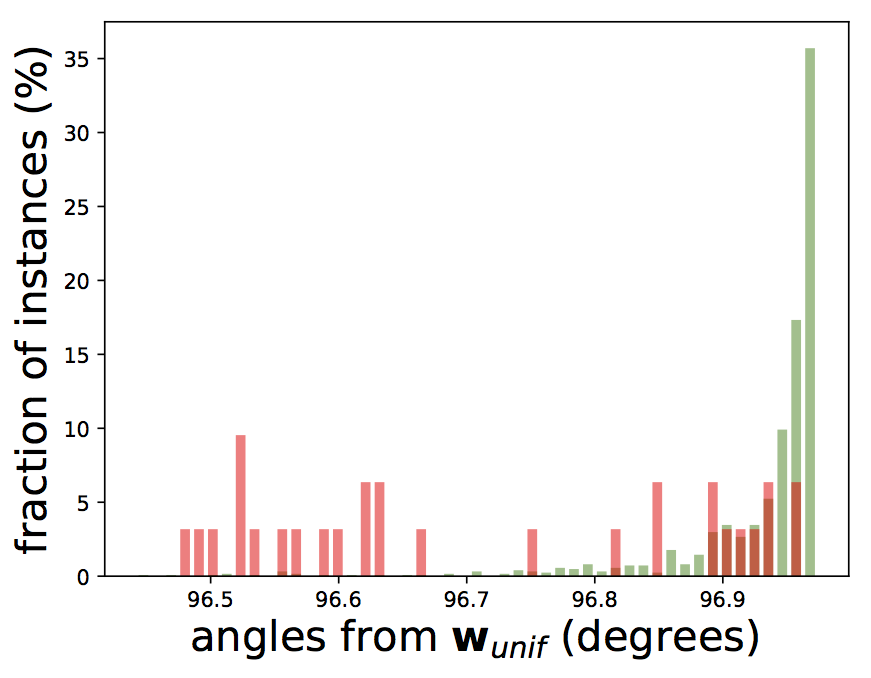}%
		\label{fig:angles_abalone}}
	\subfloat[ANN-Thyroid]{\includegraphics[width=0.24\textwidth]{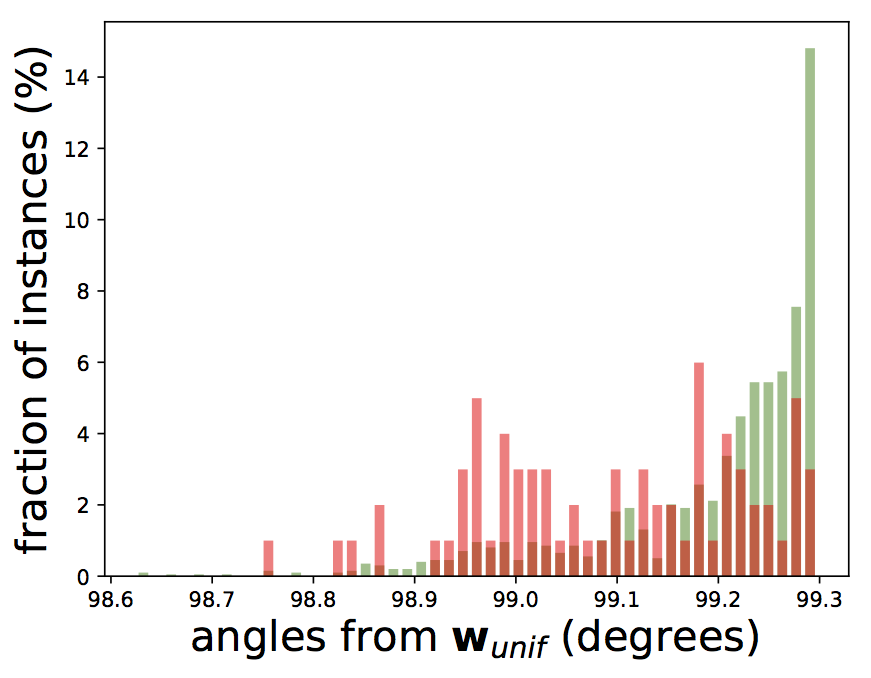}%
		\label{fig:angles_ann_thyroid_1v3}}
	\subfloat[Cardiotocography]{\includegraphics[width=0.24\textwidth]{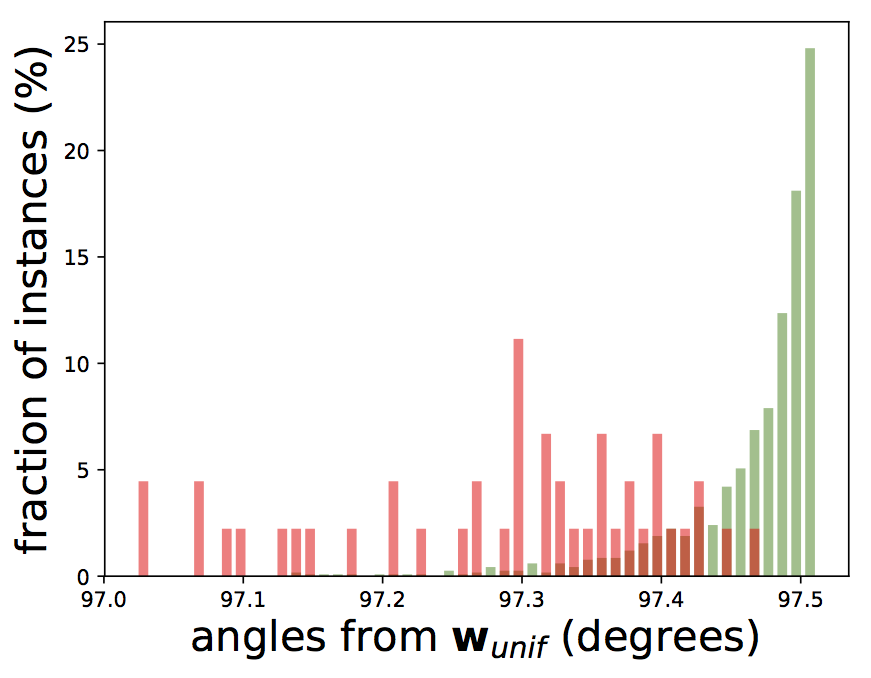}%
		\label{fig:angles_cardiotocography}}
	\subfloat[Covtype]{\includegraphics[width=0.24\textwidth]{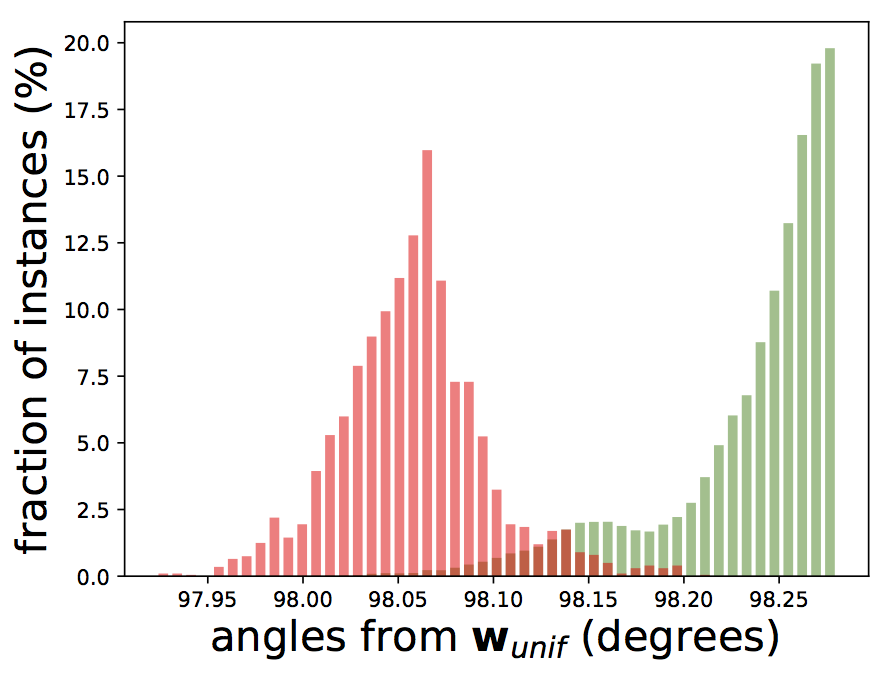}%
	\label{fig:angles_covtype}} \\
	\subfloat[Electricity]{\includegraphics[width=0.24\textwidth]{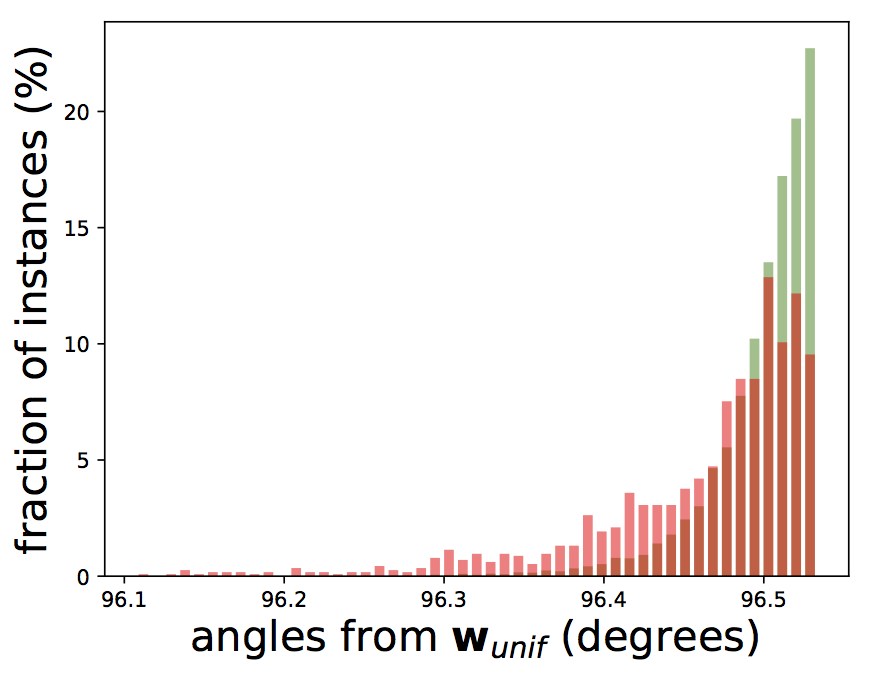}%
		\label{fig:angles_electricity}}
	\subfloat[KDDCup99]{\includegraphics[width=0.24\textwidth]{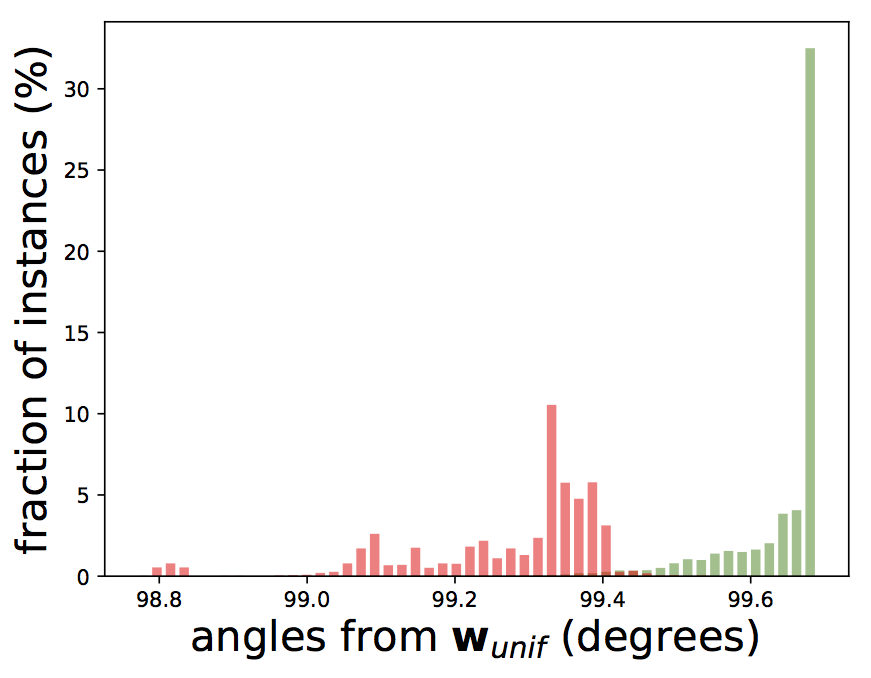}%
		\label{fig:angles_kddcup}}
	\subfloat[Mammography]{\includegraphics[width=0.24\textwidth]{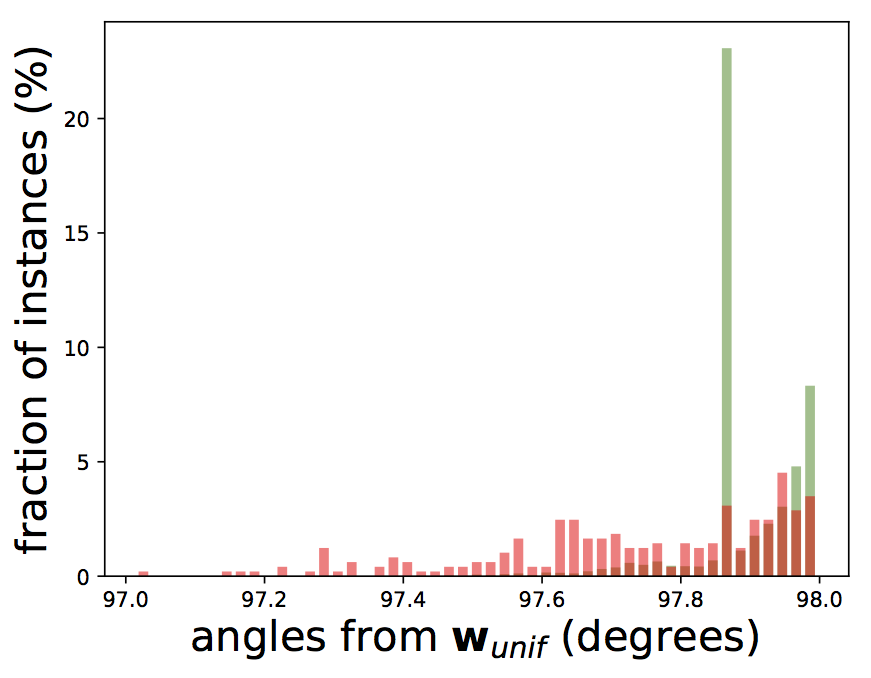}%
		\label{fig:angles_mammography}}
	\subfloat[Shuttle]{\includegraphics[width=0.24\textwidth]{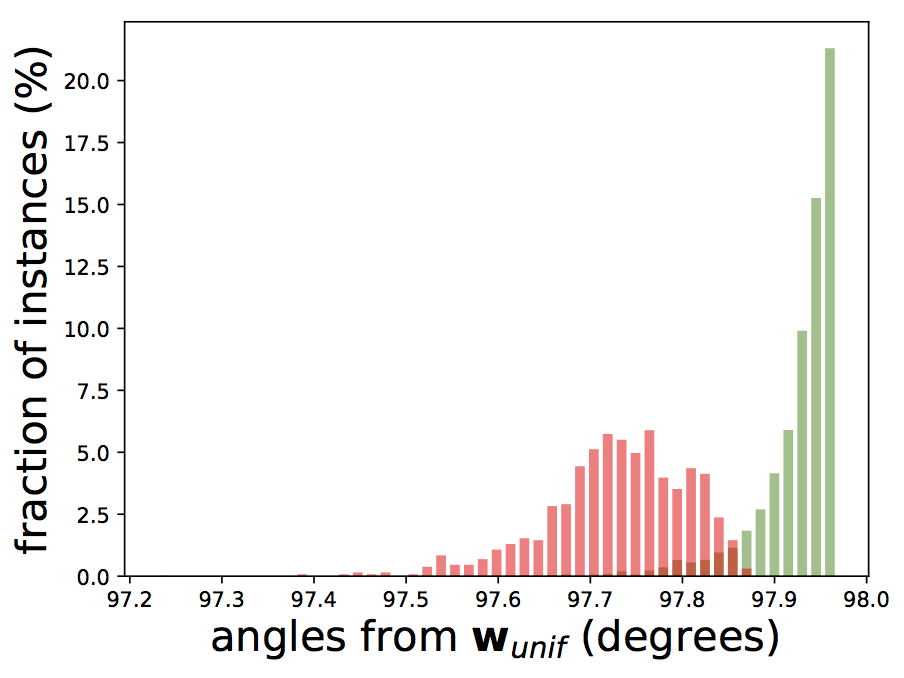}%
		\label{fig:angles_shuttle}} \\
    \subfloat[Weather]{\includegraphics[width=0.24\textwidth]{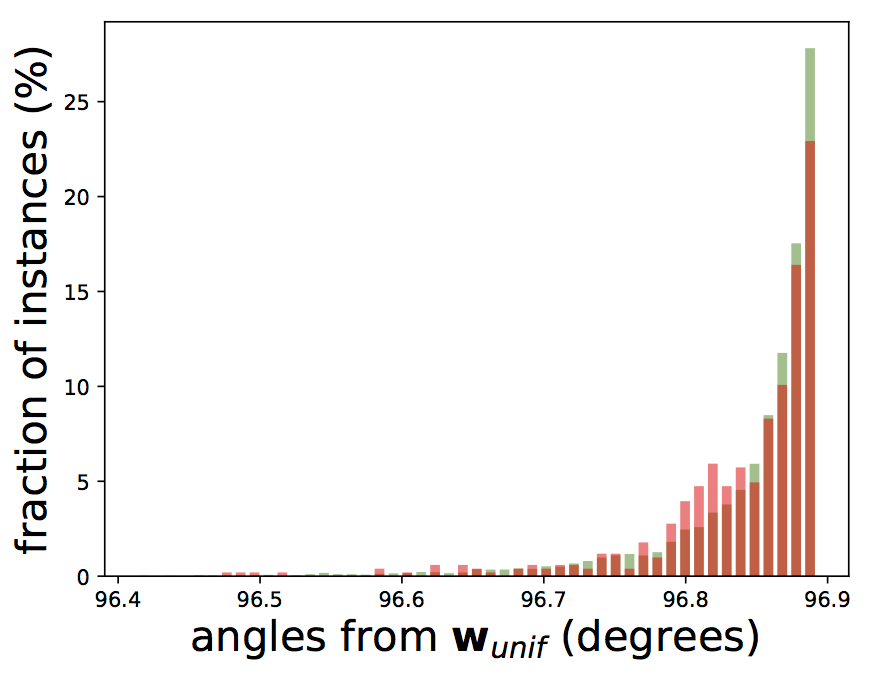}%
    	\label{fig:angles_weather}}
    \subfloat[Yeast]{\includegraphics[width=0.24\textwidth]{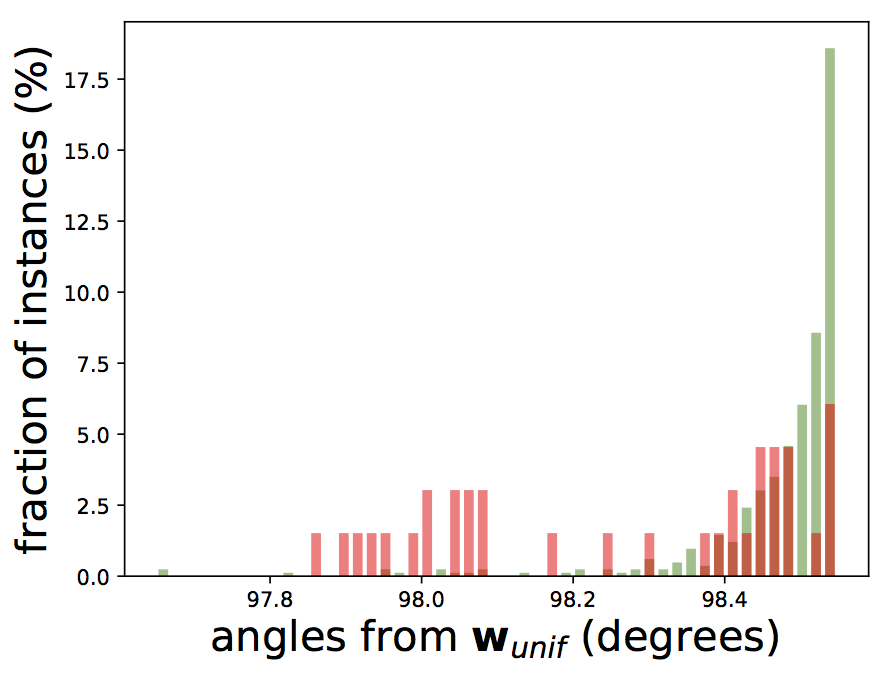}%
    	\label{fig:angles_yeast}}
	\caption{Histogram distribution of the angles between score vectors from IFOR and ${\mathbf w}_{unif}$ for all datasets. The red and green histograms show the angle distributions for anomalies and nominals respectively. Since the red histograms are closer to the left, anomalies are aligned closer to ${\mathbf w}_{unif}$.}
	\label{fig:angles}
\end{figure*}

\subsection{Human-in-the-Loop Learning for Batch Data Setting}

\noindent {\bf Experimental Setup.} All versions of HiLAD-Batch 
employ IFOR with the number of trees $T=100$ and subsample size $256$. The initial starting weights are denoted by ${\bf w}^{(0)}$. We normalize the score vector for each instance to unit length such that the score vectors lie on a unit sphere. This normalization helps adhere to the discussion in Section~\ref{sec:insights}, but is otherwise unnecessary. Figure~\ref{fig:angles} shows that ${\mathbf w}_{unif}$ tends to have a smaller angular separation from the normalized IFOR score vectors of anomalies than from those of nominals. This holds true for most of our datasets (Table~\ref{tab:datasets_full}). \textit{Weather} is a hard dataset for all anomaly detectors \cite{wu:2014}, as reflected in its angular distribution in Figure~\ref{fig:angles_weather}. In all our experiments, \texttt{Unsupervised Baseline} shows the number of anomalies detected without any feedback, i.e., using the uniform ensemble weights ${\mathbf w}_{unif}$; \texttt{HiLAD-Batch (No Prior - Unif)} and \texttt{HiLAD-Batch (No Prior - Rand)} impose no priors on the model, and start human-in-the-loop learning with ${\mathbf w}^{(0)}$ set to ${\mathbf w}_{unif}$ and a random vector respectively; \texttt{HiLAD-Batch} sets ${\mathbf w}_{unif}$ as prior, and starts with ${\mathbf w}^{(0)}={\mathbf w}_{unif}$. For HST, we present two sets of results with batch input only: \texttt{HST-Batch} with original settings ($T=25$, depth=$15$, no feedback) \cite{tan:2011}, and \texttt{HST-Batch (Feedback)} which supports feedback with HiLAD-Batch strategy (with $T=50$ and depth=$8$, a better setting for feedback). For RST, we present the results (\texttt{RST-Batch}) with only the original settings ($T=30$, depth=$15$) \cite{wu:2014} since it was not competitive with other methods on our datasets. We also compare the HiLAD-Batch variants with the AAD approach \cite{das:2016} in the batch setting (\texttt{AAD-Batch}). Batch data setting is the most optimistic for all algorithms.

\subsubsection{Results for HiLAD-Batch Instantiation}

\begin{figure}[h]
	\centering
	\subfloat[Abalone]{\includegraphics[width=0.48\linewidth]{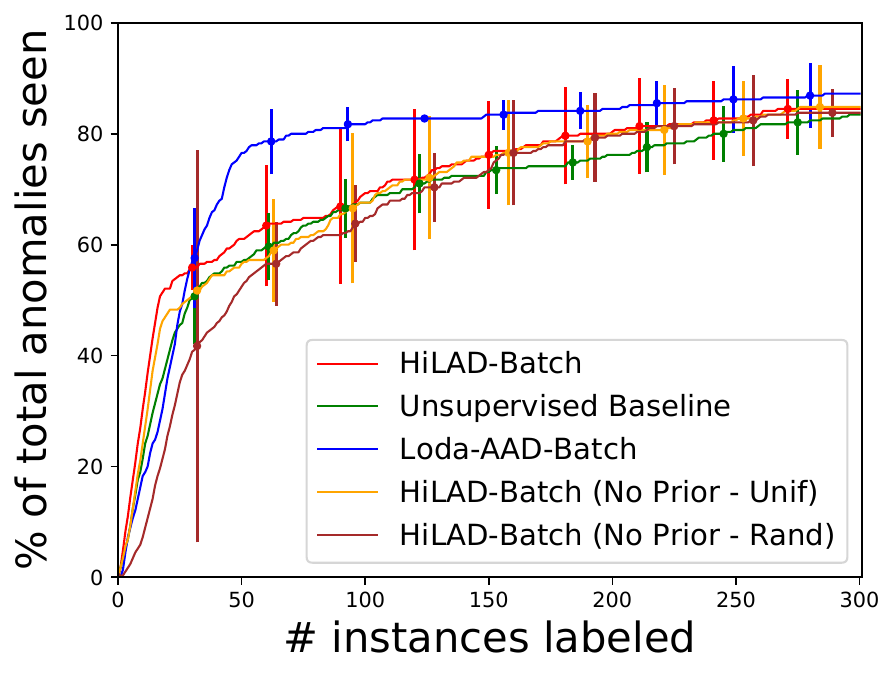}%
		\label{fig:batch_abalone}}
	\subfloat[ANN-Thyroid-1v3]{\includegraphics[width=0.48\linewidth]{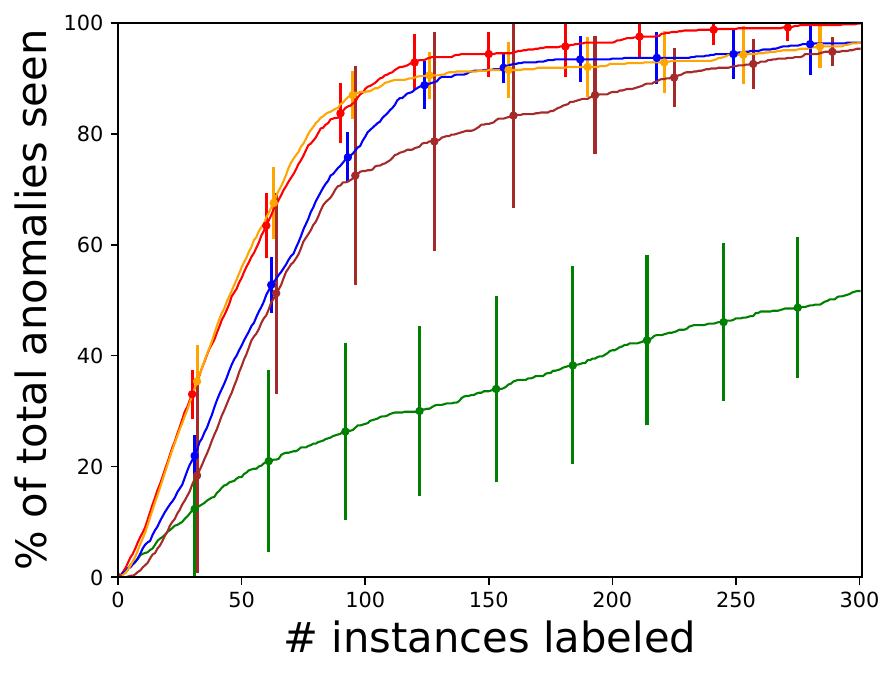}%
		\label{fig:batch_ann_thyroid_1v3}} \\
	\subfloat[Cardiotocography]{\includegraphics[width=0.48\linewidth]{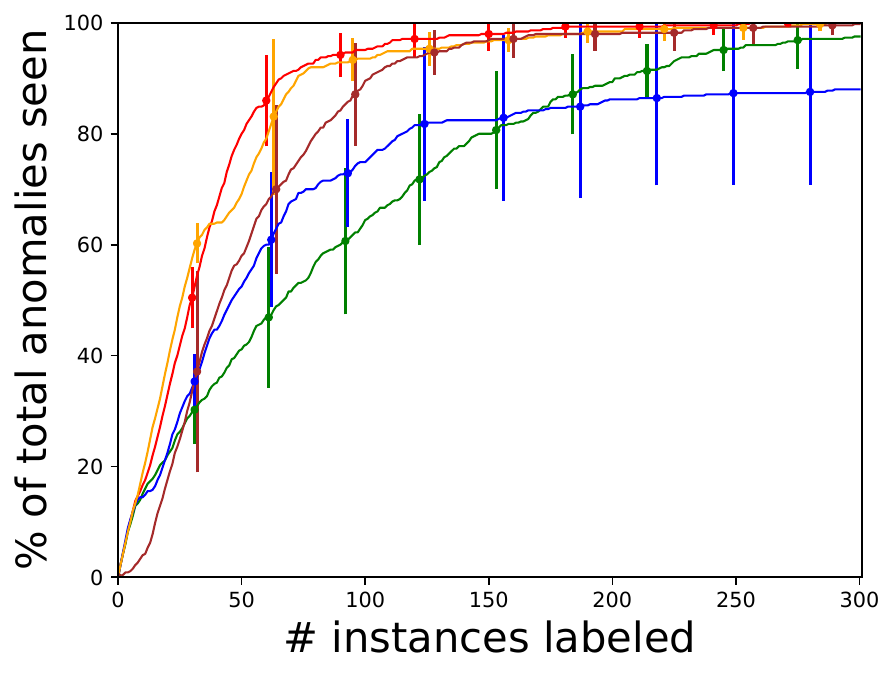}%
		\label{fig:batch_cardiotocography}}
	\subfloat[Yeast]{\includegraphics[width=0.48\linewidth]{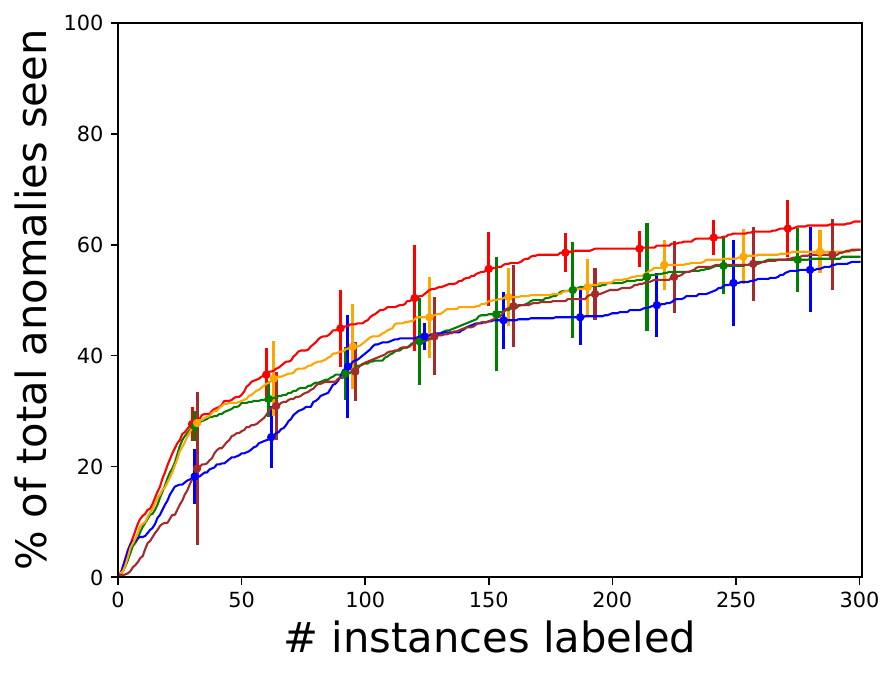}%
		\label{fig:batch_yeast}}
	\caption{Percentage of total anomalies seen vs. the number of queries for the \textbf{smaller} datasets in the \textbf{batch} setting.}
	\label{fig:batch}
\end{figure}
We set the budget $B$ to $300$ for all datasets in the batch setting. The results on the four smaller datasets \textit{Abalone, ANN-Thyroid-1v3, Cardiotocography,} and \textit{Yeast} are shown in Figure~\ref{fig:batch}. 
When the algorithm starts from sub-optimal initialization of the weights and with no prior knowledge (\texttt{HiLAD-Batch (No Prior - Rand)}), more number of queries are spent hunting for the first few anomalies, and thereafter detection improves significantly. When the weights are initialized to ${\mathbf w}_{unif}$, which is a reliable starting point (\texttt{HiLAD-Batch (No Prior - Unif)} and \texttt{HiLAD-Batch}), fewer queries are required to find the initial anomalies, and typically results in a lower variance in accuracy. Setting ${\mathbf w}_{unif}$ as prior in addition to informed initialization (\texttt{HiLAD-Batch}) performs better than without the prior (\texttt{HiLAD-Batch (No Prior - Unif)}) on \textit{Abalone, ANN-Thyroid-1v3}, and \textit{Yeast}. We believe this is because the prior helps guard against noise.

\begin{figure}[h]
	\centering
	\subfloat[Abalone]{\includegraphics[width=0.48\linewidth]{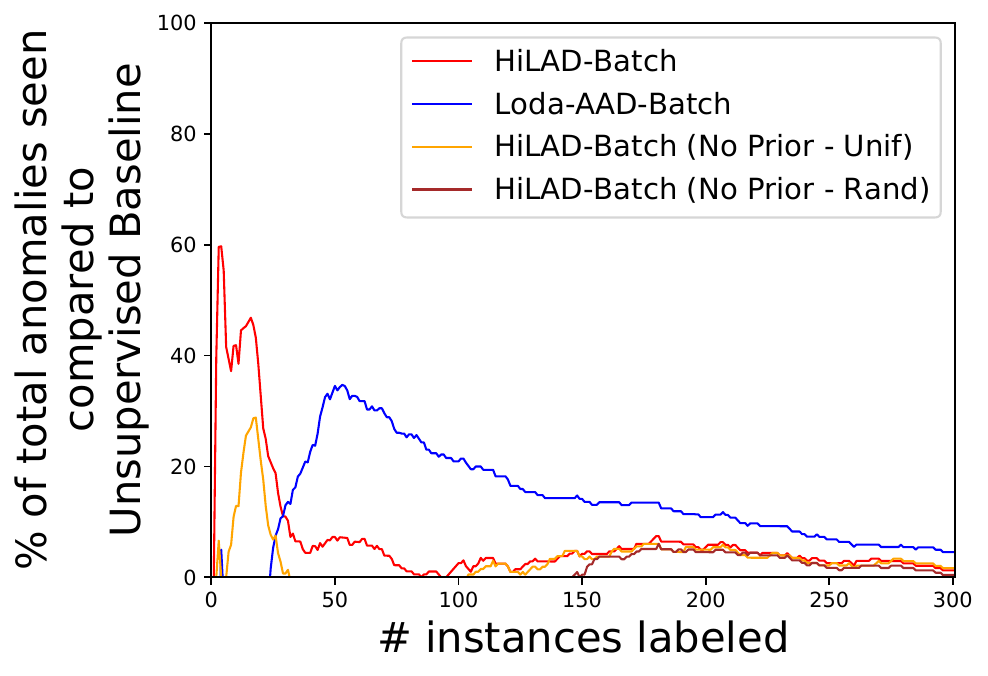}%
		\label{fig:batch_abalone}}
	\subfloat[ANN-Thyroid-1v3]{\includegraphics[width=0.48\linewidth]{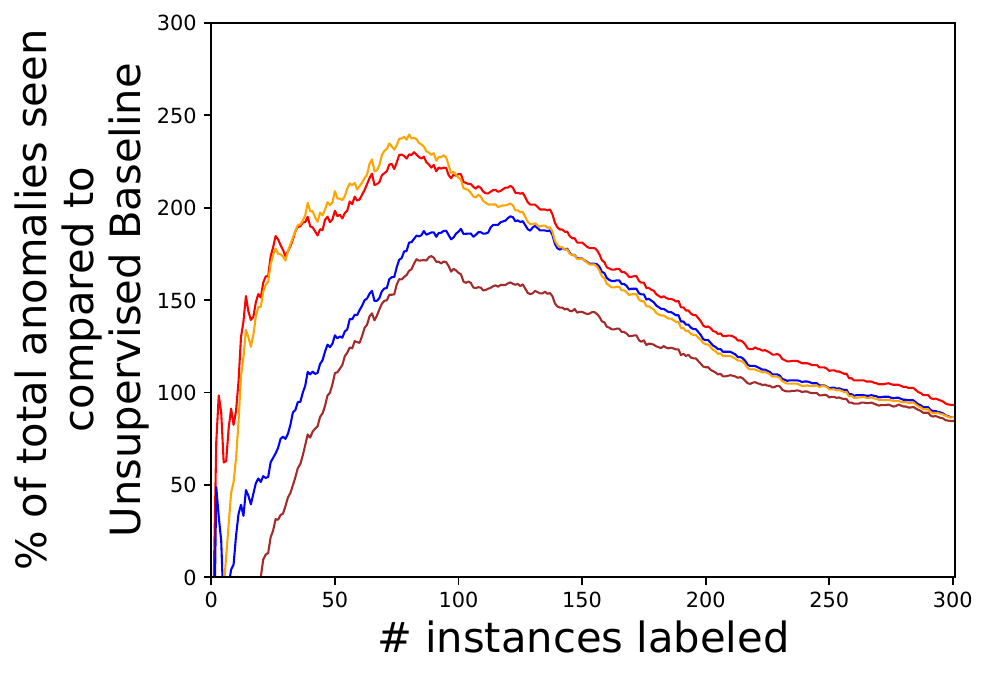}%
		\label{fig:batch_ann_thyroid_1v3}} \\
	\subfloat[Cardiotocography]{\includegraphics[width=0.48\linewidth]{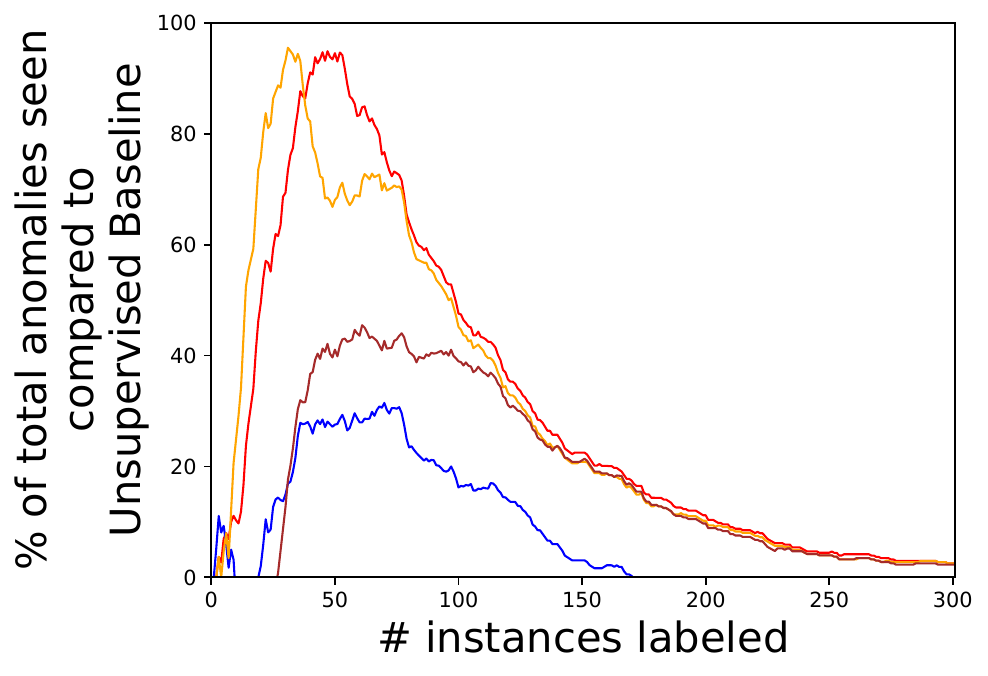}%
		\label{fig:batch_cardiotocography}}
	\subfloat[Yeast]{\includegraphics[width=0.48\linewidth]{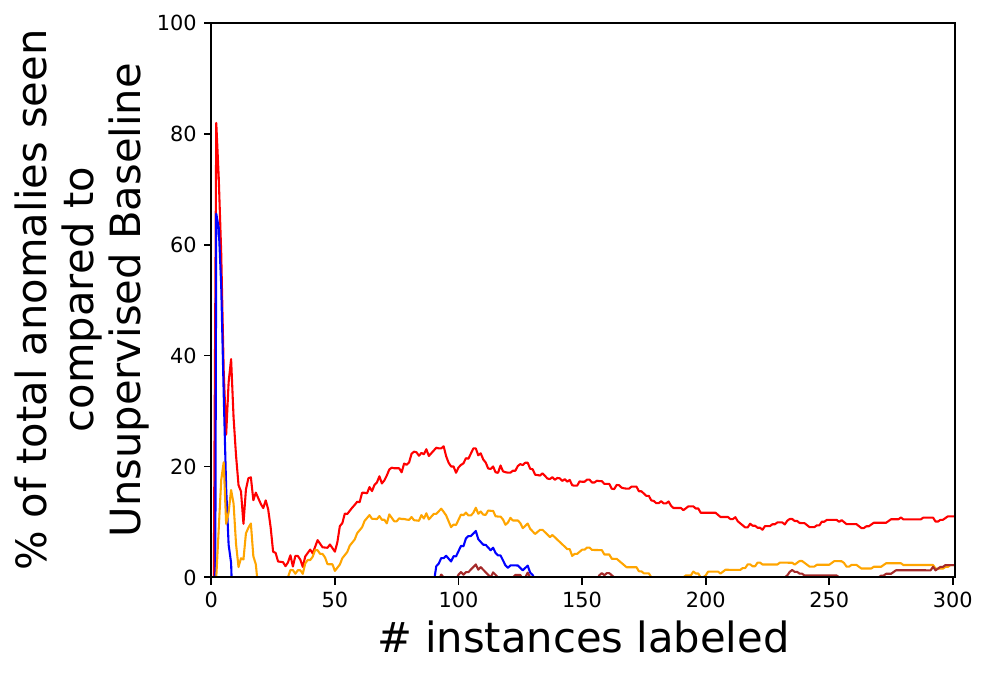}%
		\label{fig:batch_yeast}}
	\caption{Relative \% of anomalies seen compared to the \textit{Unsupervised Baseline} for \textbf{smaller} datasets in the \textbf{batch} setting.}
	\label{fig:batch-relative-change}
\end{figure}

Figure \ref{fig:batch-relative-change} shows the relative performance (\% of anomalies discovered) compared to the \texttt{Unsupervised Baseline}. It is clear that \texttt{HiLAD-Batch} discovered up to 250\% more anomalies than the baseline algorithm. A key observation to note is that \texttt{HiLAD-Batch} discovers the biggest portion of anomalies in the early phase of the label collection. For \textit{Abalone} dataset \texttt{HiLAD-Batch} discovered 40-60\% more anomalies in the initial phase whereas \texttt{Loda-AAD-Batch} was under-performing initially and recovered later. The biggest gain we saw was for \textit{Thyroid} dataset where \texttt{HiLAD-Batch} discovered $250\%$ more anomalies compared to the baseline. Other datasets (\textit{Cardiotocography, Yeast}) also resulted in higher anomaly discovery using the \texttt{HiLAD-Batch} method.

\subsubsection{Results for Diversified Query Strategy}

The diversified querying strategy \texttt{Select-Diverse} (Algorithm~\ref{alg:diverse}) employs compact descriptions to select instances. Therefore, the evaluation of its effectiveness is presented first. An interactive system can potentially ease the cognitive burden on the analysts by using descriptions to generate a ``summary'' of the anomalous instances.
\begin{figure}[ht]
	\centering
	\subfloat[Class diversity]{\includegraphics[width=0.48\textwidth]{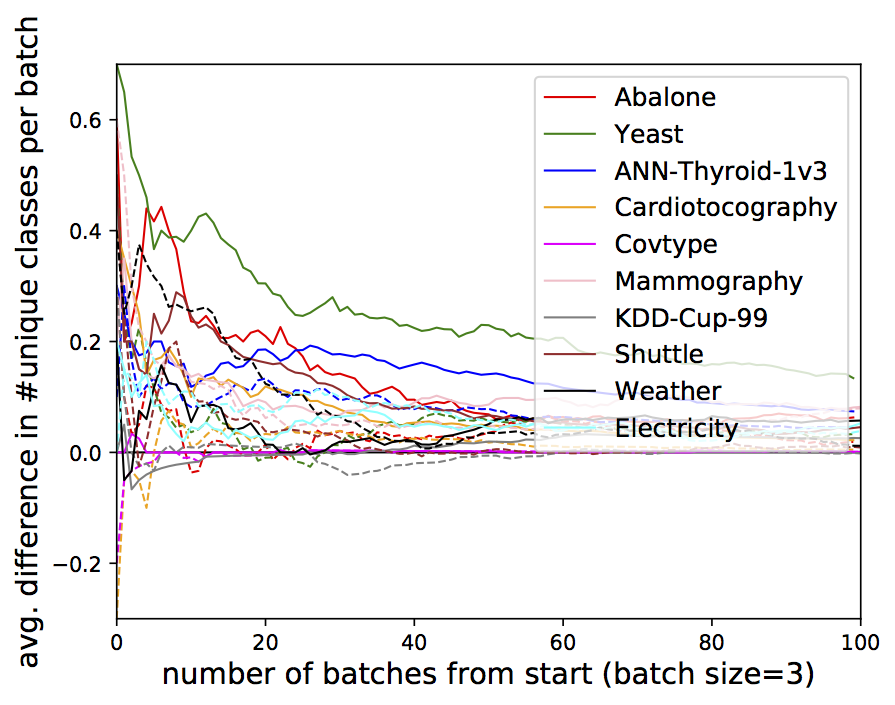}
		\label{fig:class_diversity}}
	\subfloat[Discovery performance comparison between \texttt{HiLAD-Batch-D} and \texttt{HiLAD-Batch} ]{\includegraphics[width=0.48\textwidth]{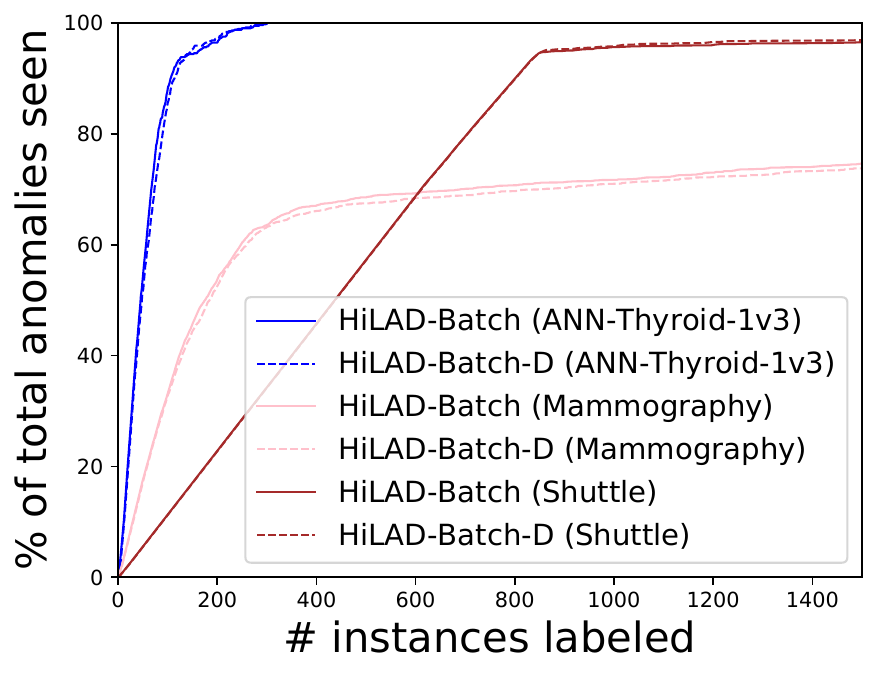}
		\label{fig:diversity_num_seen}}
	\caption{Results comparing diversified querying strategy \texttt{HiLAD-Batch-D} with baseline query strategies \texttt{HiLAD-Batch-T} and \texttt{HiLAD-Batch-R}. The x-axis in {\bf (a)} shows the number of query batches (of batch size 3). The y-axis shows the difference in the number of unique classes seen averaged across all batches till a particular batch. The solid lines in {\bf (a)} show the average difference between unique classes seen with \texttt{HiLAD-Batch-D} and \texttt{HiLAD-Batch-T}; the dashed lines show the average difference between \texttt{HiLAD-Batch-D} and \texttt{HiLAD-Batch-R}. \textbf{(b)} presents the discovery performance for \texttt{HiLAD-Batch-D} (dashed line). And the baseline is \texttt{HiLAD-Batch} (solid line)}
	\label{fig:query_diversity_all}
\end{figure}

We perform a {\em post hoc} analysis on the datasets with the knowledge of the original classes (Table~\ref{tab:datasets_full}). It is assumed that each class in a dataset represents a different data-generating process. To measure the diversity at any point in our feedback cycle, we compute the difference between the number of unique classes presented to the analyst per query batch averaged across all the past batches. The parameter $\delta$ for \texttt{Select-Diverse} was set to $5$ in all experiments. We compare three query strategies in the batch data setup: \texttt{HiLAD-Batch-T}, \texttt{HiLAD-Batch-D}, and \texttt{HiLAD-Batch-R}. \texttt{HiLAD-Batch-T} simply presents the top three most anomalous instances per query batch. \texttt{HiLAD-Batch-D} employs \texttt{Select-Diverse} to present three diverse instances out of the ten most anomalous instances. \texttt{HiLAD-Batch-R} presents three instances selected at random from the top ten anomalous instances. Finally, \texttt{HiLAD-Batch} greedily presents only the single most anomalous instance for labeling. We find that \texttt{HiLAD-Batch-D} presents a more diverse set of instances than both \texttt{HiLAD-Batch-T} (solid lines) as well as \texttt{HiLAD-Batch-R} (dashed lines) on most datasets. Figure~\ref{fig:diversity_num_seen} shows that the number of anomalies discovered (on representative datasets) with the diversified querying strategy is similar to the greedy strategy, i.e., {\em no loss in discovery rate to improve diversity}.

\subsection{Human-in-the-Loop Learning for Streaming Data Setting}
\noindent {\bf Experimental Setup.} For HiLAD-Stream instantiation, we employ IFOR with the number of trees $T=100$ and subsample size $256$. This is similar to the setup for HiLAD-Batch. In all HiLAD-Stream experiments, we set the number of queries per window $Q=20$. The total budget $B$ and the stream window size $K$ for the datasets were set respectively as follows: \textit{Covtype} (3000, 4096), \textit{KDD-Cup-99} (3000, 4096), \textit{Mammography} (1500, 4096), \textit{Shuttle} (1500, 4096), \textit{Electricity} (1500, 1024), \textit{Weather} (1000, 1024). These values are reasonable w.r.t the dataset's size, the number of anomalies, and the rate of concept drift. The maximum number of unlabeled instances residing in memory is $K$. When the last window of data arrives, then human-in-the-loop learning is continued with the final set of unlabeled data retained in the memory until the total budget $B$ is exhausted. The instances are streamed in the same order as they appear in the original public sources. When a new window of data arrives: \texttt{HiLAD-Stream~(KL~Adaptive)} dynamically determines which trees to replace based on KL-divergence, \texttt{HiLAD-Stream~(Replace $20\%$ Trees)} replaces $20\%$ oldest trees, and \texttt{HiLAD-Stream~(No~Tree~Replace)} creates the trees only once with the first window of data and only updates the weights of the fixed leaf nodes using feedback. 
\begin{figure}[h]
	\centering
	\subfloat[Covtype]{\includegraphics[width=0.45\linewidth]{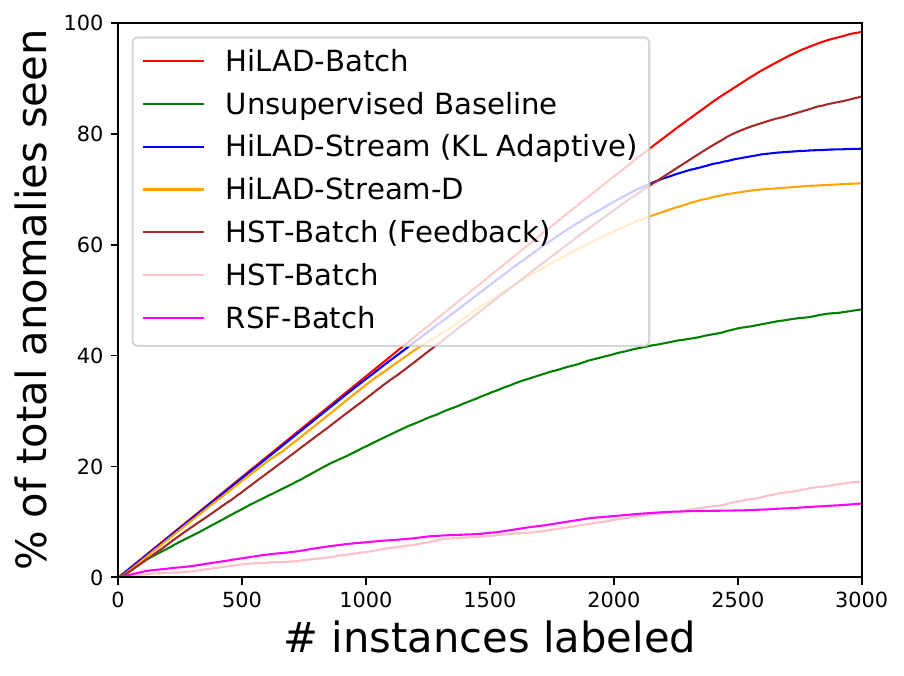}%
		\label{fig:stream_covtype}}
	\subfloat[Mammography]{\includegraphics[width=0.45\linewidth]{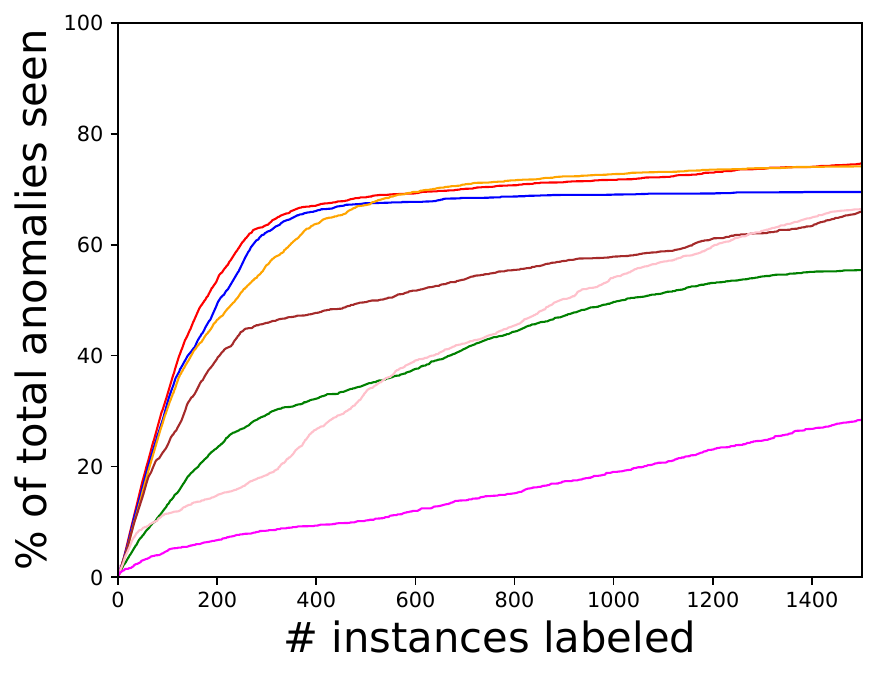}%
		\label{fig:stream_mammography}} \\
	\subfloat[KDD-Cup-99]{\includegraphics[width=0.45\linewidth]{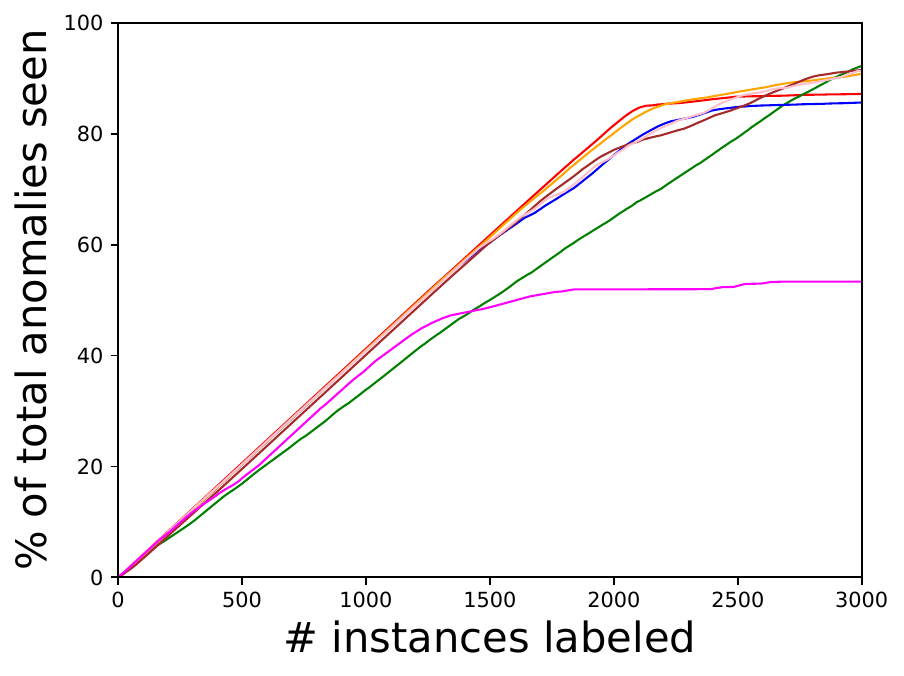}%
	\label{fig:stream_kddcup}}
	\subfloat[Shuttle]{\includegraphics[width=0.45\linewidth]{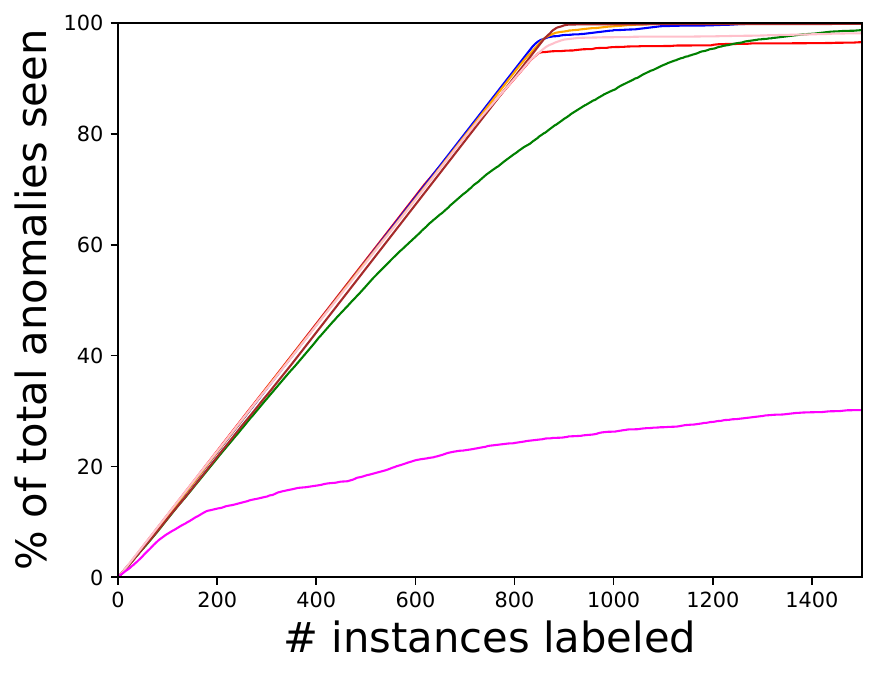}%
	\label{fig:stream_shuttle}}
	\caption{Percentage of total anomalies seen vs. number of queries for the \textbf{larger datasets in the limited memory setting}. \texttt{HiLAD-Stream~(KL~Adaptive)} and \texttt{HiLAD-Stream-D} apply the \texttt{Select-Top} and \texttt{Select-Diverse} query strategies resp. \textit{Mammography}, \textit{KDD-Cup-99}, and \textit{Shuttle} have no significant drift. \textit{Covtype}, which has a higher drift, is included here for comparison because it is large.}
	\label{fig:limited_memory}
\end{figure}

\subsubsection{Results for HiLAD-Stream Instantiation}

\noindent\textbf{Limited memory setting with no concept drift.} The results on the four larger datasets are shown in Figure~\ref{fig:limited_memory}. The performance is similar to what is seen on the smaller datasets. Among the unsupervised algorithms in the batch setting, IFOR (\texttt{Unsupervised Baseline}) and HST (\texttt{HST-Batch}) are competitive, and both are better than RSF (\texttt{RSF-Batch}). With feedback, \texttt{HiLAD-Batch} is consistently the best performer. HST with feedback (\texttt{HST-Batch (Feedback)}) always performs better than \texttt{HST-Batch}. The streaming algorithm with feedback, \texttt{HiLAD-Stream (KL~Adaptive)}, significantly outperforms \texttt{Unsupervised Baseline} and is competitive with \texttt{HiLAD-Batch}. \texttt{HiLAD-Stream (KL~Adaptive)} performs better than \texttt{HST-Batch (Feedback)} as well. \texttt{HiLAD-Stream-D} which presents a more diverse set of instances for labeling performs similar to \texttt{HiLAD-Stream (KL Adaptive)}. These results demonstrate that the {\em feedback-tuned anomaly detectors generalize to unseen data}.

Figure~\ref{fig:drift_detection_all_non_streaming} shows drift detection on datasets which do not have any drift. The streaming window size for each dataset was set commensurate to its total size: \textit{Abalone}($512$), \textit{ANN-Thyroid-1v3}($512$), \textit{Cardiotocography}($512$), \textit{Covtype}($4096$), \textit{Electricity}($1024$), \textit{KDDCup99}($4096$), \textit{Mammography}($4096$), \textit{Shuttle}($4096$), \textit{Weather}($1024$), and \textit{Yeast}($512$).

\begin{figure}[h]
	\centering
	\subfloat[ANN-Thyroid-1v3]{
		\includegraphics[width=0.47\textwidth]{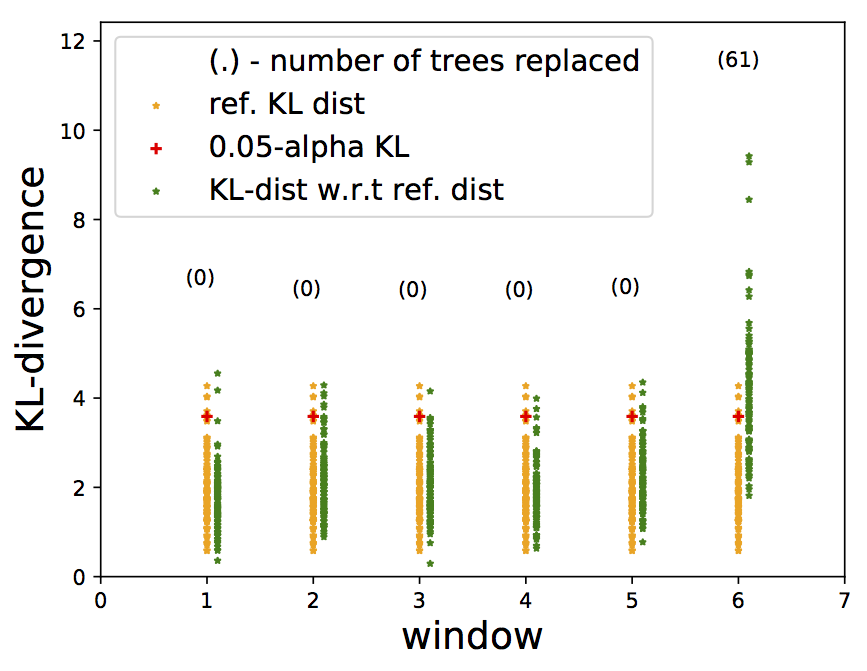}
		\label{fig:drift_ann_thyroid_1v3}}
	\subfloat[Covtype]{
		\includegraphics[width=0.47\textwidth]{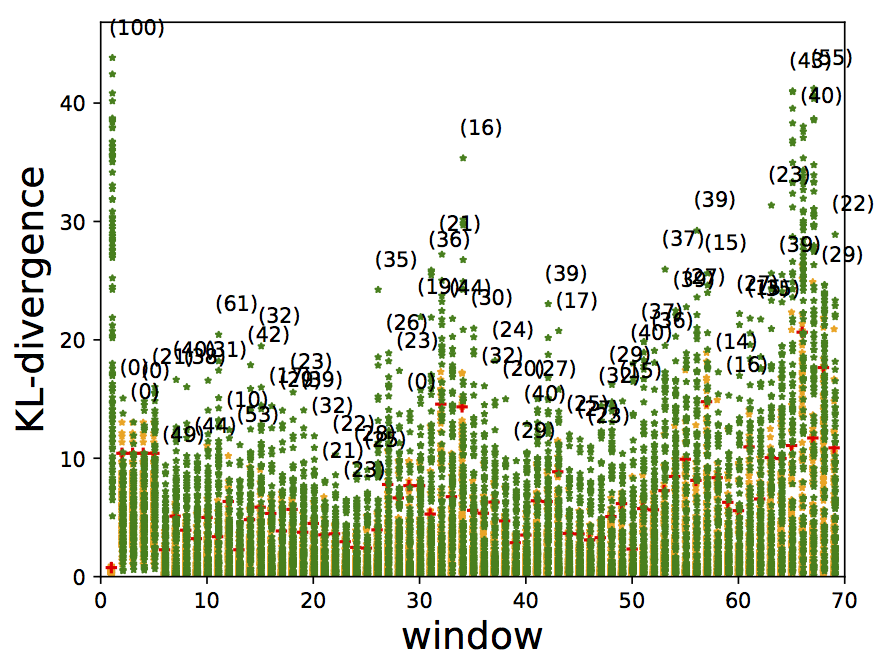}
		\label{fig:drift_covtype}}
	\caption{Results for drift detection across windows. {\bf (a)} When there is no drift, such as in {\em ANN-Thyroid-1v3}, then no trees are replaced for most of the windows, and the older model is retained. {\bf (b)} If there is drift, such as in {\em Covtype}, then the trees are more likely to be replaced.}
	\label{fig:drift}
\end{figure}

\begin{figure}
	\centering 
	\subfloat[Abalone]{
		\includegraphics[width=0.45\textwidth]{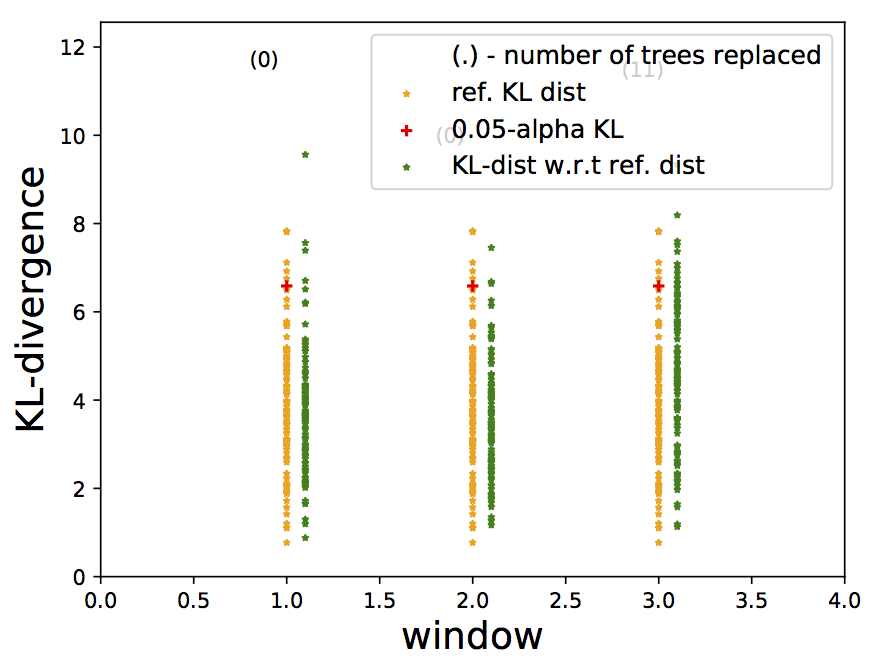}
		\label{fig:concept_drift_abalone}}
	\subfloat[Cardiotocography]{
		\includegraphics[width=0.45\textwidth]{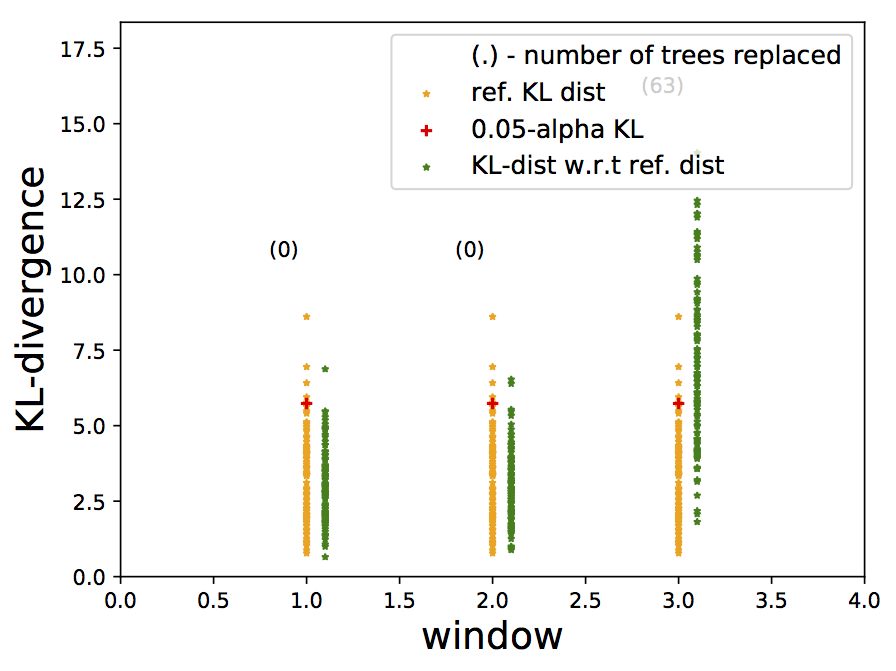}
		\label{fig:concept_drift_cardiotocography_1}} \\
	\subfloat[Yeast]{
		\includegraphics[width=0.45\textwidth]{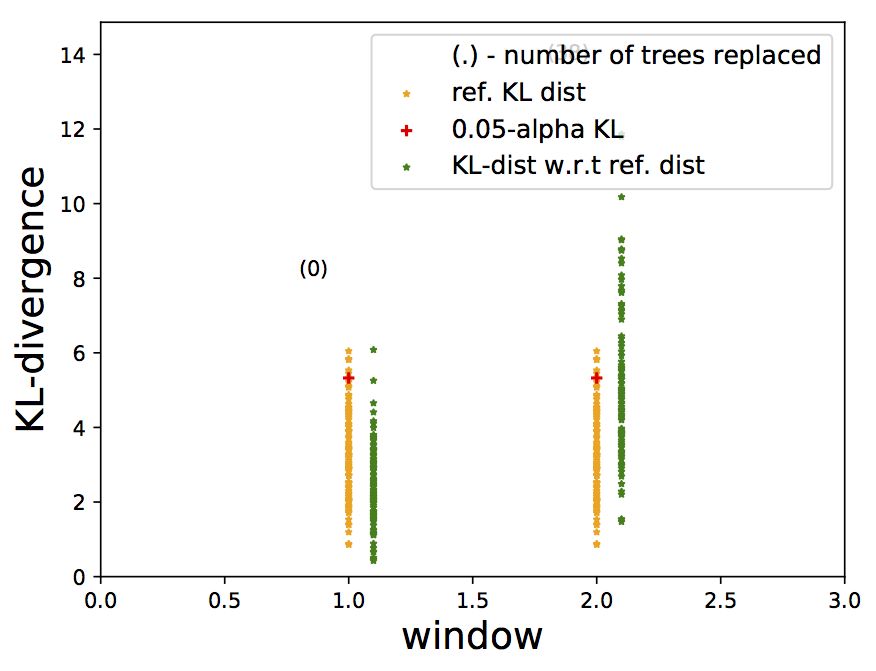}
		\label{fig:concept_drift_yeast}} 
	\subfloat[Mammography]{
		\includegraphics[width=0.45\textwidth]{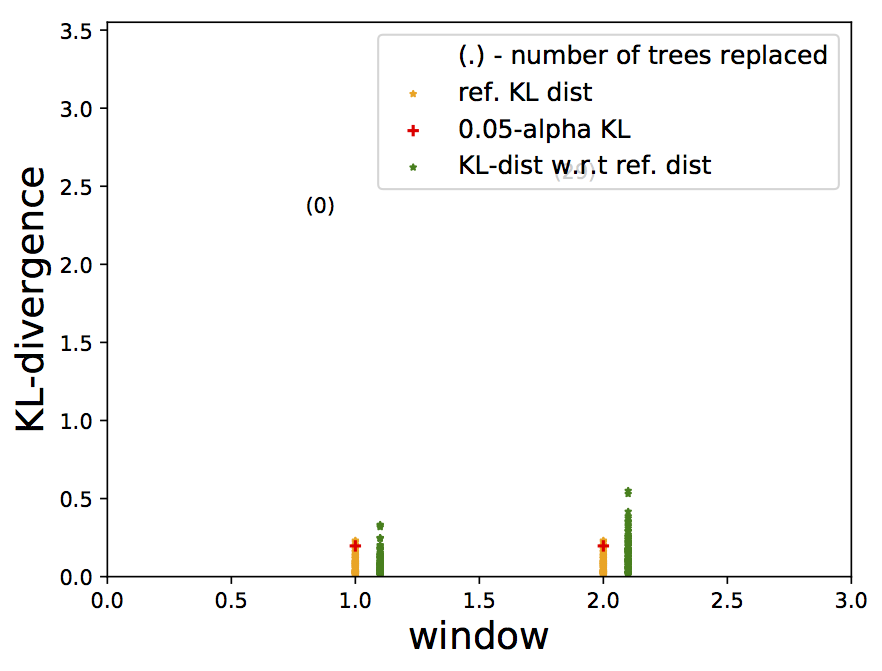}
		\label{fig:fig:concept_drift_mammography}} \\
	\subfloat[KDD-Cup-99]{
		\includegraphics[width=0.45\textwidth]{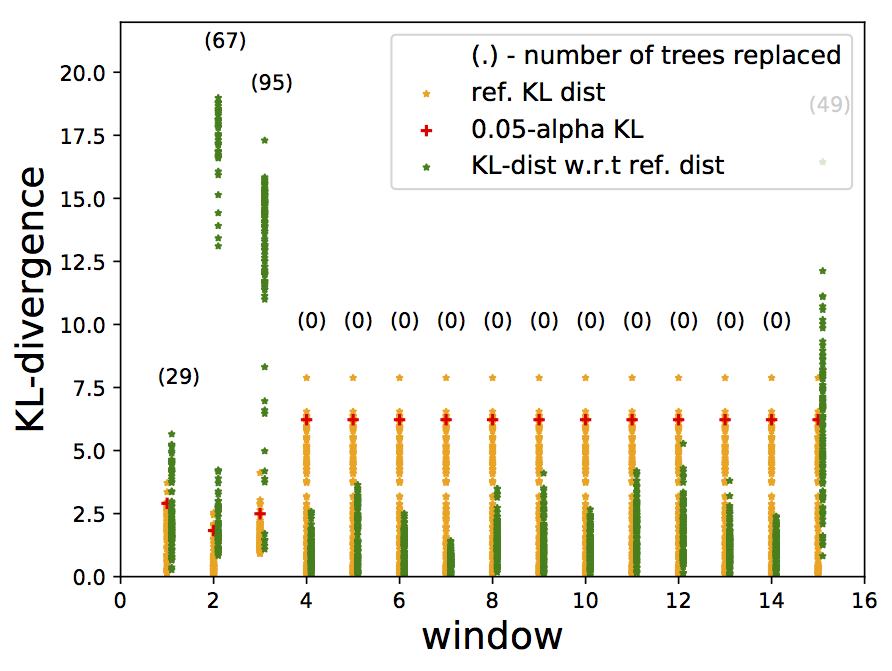}
		\label{fig:fig:concept_drift_kddcup}} 
	\subfloat[Shuttle]{
		\includegraphics[width=0.45\textwidth]{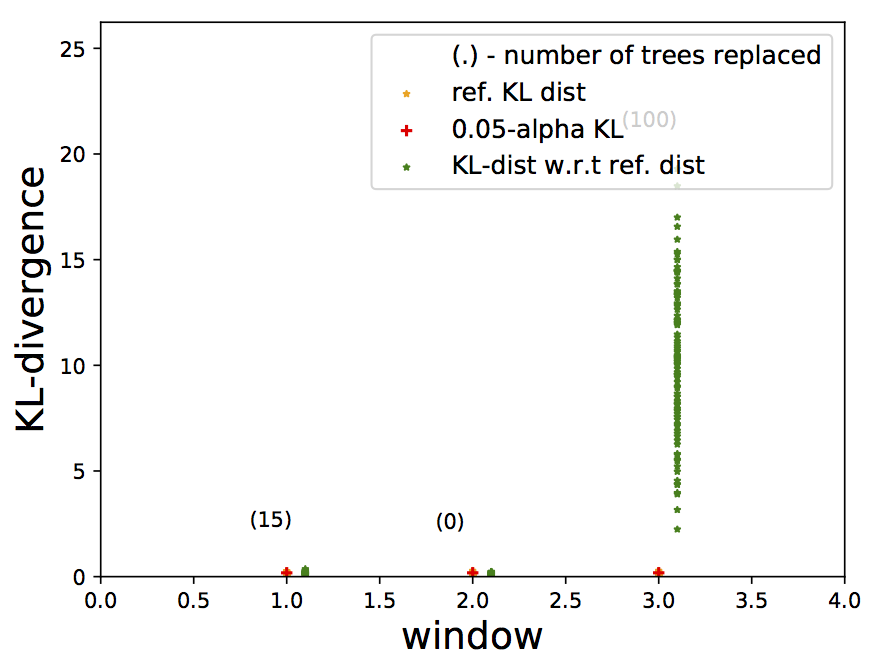}
		\label{fig:fig:concept_drift_shuttle_1v23567}}
	\caption{The last data window in each dataset usually has much fewer instances and therefore, its distribution is very different from the previous window despite there being no data drift. \textbf{Therefore, we ignore the drift in the last window.} We did not expect \textit{Abalone}, \textit{Cardiotocography}, \textit{KDDCup99}, \textit{Mammography}, \textit{Shuttle}, \textit{Yeast}, and \textit{ANN-Thyroid-1v3} (Figure~\ref{fig:drift_ann_thyroid_1v3}) to have much drift in the data. This can also be seen in the plots where most of the windows in the middle of streaming did not result in too many trees being replaced (the numbers in the parenthesis are mostly zero).}
	\label{fig:drift_detection_all_non_streaming}
\end{figure}

\vspace{1.0ex}

\begin{figure}
	\centering 
	\subfloat[Covtype]{
		\includegraphics[width=0.3\textwidth]{stream_diff/test_concept_drift_covtype}
		\label{fig:concept_drift_covtype}}
	\subfloat[Electricity]{
		\includegraphics[width=0.3\textwidth]{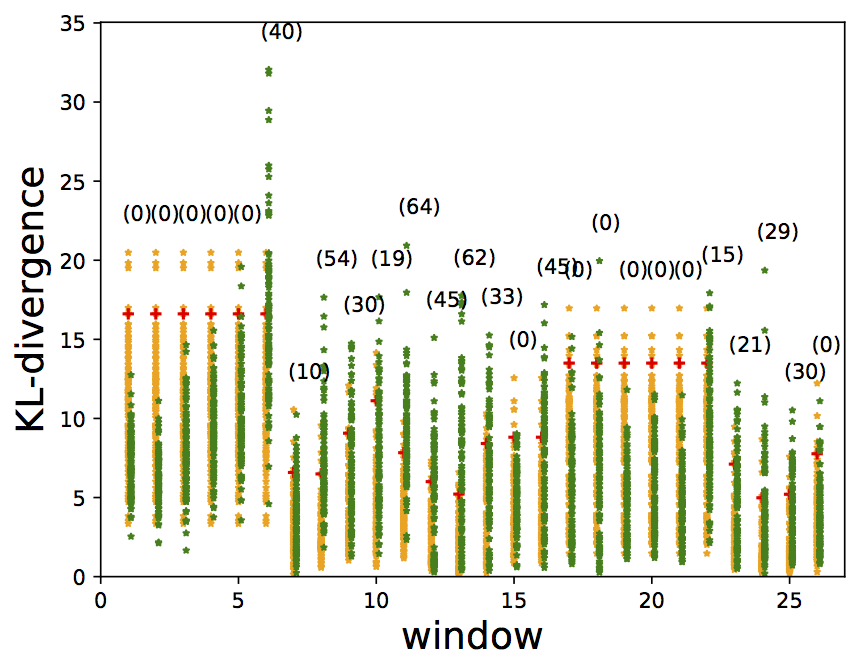}
		\label{fig:concept_drift_electricity}}
	\subfloat[Weather]{
		\includegraphics[width=0.3\textwidth]{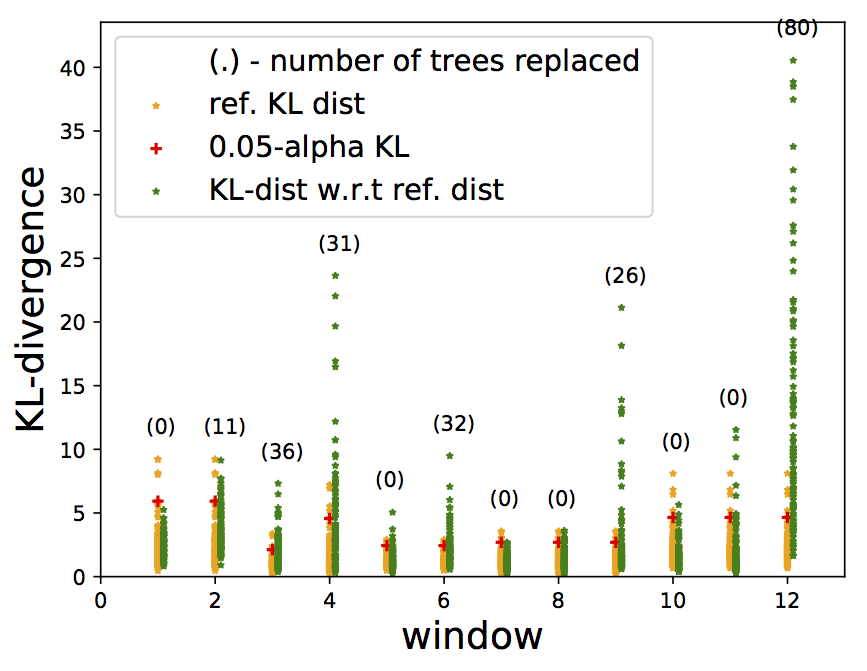}
		\label{fig:concept_drift_weather}} \\
	\subfloat[Covtype]{
		\includegraphics[width=0.3\textwidth]{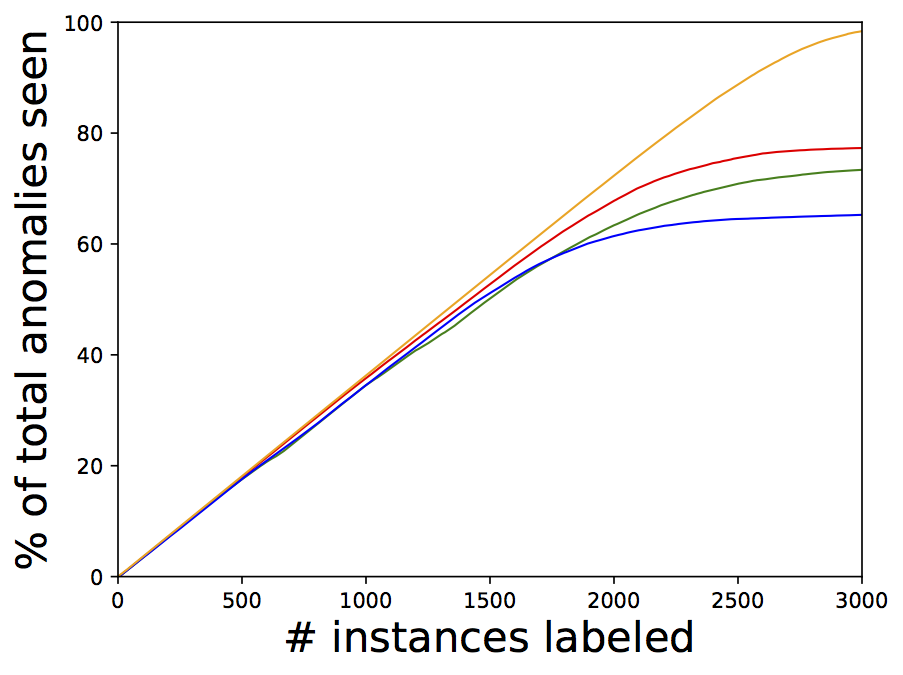}
		\label{fig:concept_drift_covtype_num}}
	\subfloat[Electricity]{
		\includegraphics[width=0.3\textwidth]{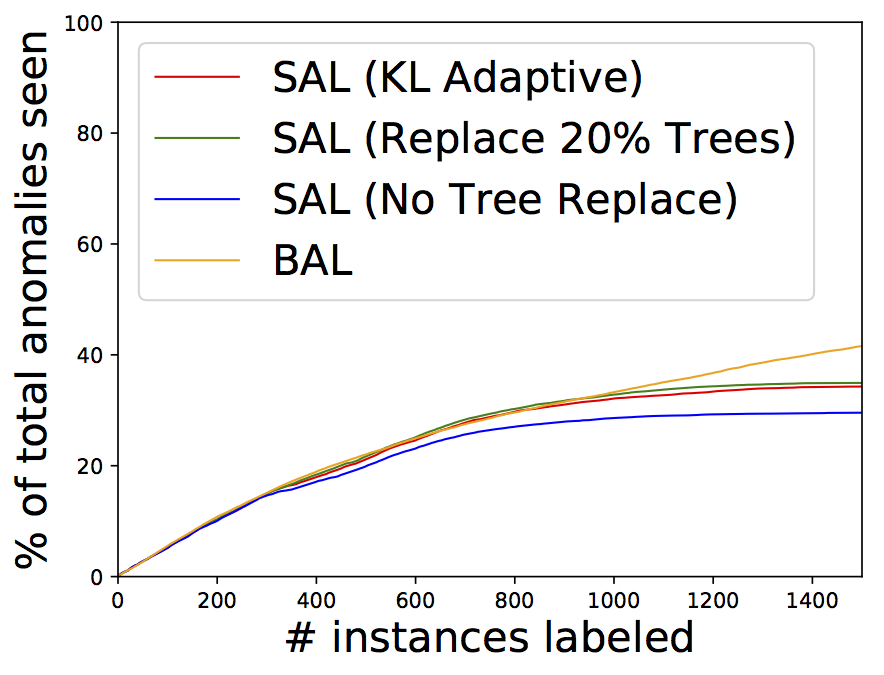}
		\label{fig:concept_drift_electricity_num}}
	\subfloat[Weather]{
		\includegraphics[width=0.3\textwidth]{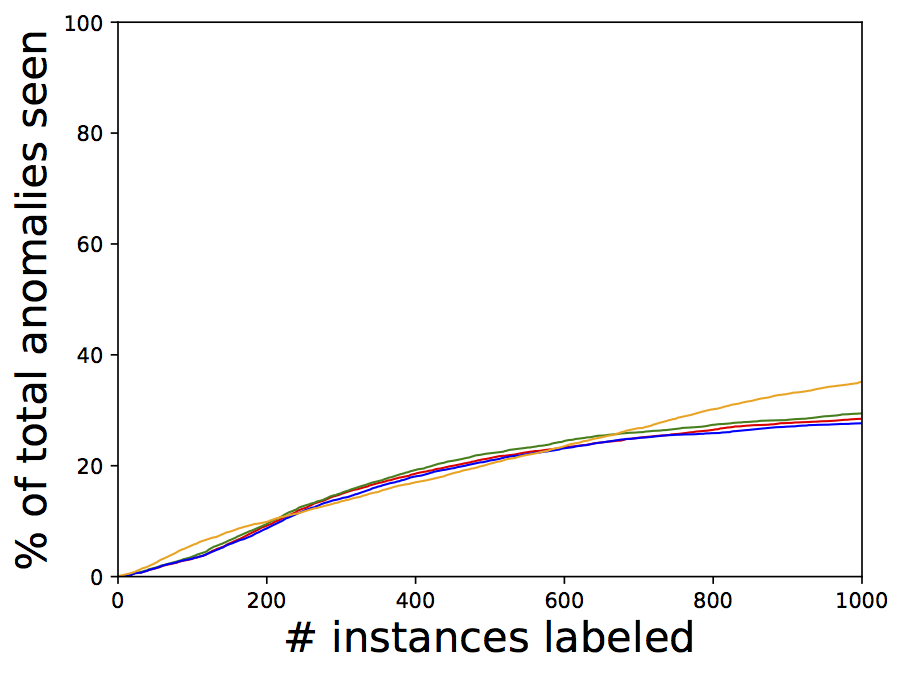}
		\label{fig:concept_drift_weather_num}}
	\caption{Results for integrated drift detection and label feedback with HiLAD-Stream algorithm. The top row shows the number of trees replaced per window when a drift in the data was detected relative to previous window(s). The bottom row shows the percentage of total anomalies seen vs. number of queries for the \textbf{streaming datasets with significant concept drift}.}
	\label{fig:drift_detection_all_streaming}
\end{figure}

\noindent\textbf{Streaming setting with concept drift.} Figure~\ref{fig:drift_detection_all_streaming} shows the results after integrating drift detection and label feedback with HiLAD-Stream for the datasets which are expected to have significant drift. Both \textit{Covtype} and \textit{Electricity} show more drift in the data than \textit{Weather} (top row in Figure~\ref{fig:drift_detection_all_streaming}). The rate of drift determines how fast the model should be updated. If we update the model too fast, then we loose valuable knowledge gained from feedback which is still valid. On the other hand, if we update the model too slowly, then the model continues to focus on the stale subspaces based on the past feedback. It is hard to find a common rate of update that works well across all datasets (such as replacing $20\%$ trees with each new window of data). Figure~\ref{fig:drift_detection_all_streaming} (bottom row) shows that the adaptive strategy (\texttt{HiLAD-Stream~(KL~Adaptive)}) which replaces obsolete trees using KL-divergence, as illustrated in Figure~\ref{fig:drift_detection_all_streaming} (top row), is robust and competitive with the best possible configuration, i.e., \texttt{HiLAD-Batch}.

\subsubsection{Parameter sensitivity analysis for \texttt{HiLAD-Stream}}

\label{subsec:streaming_param}

HiLAD-Stream depends on drift detection measure. It has several approaches to replace ensemble members: i) KL-adaptive, ii) a fixed amount of tree replacement, and iii) no replacement. \textit{KL-adaptive} case, we focused on selection of 95 percentile threshold to discard less important ensemble members(trees). We changed the decision threshold to observe \texttt{HiLAD-Stream's} sensitivity. Similar exploration was performed for fixed tree replacement parameters from a wide range of parameters 5\%, 10\%, 15\%, 20\% 25\%. Table \ref{tab:streaming_sensitivity} presents our algorithms detection performance with the variation of parameters. KL based threshold selection (95\%) is the most stable one. In a real-world scenario, determining the fixed replacement factors is infeasible, and KL-divergence based threshold approach will be the recommended choice. We present support for this claim in Tables \ref{tab:param_sensitivity_covtype}, \ref{tab:param_sensitivity_electricty}, \ref{tab:param_sensitivity_weather} for various number of feedback. Our recommended choice seems to discover comparable amount of anomalies compared to fixed settings. We highlighted the performance of our algorithm bold and underlined the best result achieved with fixed drift amount. In a practical scenario, the amount of drift is unknown hence setting the right configuration is always a challenging decision. Whereas, our principled approach \texttt{HiLAD-Stream} determines such thresholds dynamically and can reduce the burden from human experts on setting the fixed parameter.

\begin{table}[]
	\centering
\caption{The fraction of total anomalies seen for all three datasets for streaming experiment. We explored our algorithms tree replacement fraction using both (fixed replacement and KL-threshold) approaches. It seems that KL-Divergence based technique with 95\% showed stable performance over fixed tree replacement (marked as \textbf{bold}). We also \underline{underlined} the best results obtained from all datasets.}
\label{tab:streaming_sensitivity}
\begin{tabular}{clrrr}
\cline{3-5}
\multicolumn{1}{l}{}   &        & \multicolumn{3}{c}{Datasets (\# of feedbacks)}   \\ \hline
\begin{tabular}[c]{@{}c@{}}Tree replacement\\ technique\end{tabular} &
  Amount &
  \multicolumn{1}{c}{\begin{tabular}[c]{@{}c@{}}Electricity\\ (1500)\end{tabular}} &
  \multicolumn{1}{c}{\begin{tabular}[c]{@{}c@{}}Weather\\ (1000)\end{tabular}} &
  \multicolumn{1}{c}{\begin{tabular}[c]{@{}c@{}}Covtype\\ (3000)\end{tabular}} \\ \hline
\multirow{8}{*}{Fixed} & 5\%    & 33.32          & 26.28          & 72.26          \\ 
                       & 10\%   & 34.29          & 27.91          & 72.59          \\ 
                       & 15\%   & 35.49          & 27.80          & 72.90          \\ 
                       & 20\%   & 35.98          & {\ul 29.13}    & 72.64          \\ 
                       & 25\%   & 35.96          & 28.25          & 75.03          \\ 
                       & 30\%   & 35.89          & 27.90          & 77.86          \\ 
                       & 50\%   & 38.28          & 25.76          & 76.21          \\ 
                       & 80\%   & {\ul 38.64}    & 20.63          & 63.49          \\ \hline
\multirow{3}{*}{\begin{tabular}[c]{@{}c@{}}KL\\ threhsold \\ (in percentile)\end{tabular}} &
  92.5\% &
  32.70 &
  25.08 &
  {\ul 80.58} \\ 
                       & 95\%   & \textbf{35.33} & \textbf{28.08} & \textbf{80.44} \\ 
                       & 97.5\% & 35.14          & 28.32          & 79.76          \\ \hline
\end{tabular}
\end{table}

\subsubsection{Anomaly characterization from label feedback using \texttt{HiLAD}}

\label{subsec:anomaly_characterization}

HiLAD can identify anomalies from different subspaces based on feedback. It takes into account the users' preferences (data point type and spatial location) to re-weight different subspaces. In contrast, classical anomaly detection approaches always propose candidates from the same ranked list. We demonstrate the performance of \texttt{HiLAD} on three cases (one on synthetic, and two others from representative real-world datasets (\textit{Yeast} and \textit{Abalone}).

\begin{figure}
    \subfloat[Anomaly(top right)]{
		\includegraphics[width=0.23\textwidth]{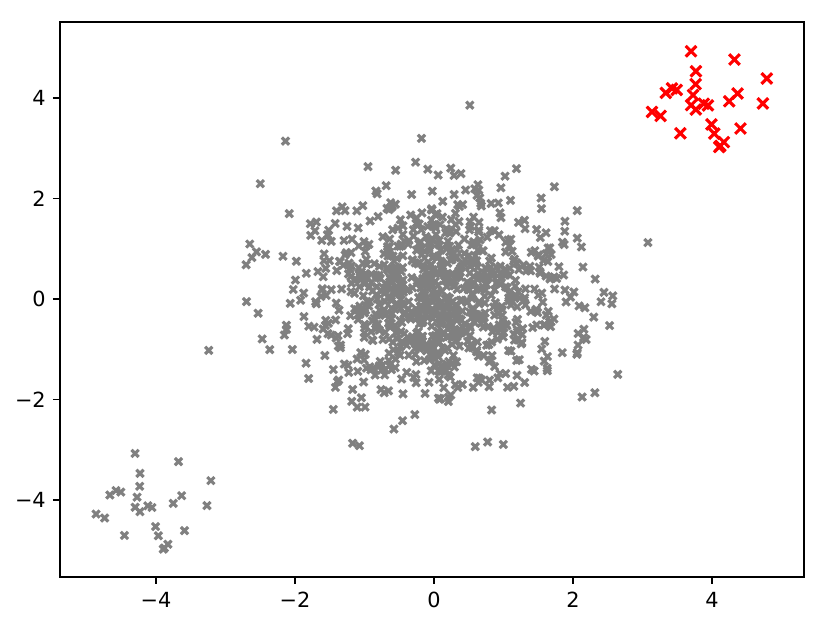}}
	\subfloat[Contour plots]{
		\includegraphics[width=0.23\textwidth]{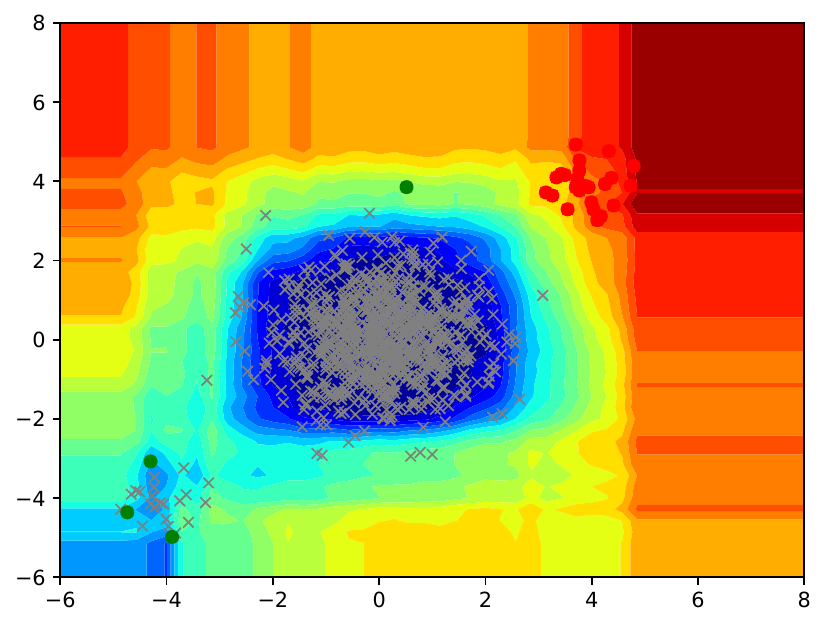}}
    \subfloat[Anomaly(bottom-left)]{
		\includegraphics[width=0.23\textwidth]{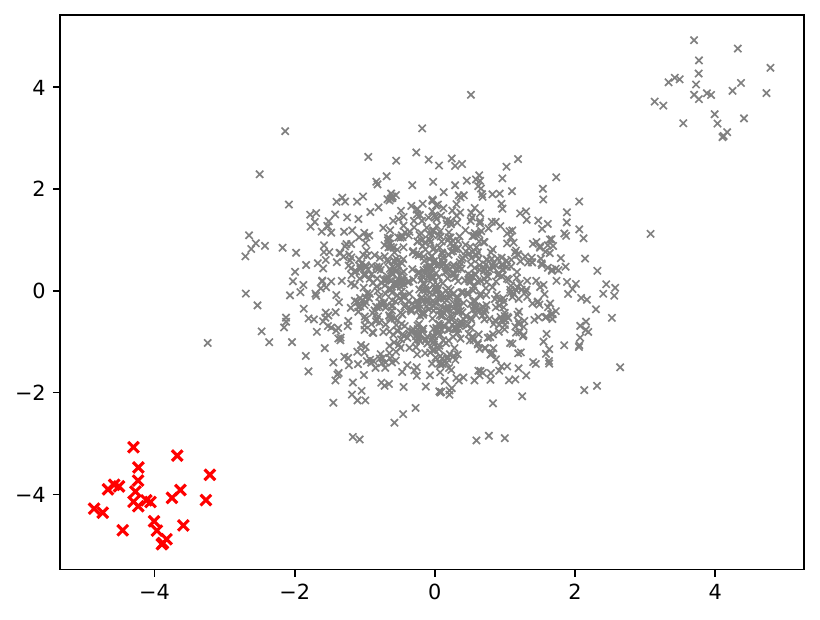}}
	\subfloat[Contour plots]{
		\includegraphics[width=0.23\textwidth]{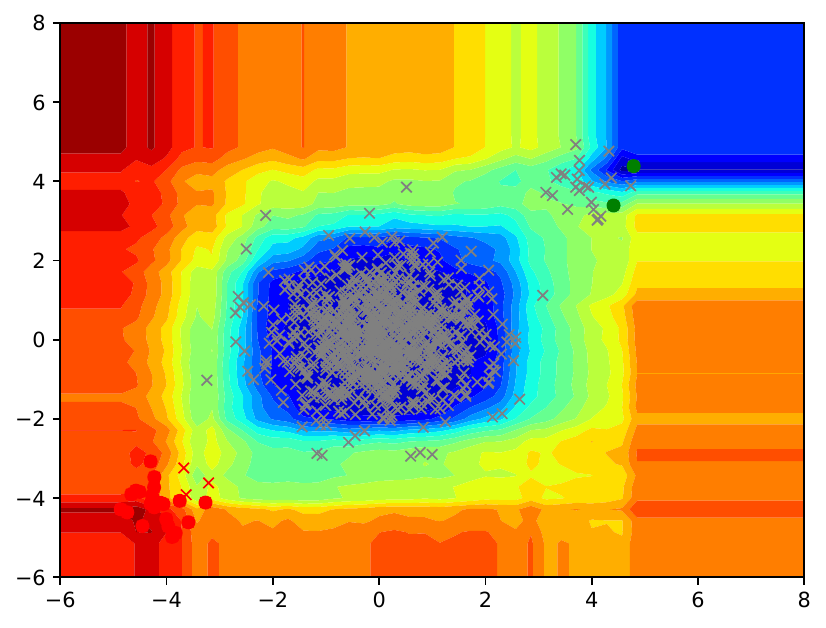}}
    \caption{Class-wise feedback experiment with synthetic datasets. Figure (a) and (c) are presenting two types of anomalies located two corners. And (b), (d) are showing how the algorithm rank data points across the data space.}
    \label{fig:class_feedback_synth}
\end{figure}
\begin{figure}
	\centering 
	\subfloat[Abalone]{
		\includegraphics[width=0.3\textwidth]{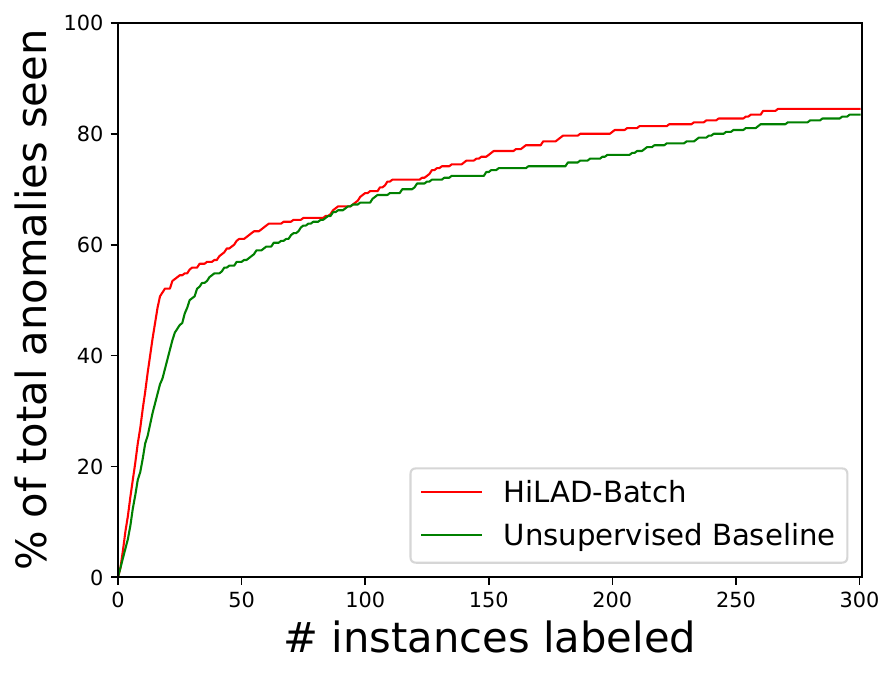}
		\label{fig:abalone_feedback}}
	\subfloat[Abalone(Class 3)]{
		\includegraphics[width=0.3\textwidth]{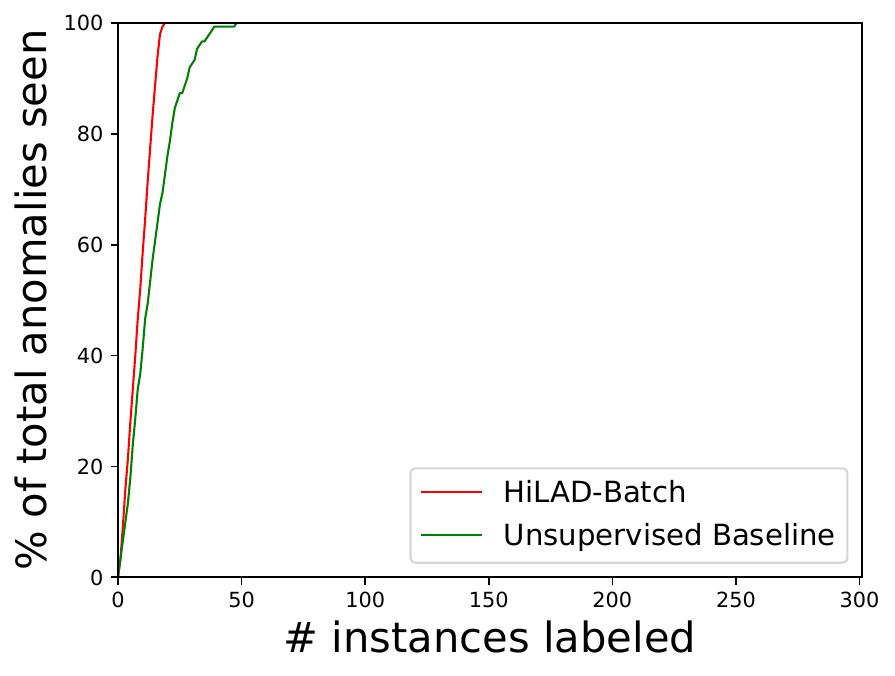}
		\label{fig:abalone_feedback_3}}
	\subfloat[Abalone(Class 21)]{
		\includegraphics[width=0.3\textwidth]{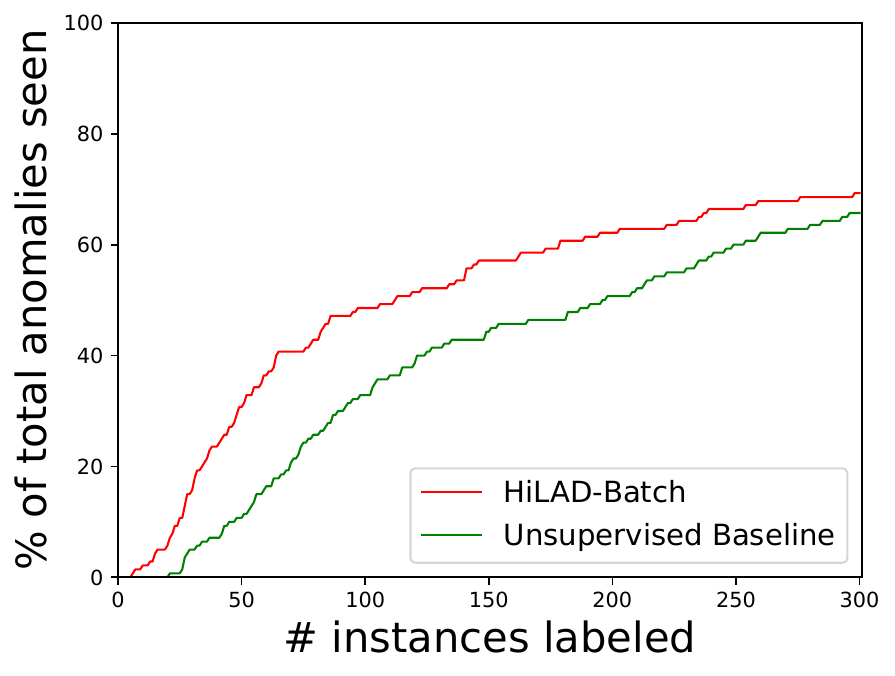}
		\label{fig:abalone_feedback_21}}
	\caption{Abalone dataset contains anomalies from two different classes. They are known as Class 3 and Class 21. The above figures are (a) feedback with HiLAD without class preferences (b) domain expert is interested in examples from \textit{Class 21}. (c) \textit{Class 3} is the point of interest. In all cases, we notice that getting feedback on a particular class can help us to recognize more anomalies from any particular class. (Note that for Figure (b) and (c) Y-axis represents what fraction of anomalies were discovered from that class.)}
	\label{fig:abalone}
\end{figure}
\begin{figure}
	\subfloat[Yeast]{
		\includegraphics[width=0.22\textwidth]{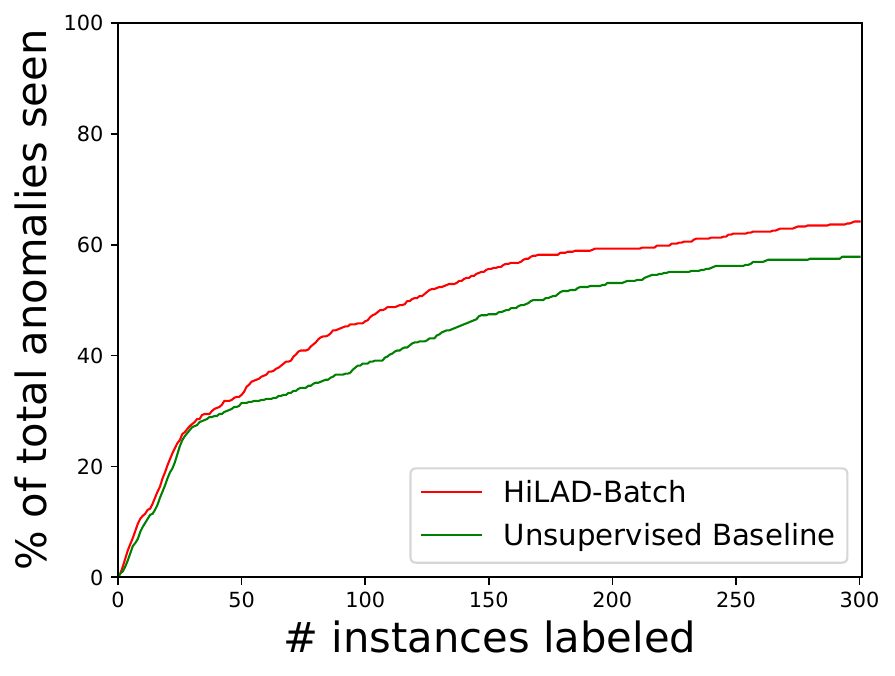}
		\label{fig:yeast_feedback}}
	\subfloat[Yeast(Class POX)]{
		\includegraphics[width=0.22\textwidth]{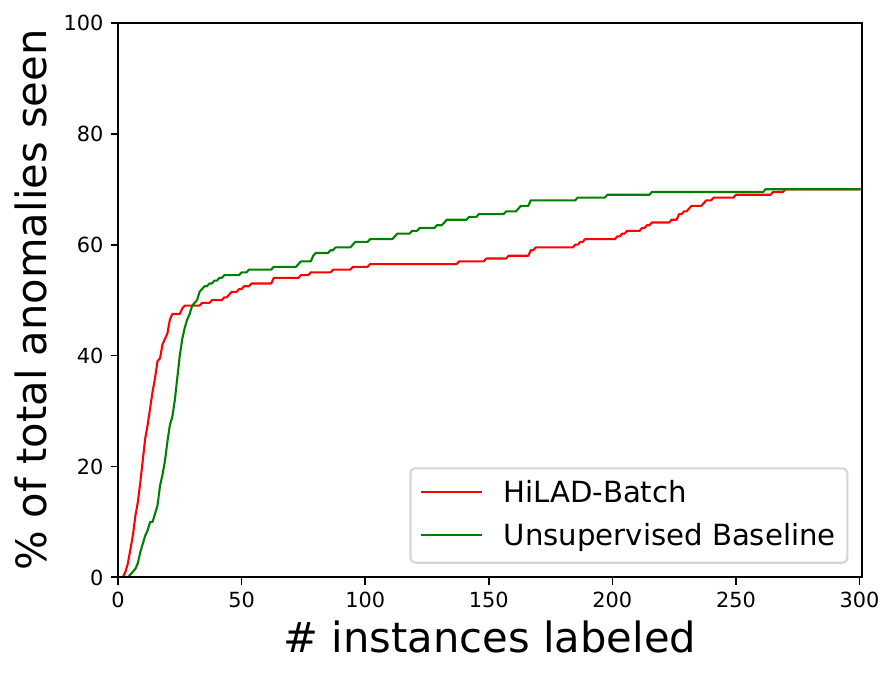}
		\label{fig:yeast_feedback_pox}}
	\subfloat[Yeast(Class VAC)]{
		\includegraphics[width=0.22\textwidth]{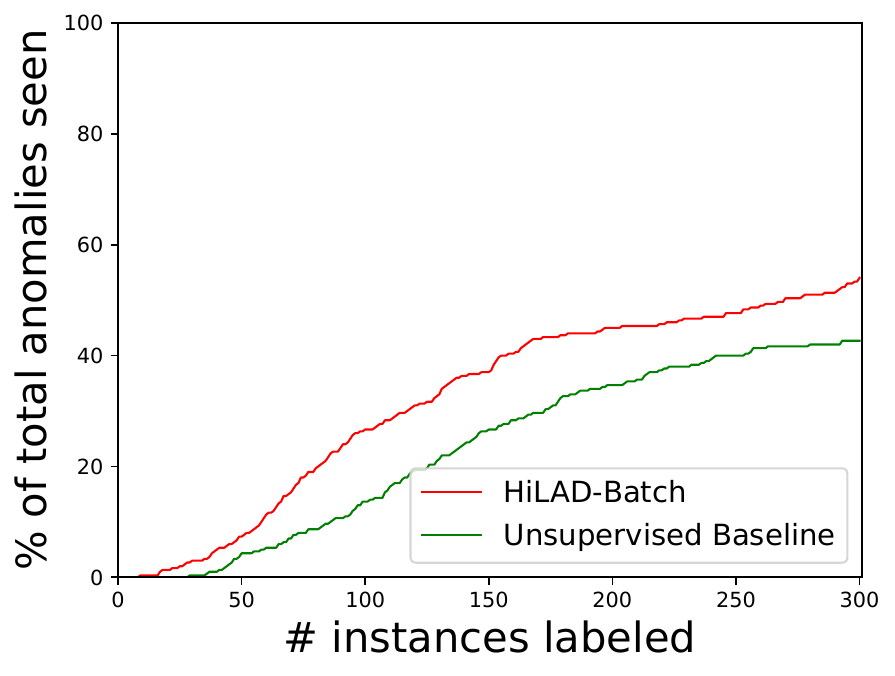}
		\label{fig:yeast_feedback_vac}}
	\subfloat[Yeast(Class ERL)]{
		\includegraphics[width=0.22\textwidth]{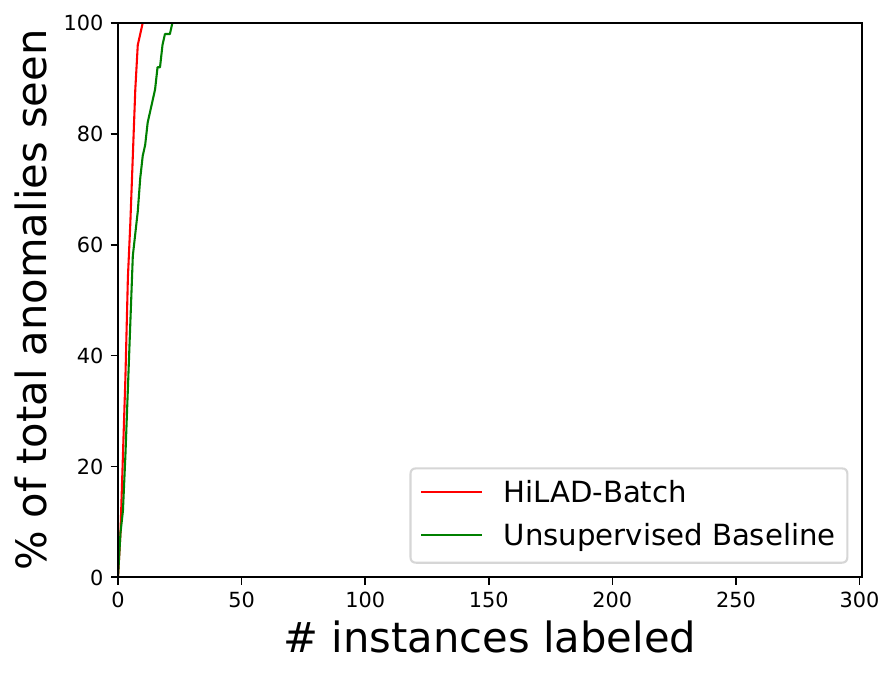}
		\label{fig:yeast_feedback_erl}}
	\caption{Yeast contains anomalies from three classes known as \textit{VAC, POX, ERL}. These plots are presenting HiLAD-Batch is able to identify anomalous examples from a specific type of examples. (Note that for Y-axis represents what fraction of anomalies were discovered from that class.)}
	\label{fig:yeast}
\end{figure}

\begin{enumerate}
    \item \textbf{Synthetic dataset} with two groups of anomalies created to provide controlled feedback. It's clear from the score contour plots  (Figure \ref{fig:class_feedback_synth}) that our algorithm pays attention where the domain expert is interested.
    \item \textbf{Abalone} contains two anomaly classes (\textit{Class 3} and \textit{Class 21}). In our experiments, we provided feedback from one class as anomalies and found that \texttt{HiLAD} ranked anomalies from that class higher  as shown in Figure \ref{fig:abalone}.
    \item \textbf{Yeast} contains three anomaly classes (\textit{POX}, \textit{VAC}, and \textit{ERL}). The results of the experiment where we provide feedback on only one class of anomalies is shown in Figure \ref{fig:yeast}.
\end{enumerate}

\begin{figure}
    \subfloat[Abalone tSNE with class label]{
		\includegraphics[width=0.48\textwidth]{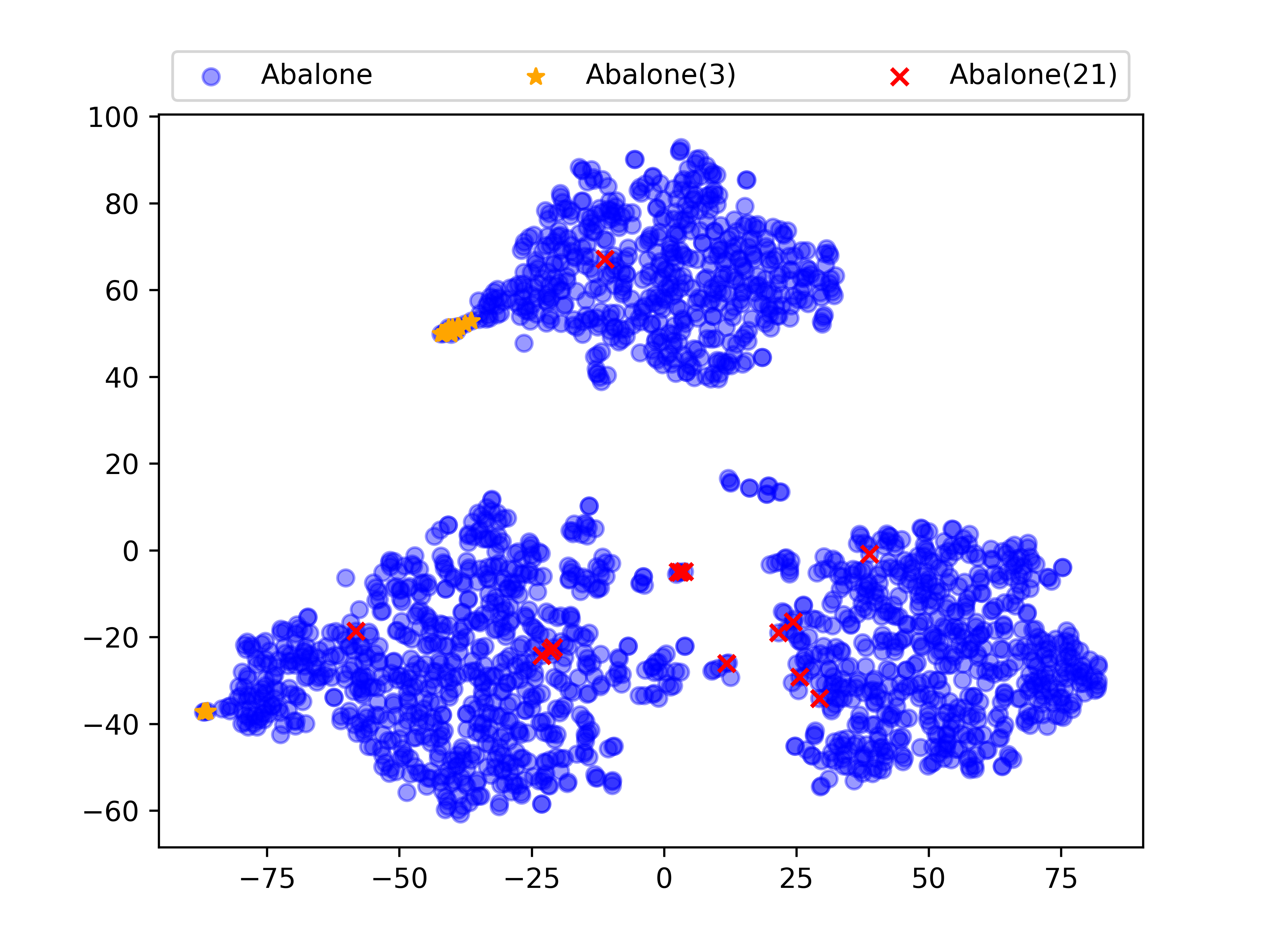}
  }
	\subfloat[Yeast tSNE with class label]{
		\includegraphics[width=0.48\textwidth]{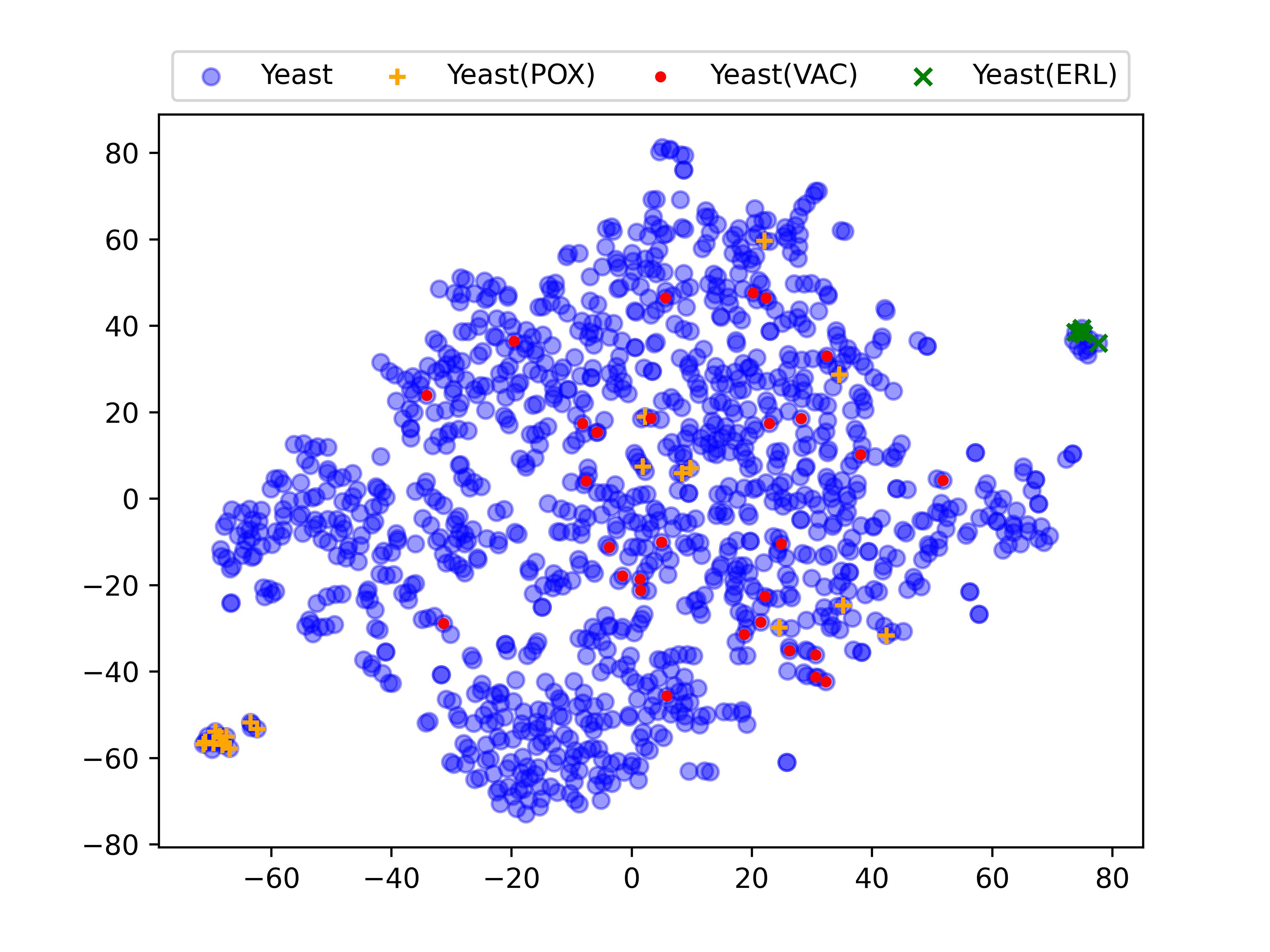}
  }
    \caption{tSNE plot for Abalone and Yeast. Samples for each class are marked and color codes. They are widespread on the data space}
    \label{fig:class_feedback_tsne}
\end{figure}

We see similar behavior of the algorithm for classes \textit{VAC}(Figure \ref{fig:yeast_feedback_vac}) and \textit{ERL}(Figure \ref{fig:yeast_feedback_erl}) as we did for both classes of the \textit{Abalone} dataset. However, for class \textit{POX} we see an initial higher gain with \texttt{HiLAD-Batch} than the unsupervised baseline before the unsupervised baseline overtakes \texttt{HiLAD-Batch}. We took a closer look into the cause for this using t-SNE plots (Figure \ref{fig:class_feedback_tsne}). Here we see that there is one clustered group of anomalies (lower left) that gets discovered fast initially by \texttt{HiLAD-Batch} and thus explains its initial steeper rise. But then, rest of the anomalies belonging to \textit{POX} are mixed within the nominal data and hence harder to discover. Since these might be spread out among nominals, positive feedback on one such anomaly does not get propagated to other anomalies efficiently. Moreover, a negative feedback on a nominal data instance (a very likely scenario for this situation) lowers the priority of an anomaly that is present in its proximity.

\subsection{Comparison with Recent Semi-Supervised Approaches}
\label{subsec:semisupervised-baseline-comparison}

\begin{figure}
	\centering
    \subfloat[Abalone]{\includegraphics[width=0.3\linewidth]{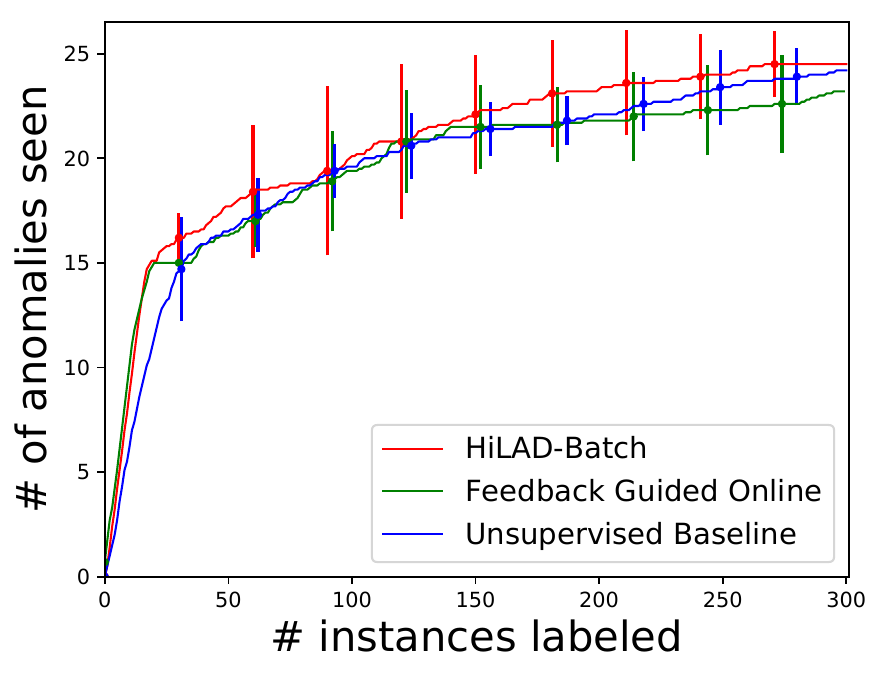}%
		\label{fig:fb_abalone}}
	\subfloat[ANN-Thyroid-1v3]{\includegraphics[width=0.3\linewidth]{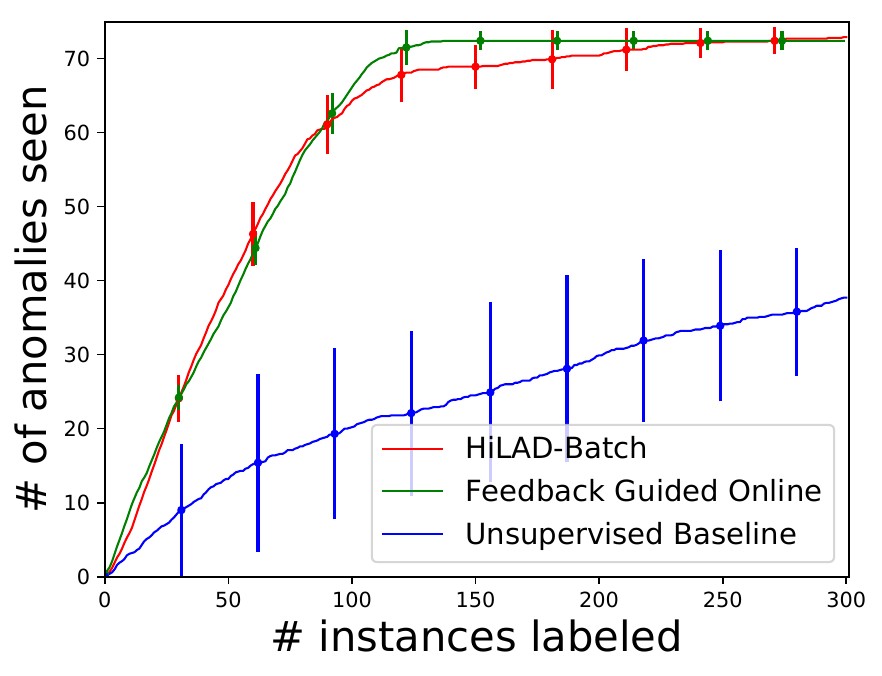}%
		\label{fig:fb_ann_thyroid_1v3}}
	\subfloat[Cardiotocography]{\includegraphics[width=0.3\linewidth]{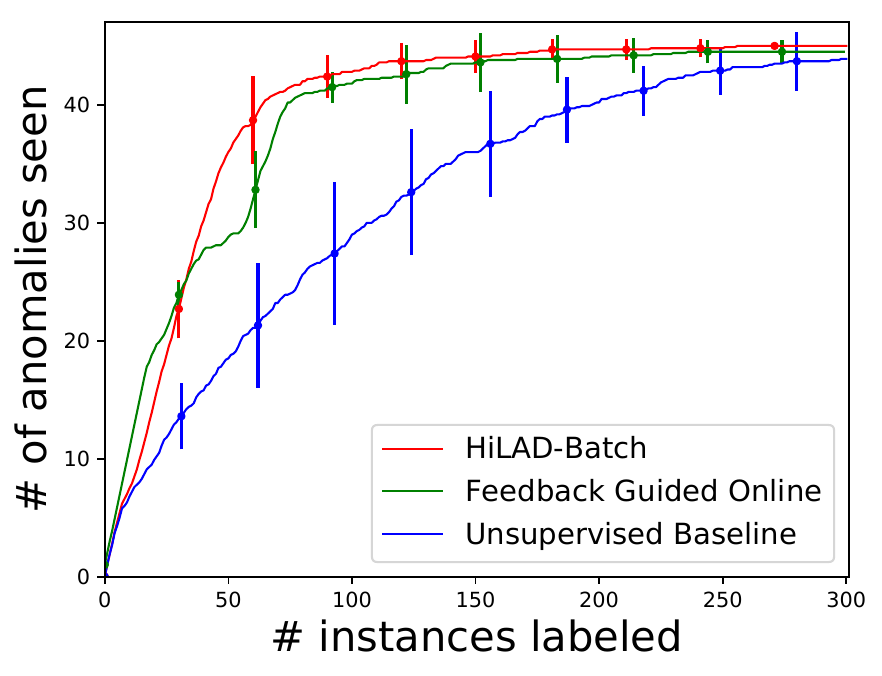}%
		\label{fig:fb_cardiotocography}}\\
	\subfloat[Electricity]{\includegraphics[width=0.3\linewidth]{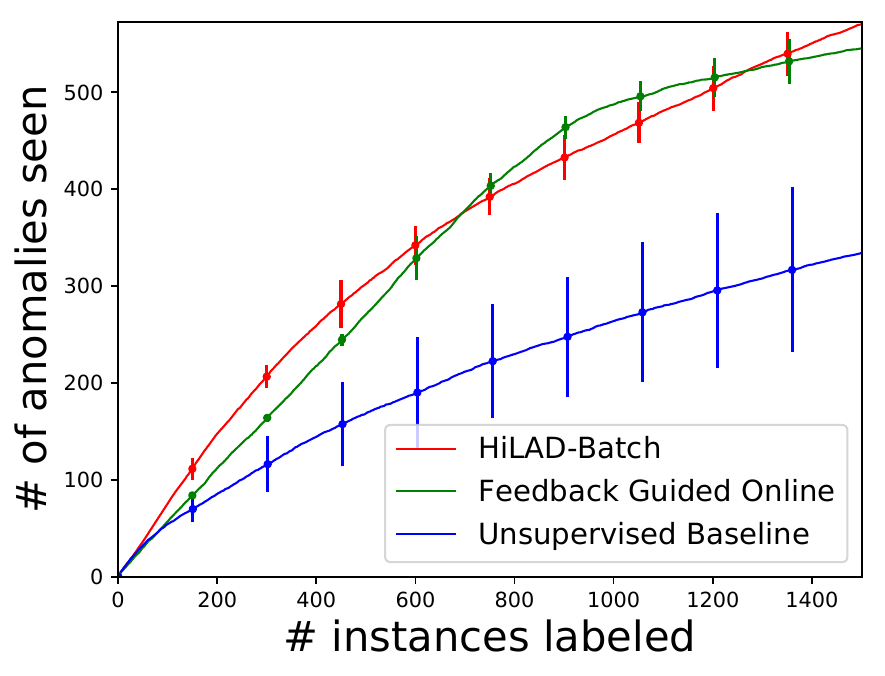}%
		\label{fig:fb_electricty}}
	\subfloat[Mammography]{\includegraphics[width=0.3\linewidth]{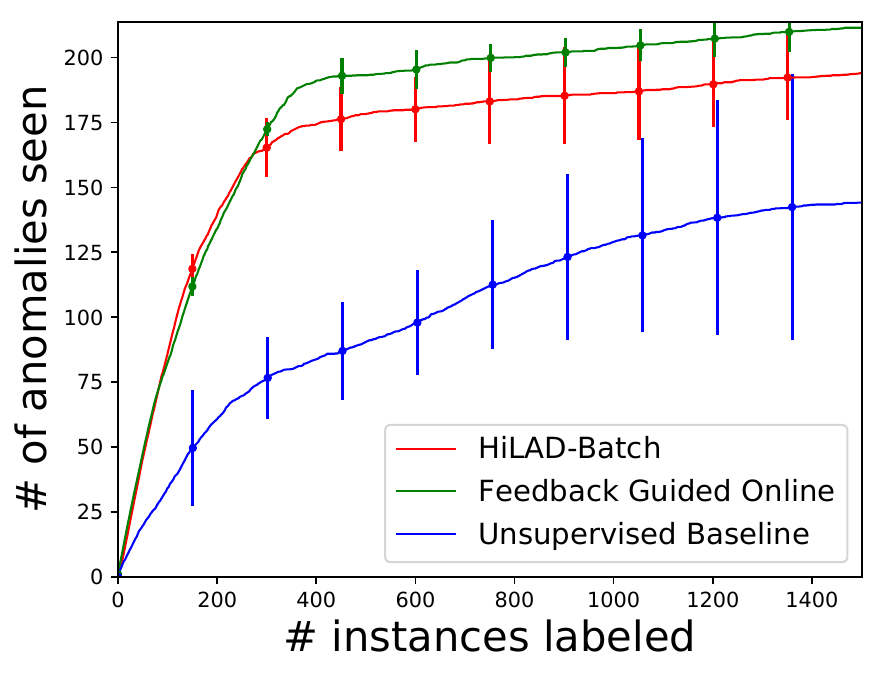}%
		\label{fig:fb_mammography}}
	\subfloat[Shuttle]{\includegraphics[width=0.3\linewidth]{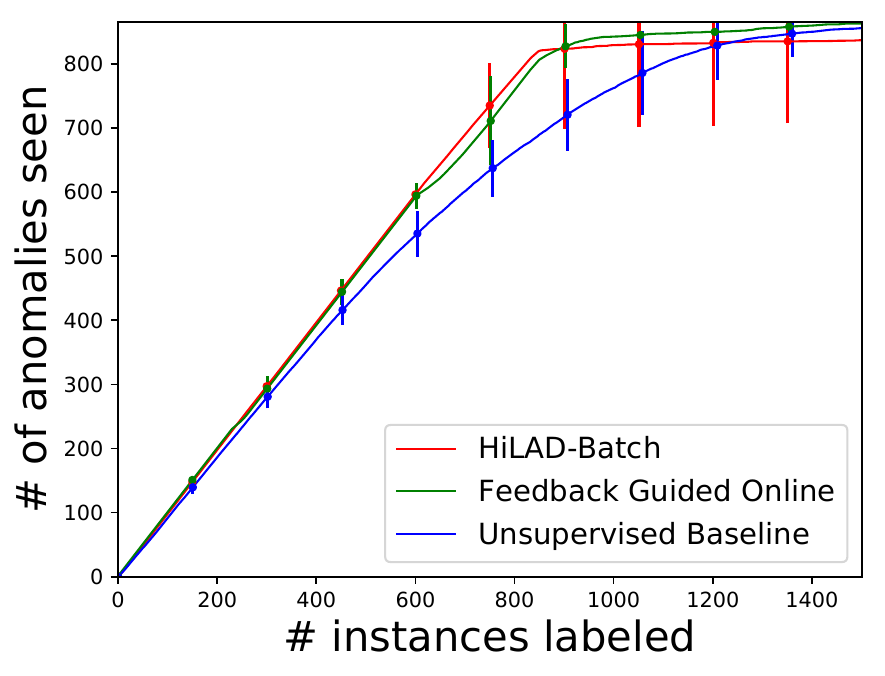}%
		\label{fig:fb_shuttle}}\\
	\subfloat[Weather]{\includegraphics[width=0.35\linewidth]{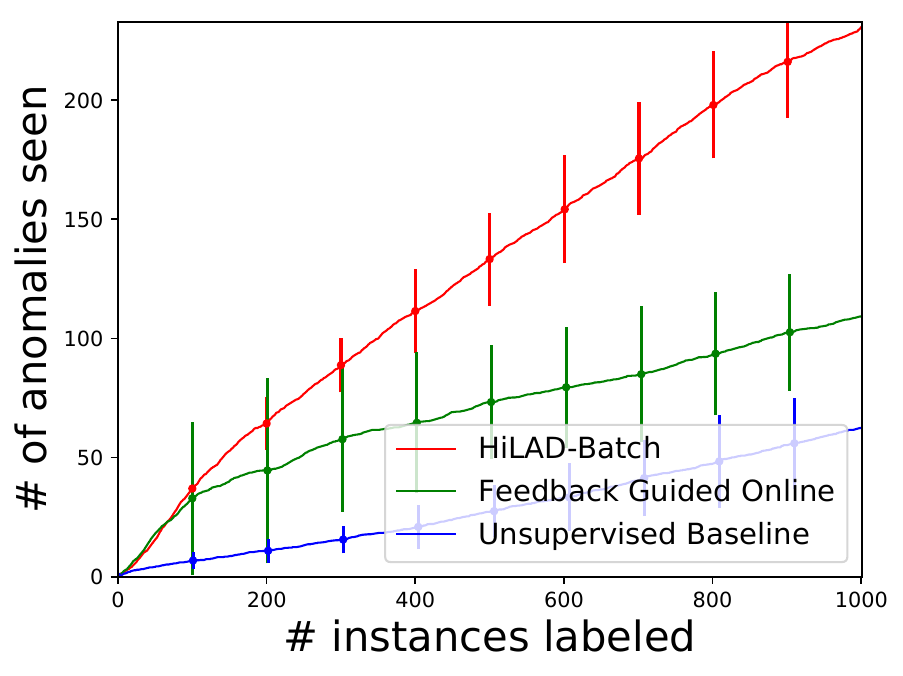}%
		\label{fig:fb_weather}}
	\subfloat[Yeast]{\includegraphics[width=0.35\linewidth]{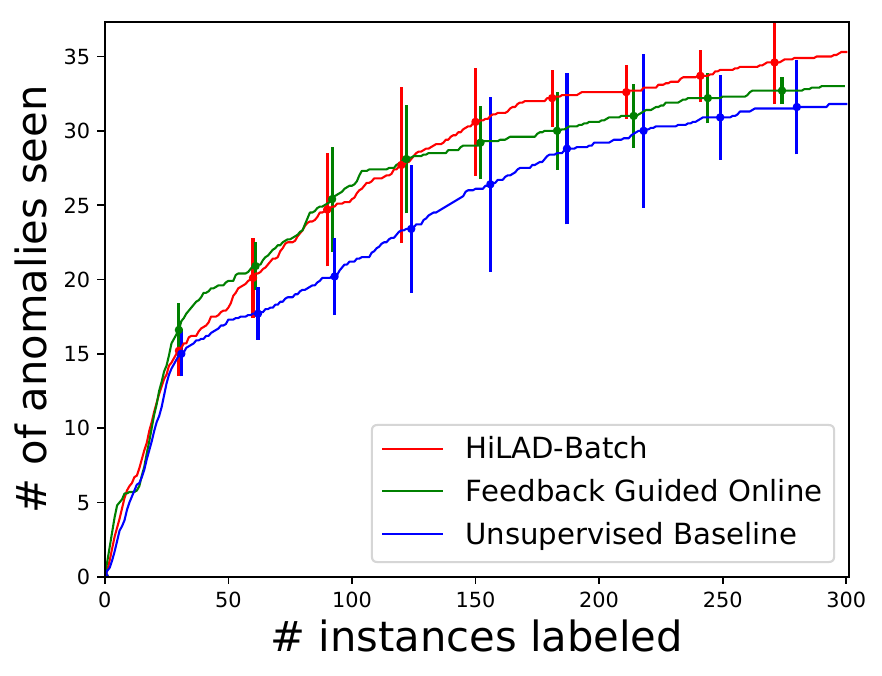}%
		\label{fig:fb_yeast}}
		
    \caption{Results comparing HiLAD-Batch with feedback-guided anomaly detection via online optimization \cite{siddiqui:2018}. \texttt{HiLAD-Batch} is the tree-based model implemented in our codebase and employs the AAD loss (anomalies score higher than $\tau$-th quantile score and nominals lower). \texttt{Feedback Guided Online} employs the linear loss in Siddiqui et al., 2018. \texttt{Unsupervised Baseline} is the unsupervised Isolation Forest baseline. Both approaches perform similar on most datasets. While \texttt{HiLAD-Batch} has slightly poor accuracy on \textit{Mammography} than \texttt{Feedback Guided Online}, \texttt{HiLAD-Batch} performs much better on \textit{Weather}.}
    \label{fig:fbonline_all}
\end{figure}

We propose effective instantiations of the generic human-in-the-loop learning framework \texttt{HiLAD} for anomaly discovery. Prior works including \cite{siddiqui:2018}, \cite{das:2016}, \cite{das:2017} are notable methods in this line of work. All these prior methods used ensemble based approaches as a black-box system \cite{pevny:2015}. However, we used tree-based ensembles / Isolation Forest \cite{liu:08} and exploited the inherent structure of the trees to effectively utilize human feedback. Our experimental evaluation considers all these works as baselines. Moreover, we did a fine-grained experimental comparison with \cite{siddiqui:2018}. Figure~\ref{fig:fbonline_all} shows the comparison of \texttt{HiLAD-Batch} with feedback-guided anomaly detection via online optimization (\texttt{Feedback-guided Online}) \cite{siddiqui:2018}. The results for {\em KDDCup99} and {\em Covtype} could not be included for \texttt{Feedback-guided Online} because their code\footnote{\url{https://github.com/siddiqmd/FeedbackIsolationForest}(retrieved on 10-Oct-2018)} resulted in error ({\em Segmentation Fault}) when run with 3000 feedback iterations (a reasonable budget for the large datasets).

\subsubsection{Comparison with Limited Labeled Set}
\label{subsec:limited-label-baseline-comparison}

In this section, we present a comparison with SOEL using the same experimental setup followed in \cite{li2023-soel}. SOEL is a semi-supervised algorithm which, as per its original design, does not expect a continuous user feedback cycle. Instead, SOEL assumes that a limited amount of labeled data is already available. Under the experimental setup, the datasets are first split into two parts (\textit{train, test}). Next, a fixed number of labels $B$ are collected by following one of the nine query strategies from previous works: (1) \textit{Mar} \cite{gornitz:2013}, (2) \textit{Hybr1} \cite{gornitz:2013}, (3) \textit{Pos1} \cite{Pimentel_2020}, (4) \textit{Pos2} \cite{barnabe2015active}, (5) \textit{Rand1} \cite{Ruff2018DeepOC}, (6) \textit{Rand2} \cite{trittenbach2021overview}, (7) \textit{Hybr2}, (8) \textit{Hybr3} \cite{ning2022deep}, and (9) \textit{SOEL} \cite{li2023-soel}. Finally, the F1-score for each query strategy is reported averaged over five runs. The train-test splits and initialization were random for each experiment run. \textit{Hybr2} is attributed to a strategy discussed in \url{https://github.com/shubhomoydas/ad_examples} but was implemented incorrectly in \cite{li2023-soel} with minimum euclidean distance between queried instances. The correct implementation \textit{maximizes} the euclidean distance to improve diversity.

Our first set of comparisons with SOEL (Table \ref{tab:SOEL-2023-baseline}) are based on the four tabular datasets from \cite{li2023-soel}. In these datasets, more than 30\% samples are considered anomalies, especially in the \textit{test} set. \textbf{HiLAD} performed best on two out of four datasets. \textbf{SOEL} performed best on \textit{Satellite}, and \textbf{Hybr1} was best on \textit{Ionosphere}. Arguably, this setup has a much higher proportion of anomalies than in real-world scenarios.

Our second set of comparisons with SOEL are based on the seven datasets from Table \ref{tab:datasets_full} where \% of anomalies present in the test set are more realistic ($<10\%$). We follow the same training methodology as in the previous experiment and report the results in Table \ref{tab:labeled-set-models-comparison}. Here, \textbf{HiLAD} achieved the best F1-score on six out of seven datasets. A key takeaway from both sets of results (Table \ref{tab:SOEL-2023-baseline} and Table \ref{tab:labeled-set-models-comparison}) is that HiLAD's performance is consistent across different datasets. 

\begin{figure}
	\centering
    \subfloat[Abalone]{\includegraphics[width=0.32\linewidth]{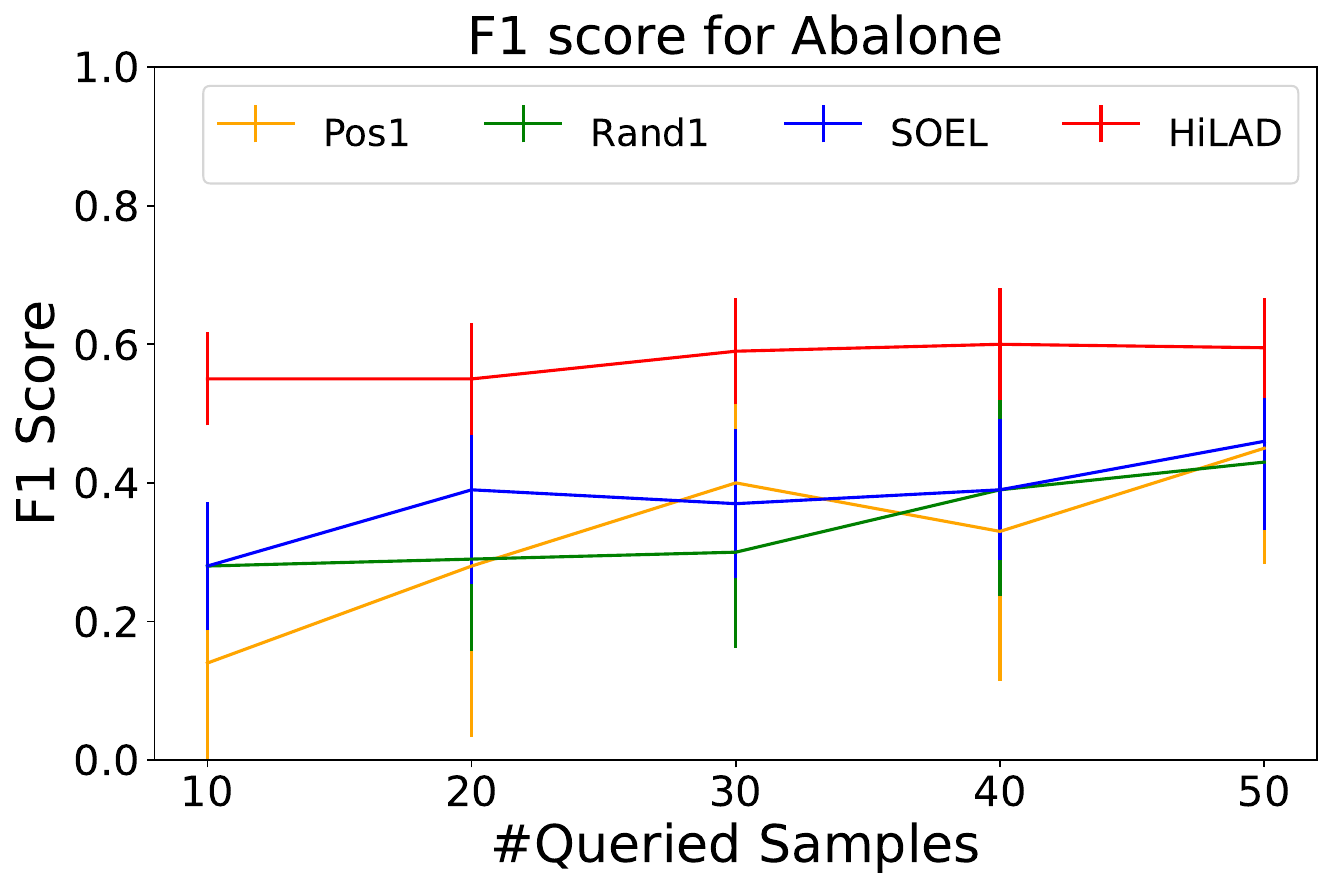}%
		\label{fig:semi_abalone}}
	\subfloat[ANN-Thyroid-1v3]{\includegraphics[width=0.32\linewidth]{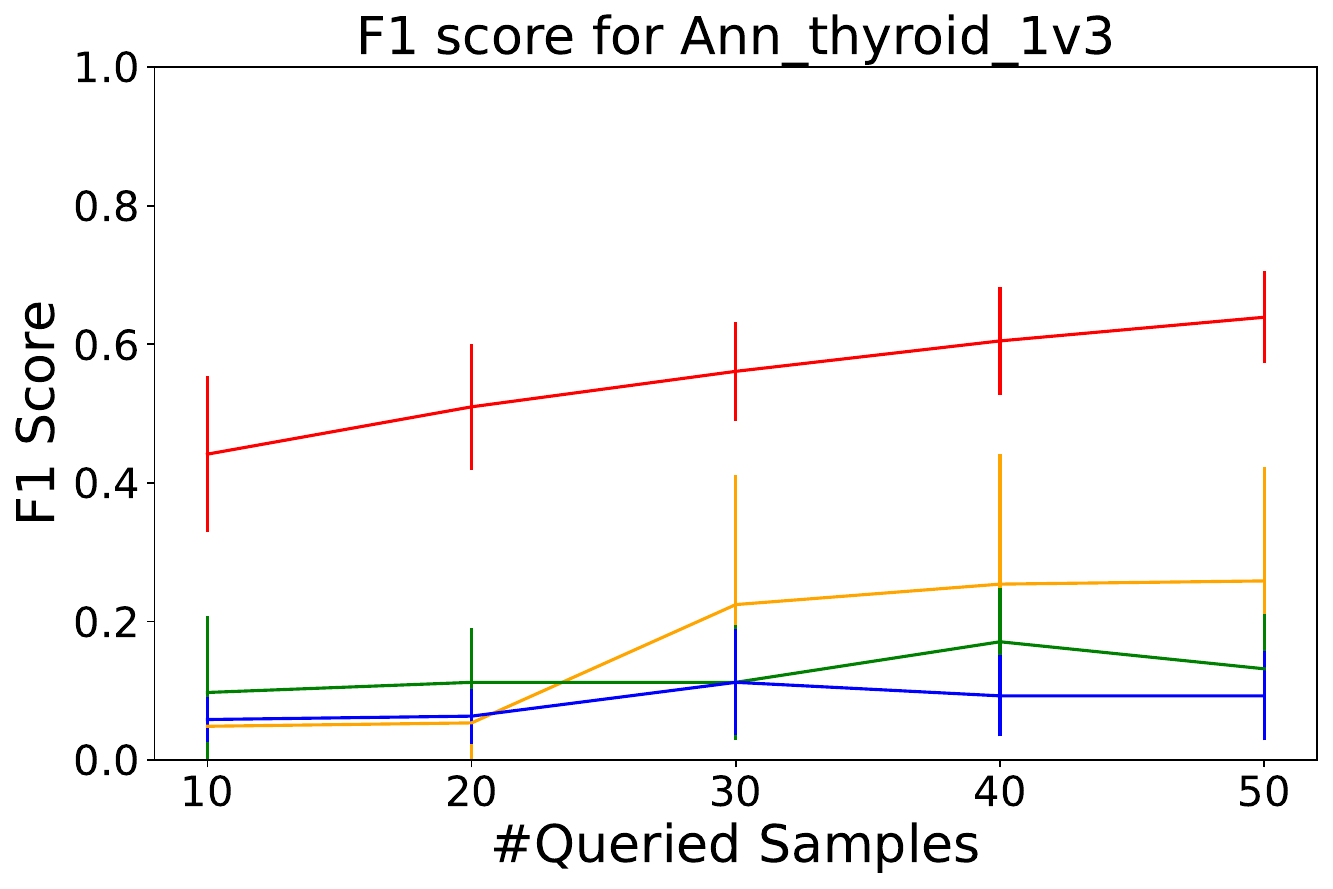}%
		\label{fig:semi_ann_thyroid_1v3}}
	\subfloat[Cardiotocography]{\includegraphics[width=0.32\linewidth]{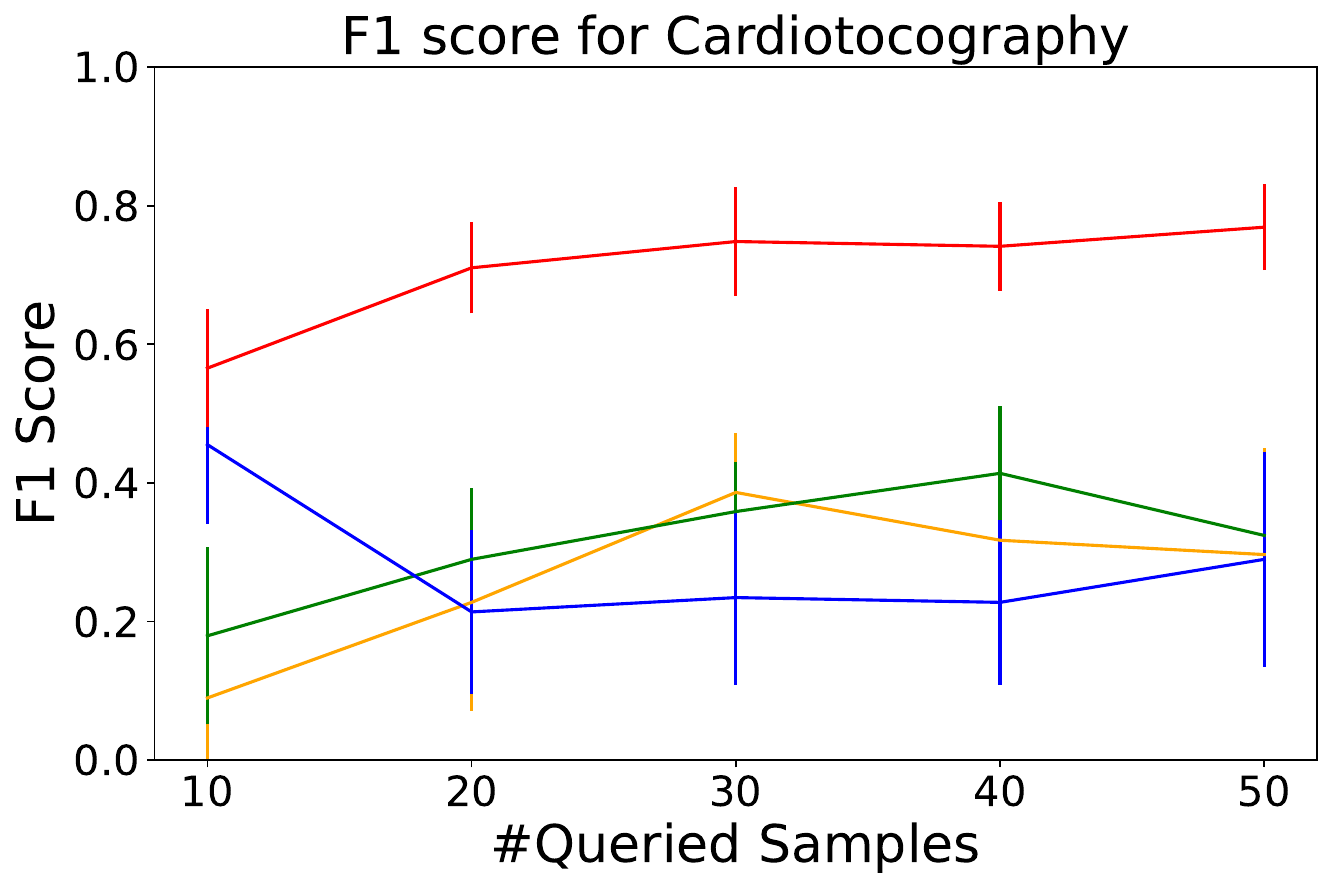}%
		\label{fig:semi_cardiotocography}}\\
  	\subfloat[KDD-Cup-99]{\includegraphics[width=0.32\linewidth]{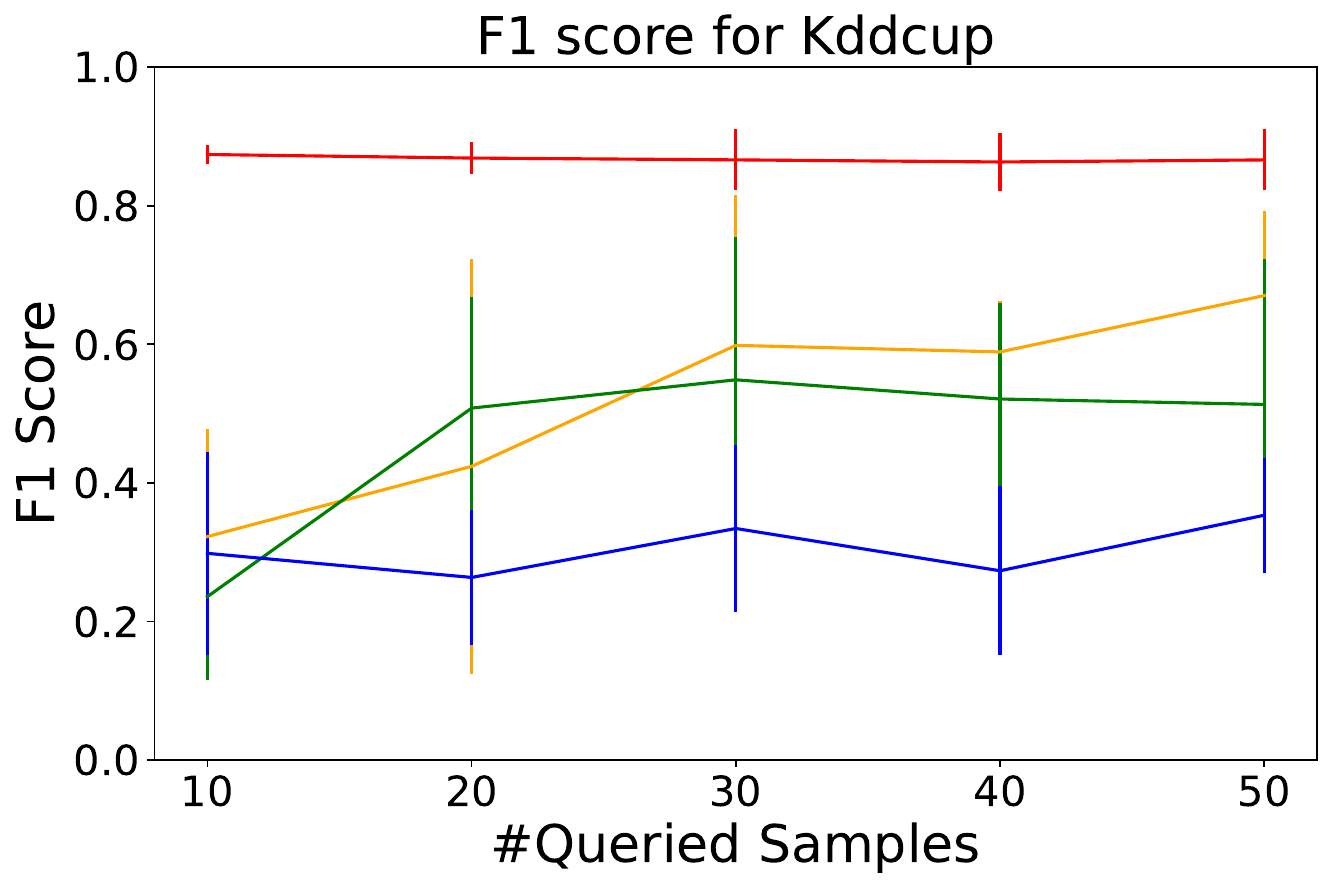} %
		\label{fig:semi_kdd_cup}}
	\subfloat[Mammography]{\includegraphics[width=0.32\linewidth]{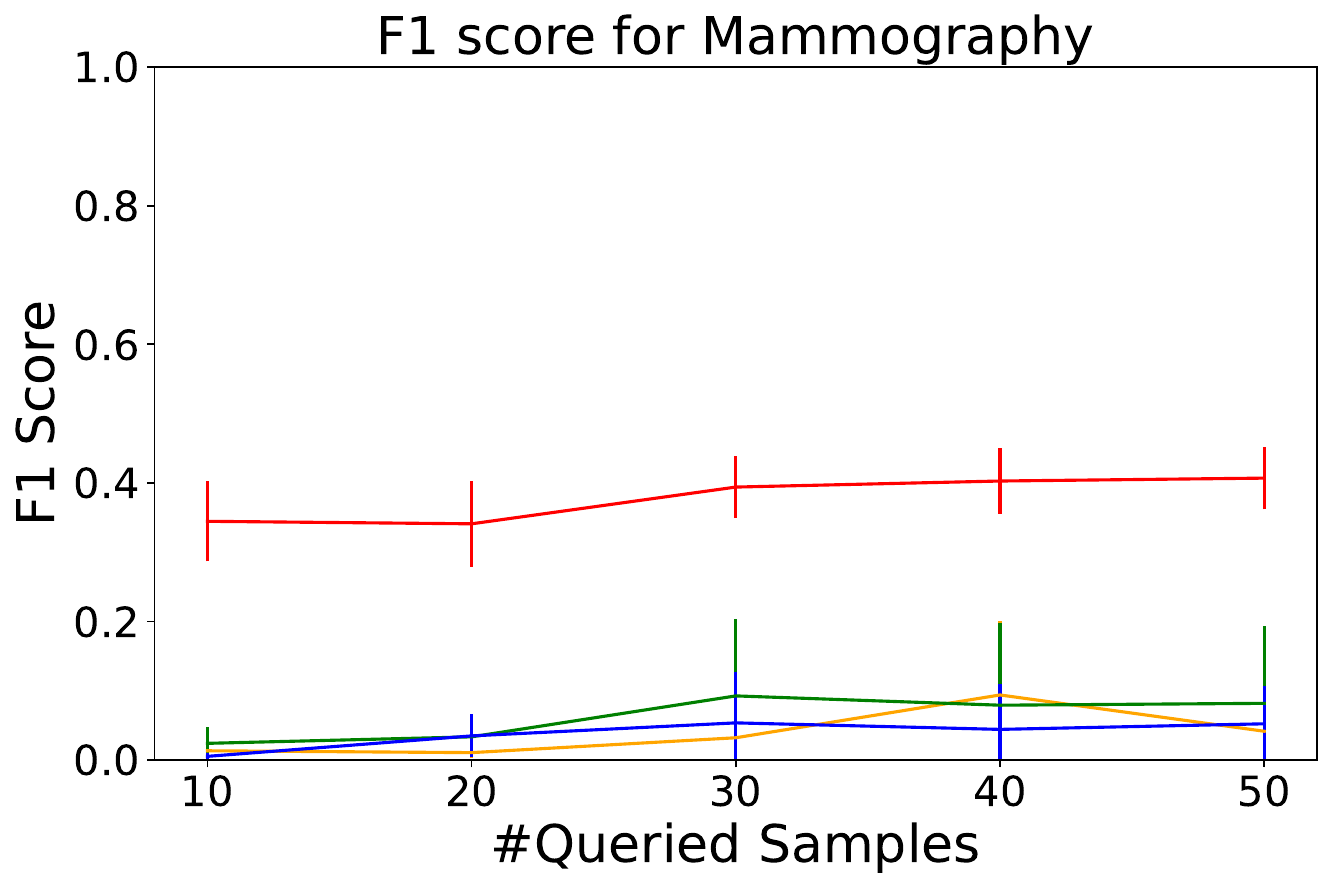}%
		\label{fig:semi_mammography}}
	\subfloat[Shuttle]{\includegraphics[width=0.32\linewidth]{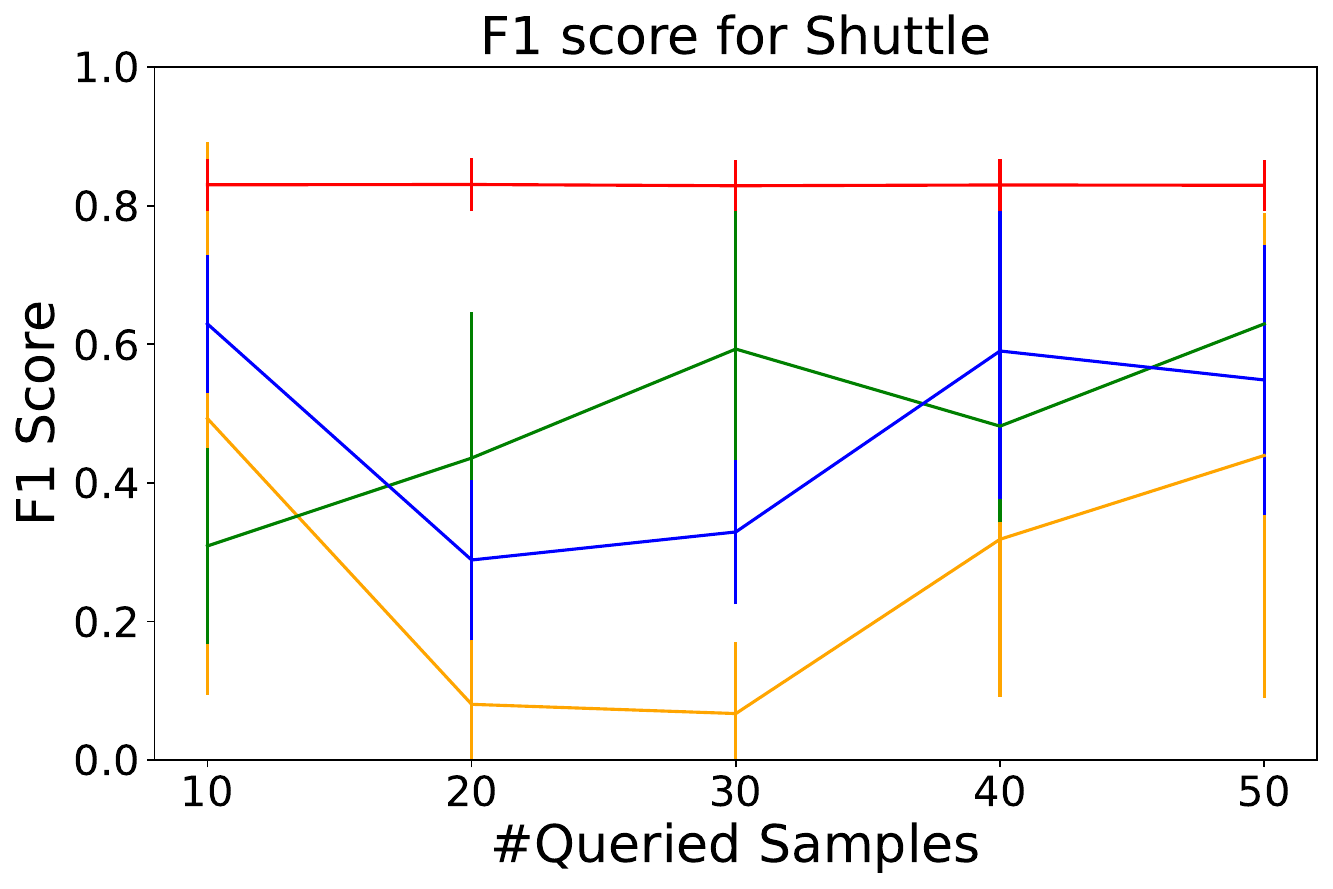}%
		\label{fig:semi_shuttle}}\\
	\subfloat[Yeast]{\includegraphics[width=0.32\linewidth]{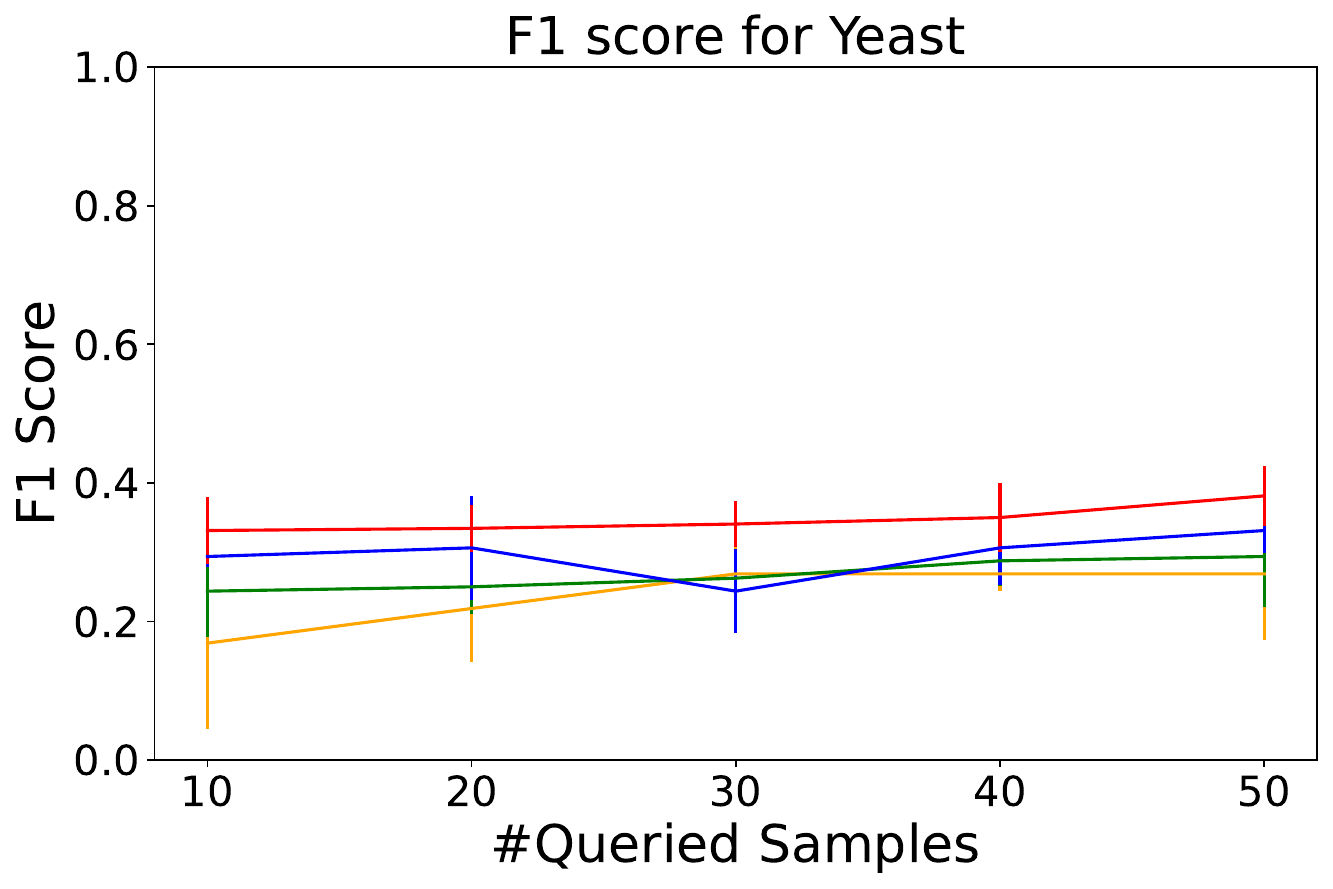}%
		\label{fig:semi_yeast}}
  \hspace{25.5em}

    \caption{Comparison of F1 scores on the seven datasets with different query budgets (10, 20, 30, 40, and 50). HiLAD performs better as it gets more labeled samples.}
    \label{fig:aad_vs_other_semi_supervised_all}
\end{figure}

Our third set of experiments are again based on datasets from Table \ref{tab:datasets_full}. Here, we compare the performance of algorithms as the amount of labeled data $B$ during training time is increased in the following order: $10, 20, 30, 40, 50$. The results are reported in Figure \ref{fig:aad_vs_other_semi_supervised_all}. For some datasets, we didn't have 50 anomalous samples present. One interesting observation is that with 10 labeled samples, HiLAD was not the best performer for \textit{Cardiotocography}. However, as more labeled samples were added, HiLAD's performance improved, unlike other methods whose performance didn't.

\begin{table}[]
\centering
\caption{F1-score (\%) with standard deviation for anomaly detection on tabular data when the query budget $|B|$ = 10. HiLAD performs the best on two of the  four datasets whereas SOEL and Hybr1 performed the best on the rest two.}
\label{tab:SOEL-2023-baseline}
\resizebox{\columnwidth}{!}{%
\begin{tabular}{crrrrrrrrrr}
\hline
Dataset name & \multicolumn{1}{c}{\textbf{Mar}} & \multicolumn{1}{c}{\textbf{Hybr1}} & \multicolumn{1}{c}{\textbf{Pos1}} & \multicolumn{1}{c}{\textbf{Pos2}} & \multicolumn{1}{c}{\textbf{Rand1}} & \multicolumn{1}{c}{\textbf{Rand2}} & \multicolumn{1}{c}{\textbf{Hybr2}} & \multicolumn{1}{c}{\textbf{Hybr3}} & \multicolumn{1}{c}{\textbf{SOEL}} & \multicolumn{1}{c}{\textbf{HiLAD}} \\ \hline
BreastW & 81.6±0.7 & 83.3±2.0 & 58.6±7.7 & 81.3±0.8 & 87.1±1.0 & 82.9±1.1 & 55.0±6.0 & 79.6±4.9 & 93.9±0.5 & \textbf{0.958±0.00} \\
Ionosphere & 91.9±0.3 & \textbf{92.3±0.5} & 56.1±6.2 & 91.1±0.8 & 91.1±0.3 & 91.9±0.6 & 64.0±4.6 & 88.2±0.9 & 91.8±1.1 & 0.843±0.02 \\
Pima & 50.1±1.3 & 49.2±1.9 & 48.5±0.4 & 52.4±0.8 & 53.6±1.1 & 51.9±2.0 & 53.8±4.0 & 48.4±0.7 & 55.5±1.2 & \textbf{0.635±0.02} \\
Satellite & 64.2±1.2 & 66.2±1.7 & 57.0±3.0 & 56.7±3.2 & 67.7±1.2 & 66.6±0.8 & 48.6±6.9 & 56.9±7.0 & \textbf{71.1±1.7} & 0.674±0.00 \\ \hline
\end{tabular}%
}
\end{table}

\begin{table}[]
\centering
\caption{F1-score (\%) with standard deviation for anomaly detection on tabular data when the query budget $|B|$ = 10. HiLAD performs the best on six of the seven datasets.}
\label{tab:labeled-set-models-comparison}
\resizebox{\columnwidth}{!}{%
\begin{tabular}{crrrrrrrrrr}
\hline
\textbf{Dataset} & \multicolumn{1}{c}{\textbf{Rand1}} & \multicolumn{1}{c}{\textbf{Rand2}} & \multicolumn{1}{c}{\textbf{Mar}} & \multicolumn{1}{c}{\textbf{Pos1}} & \multicolumn{1}{c}{\textbf{Pos2}} & \multicolumn{1}{c}{\textbf{Hybr1}} & \multicolumn{1}{c}{\textbf{Hybr2}} & \multicolumn{1}{c}{\textbf{Hybr3}} & \multicolumn{1}{c}{\textbf{SOEL}} & \multicolumn{1}{c}{\textbf{HiLAD}} \\ \hline
Abalone & 0.28±0.04 & 0.31±0.15 & 0.3±0.18 & 0.14±0.14 & 0.23±0.2 & 0.32±0.21 & 0.25±0.05 & 0.31±0.1 & 0.28±0.09 & \textbf{0.47±0.13} \\
ANN-Thyroid-1v3 & 0.1±0.11 & 0.15±0.12 & 0.1±0.11 & 0.05±0.02 & 0.11±0.1 & 0.13±0.13 & 0.19±0.08 & 0.09±0.09 & 0.06±0.03 & \textbf{0.44±0.11} \\
Cardiotocography & 0.18±0.13 & 0.23±0.09 & 0.29±0.07 & 0.09±0.12 & 0.24±0.1 & 0.27±0.08 & \textbf{0.59±0.1} & 0.41±0.15 & 0.46±0.11 & 0.57±0.08 \\
KDD-Cup-99 & 0.24±0.12 & 0.3±0.16 & 0.15±0.06 & 0.32±0.15 & 0.23±0.14 & 0.29±0.13 & 0.4±0.1 & 0.34±0.05 & 0.3±0.15 & \textbf{0.87±0.01} \\
Mammography & 0.02±0.02 & 0.01±0.01 & 0±0 & 0.01±0.02 & 0±0.01 & 0.02±0.02 & 0.01±0.01 & 0.01±0.02 & 0.01±0.01 & \textbf{0.34±0.06} \\
Shuttle & 0.31±0.14 & 0.36±0.19 & 0.42±0.23 & 0.49±0.4 & 0.52±0.16 & 0.46±0.17 & 0.66±0.29 & 0.55±0.18 & 0.63±0.1 & \textbf{0.83±0.04} \\
Yeast & 0.24±0.07 & 0.21±0.09 & 0.24±0.02 & 0.17±0.12 & 0.24±0.05 & 0.28±0.02 & 0.21±0.11 & 0.25±0.09 & 0.29±0.02 & \textbf{0.33±0.05} \\ \hline
\end{tabular}%
}
\end{table}

\color{black}

\subsection{Broader Applicability of Insights and Algorithmic Ideas}

We employed the tree-based approach as a device to demonstrate how our generic mechanism for incorporating labeled anomalies can be applied to ensembles of anomaly detectors. This technique can be applied in situations where an ensemble of detectors can be trained inexpensively (e.g., using feature bagging \cite{lazarevic:05}, LODA \cite{pevny:2015}, IFOR \cite{liu:08} etc.) or with more sophisticated methods \cite{senator:2013}. We have focused on tree-based methods in this paper due to their popularity and frequent state-of-the-art performance. Our algorithm makes an assumption that the data is \textit{i.i.d.}. distributed in the form of tabular data with feature values and hence, we did not apply it \textit{directly} to raw data such as text and images. However, this is not a serious limitation. Deep-learning-based anomaly detection methods that use deep learning to extract features \cite{pang2021deep} can employ our proposed algorithm on the representation of the data in the latent space where it is assumed that the data is \textit{i.i.d.} distributed.

The proposed technique treats the subspaces at the tree leaves to be ensemble components. GLAD \cite{das2018glad} uses the insights from this algorithm and extends it to the extreme case where each instance is treated as a subspace; and then, instead of trees, uses neural networks to incorporate feedback using a conceptually similar loss function as proposed this paper. The strong results of GLAD algorithm demonstrate that the key ideas behind the proposed algorithm are versatile.

\subsection{Summary of Experimental Findings}

We briefly summarize the main findings of our empirical evaluation of HiLAD framework.

\begin{itemize}
\item Uniform prior over the weights of the scoring function with tree-based ensembles to rank data instances is very effective for human-in-the-loop anomaly detection. The histogram distribution of the angles between score vectors from Isolation Forest (IFOR) and ${\mathbf w}_{unif}$ show that anomalies are aligned closer to ${\mathbf w}_{unif}$.

\item The diversified query selection strategy (\texttt{Select-Diverse}) based on compact description improves diversity over greedy query selection strategy with no loss in anomaly discovery rate.

\item The KL-Divergence based drift detection algorithm is very robust in terms of detecting and quantifying the amount of drift. In the case of limited memory settings with no concept drift, feedback tuned anomaly detectors generalize to unseen data. In the streaming setting with concept drift, our HiLAD-Stream algorithm is robust and competitive with the best possible configuration, namely, HiLAD-Batch.

\item HiLAD can discover higher anomalies with limited labeled feedback. With more labeled samples, it discovers more anomalies compared to baseline methods.

\end{itemize}
\section{Summary and Future Work}
\label{sec:conclusions}

This paper studied a human-in-the-loop learning framework using tree-based anomaly detector ensembles for discovering and interpreting anomalous data instances. We first explained the reason behind the empirical success of tree-based anomaly detector ensembles and called attention to an under-appreciated property that makes them uniquely suitable for label-efficient human-in-the-loop anomaly detection. We demonstrated the practical utility of this property by designing efficient learning algorithms to support this framework. We also showed that the tree-based ensembles can be used to compactly describe groups of anomalous instances to discover diverse anomalies and to improve interpretability. To handle streaming data settings, we developed a novel algorithm to detect the data drift and associated algorithms to take corrective actions. This algorithm is not only robust, but can also be employed broadly with any ensemble anomaly detector whose members can compute sample distributions such as tree-based and projection-based detectors. 

Our immediate future work includes deploying human-in-the-loop anomaly detection algorithms in real-world systems to measure their accuracy and usability (e.g., qualitative assessment of interpretability and explanations). Developing algorithms for interpretability and explainability of anomaly detection systems is a very important future direction.

\section*{Acknowledgments} This research was supported in part by the AgAID AI Institute for Agriculture Decision Support, supported by the National Science Foundation and United States Department of Agriculture - National Institute of Food and Agriculture award \#2021-67021-35344, and in part by Contract W911NF15-1-0461 with the US Defense Advanced Research Projects Agency (DARPA) Communicating with Computers Program and the Army Research Office.

\clearpage

\appendix
\section*{Appendix A. Parameter sensitivity for streaming algorithms}
\label{appendix:param-sensitivity}
We provide comprehensive experimental results to evaluate the performance of our active anomaly discovery algorithm for streaming data setting on the dataset described in section \ref{subsec:streaming_param}. Table
\ref{tab:param_sensitivity_electricty}, \ref{tab:param_sensitivity_covtype}, 
\ref{tab:param_sensitivity_weather} show the results for different experimental configurations. These results demonstrate that irrespective of the label feedback budget and data drift our proposed \texttt{KL-based} threshold selection approach remains stable.
\begin{table}[h]
\centering
\caption{Fraction of total anomalies discovered for \textit{Electricity} dataset. We varied the fraction of replaced trees for both fixed and KL-Divergence based approaches. KL-Divergence based approach with 95\% threshold shows stable results compared with other parameter settings (we marked this row \textbf{bold}). We also \underline{underlined} the best results obtained for different label feedback budget. Same observation applies to the other two datasets as well. Therefore, we recommend this setting (KL-Divergence with 95\% threshold) can be used as the default configuration for all datasets.}
\label{tab:param_sensitivity_electricty}
\begin{tabular}{crrrrrrrr}
\cline{3-9}
\multicolumn{1}{l}{} &        & \multicolumn{7}{c}{Feedback received}                \\ \hline
\begin{tabular}[c]{@{}c@{}}Tree replacement\\ technique\end{tabular}                      & Amount & 0    & 250   & 500   & 750   & 1000  & 1250  & 1500  \\ \hline
\multirow{8}{*}{Fixed}                                                                    & 5\%    & 0.07 & 12.49 & 20.98 & 28.20 & 31.52 & 32.93 & 33.32 \\ 
                     & 10\%   & 0.07 & 12.60 & 21.40 & 28.56 & 32.21 & 33.75 & 34.29 \\ 
                     & 15\%   & 0.07 & {\ul 12.60} & {\ul 21.66} & 29.15 & 33.01 & 34.77 & 35.49 \\ 
                     & 20\%   & 0.07 & 12.48 & 21.65 & {\ul 29.53} & 33.34 & 35.42 & 35.98 \\ 
                     & 25\%   & 0.07 & 12.27 & 21.10 & 29.01 & 33.18 & 35.32 & 35.96 \\ 
                     & 30\%   & 0.07 & 12.38 & 21.16 & 28.78 & 33.02 & 35.14 & 35.89 \\ 
                     & 50\%   & 0.07 & 12.10 & 20.55 & 28.50 & 33.81 & 36.76 & 38.28 \\ 
                     & 80\%   & 0.07 & 11.84 & 20.08 & 27.98 & {\ul 33.83} & {\ul 37.14} & {\ul 38.64} \\ \hline
\multirow{3}{*}{\begin{tabular}[c]{@{}c@{}}KL\\ threshold\\ (in percentile)\end{tabular}} & 92.5\% & 0.07 & 12.32 & 20.89 & 28.16 & 31.04 & 32.32 & 32.70 \\ 
                     & \textbf{95.0\%} & \textbf{0.07} & \textbf{12.39} & \textbf{21.33} & \textbf{28.88} & \textbf{32.83} & \textbf{34.58} & \textbf{35.33} \\ 
                     & 97.5\% & 0.07 & 12.41 & 21.60 & 29.43 & 32.85 & 34.56 & 35.14 \\ \hline
\end{tabular}
\end{table}

\begin{table}[]
\centering
\caption{Fraction of total anomalies discovered for \textit{Covtype} dataset. We varied the fraction of replaced trees for both fixed and KL-Divergence based approaches. KL-Divergence based approach with 95\% threshold shows stable results compared with other parameter settings (we marked this row \textbf{bold}). We also \underline{underlined} the best results obtained for different label feedback budget.}
\label{tab:param_sensitivity_covtype}
\begin{tabular}{crrrrrrrr}
\cline{3-9}
\multicolumn{1}{l}{}                 &        & \multicolumn{7}{c}{Feedback received}                \\ \hline
\multicolumn{1}{l}{Tree Replacement} &        & 0    & 500   & 1000  & 1500  & 2000  & 2500  & 3000  \\ \hline
\multirow{8}{*}{Fixed}               & 5\%    & 0.04 & 17.80 & 35.60 & 52.55 & 64.66 & 70.77 & 72.26 \\ 
                                     & 10\%   & 0.04 & 17.72 & 35.10 & 51.58 & 65.71 & 71.47 & 72.59 \\ 
                                     & 15\%   & 0.04 & 17.64 & 34.67 & 50.75 & 64.99 & 71.28 & 72.90 \\ 
                                     & 20\%   & 0.04 & 17.60 & 34.59 & 50.29 & 65.27 & 71.18 & 72.64 \\ 
                                     & 25\%   & 0.04 & 17.57 & 34.59 & 50.37 & 66.06 & 73.16 & 75.03 \\ 
                                     & 30\%   & 0.04 & 17.55 & 34.67 & 50.36 & 66.72 & 75.77 & 77.86 \\ 
                                     & 50\%   & 0.04 & 17.38 & 33.92 & 48.78 & 64.63 & 74.19 & 76.21 \\ 
                                     & 80\%   & 0.04 & 17.22 & 33.70 & 48.45 & 59.87 & 63.13 & 63.49 \\ \hline
\multirow{3}{*}{\begin{tabular}[c]{@{}c@{}}KL\\ threhsold\\ (in percentile)\end{tabular}} &
  92.5\% &
  0.04 &
  17.91 &
  35.84 &
  {\ul 53.04} &
  68.58 &
  78.03 &
  {\ul 80.58} \\ 
 & 
  \textbf{95.0\%} &
  \textbf{0.04} &
  {\ul \textbf{17.91}} &
  {\ul \textbf{35.85}} &
  \textbf{53.01} &
  {\ul \textbf{69.17}} &
  {\ul \textbf{78.42}} &
  \textbf{80.44} \\ 
                                     & 97.5\% & 0.04 & 17.88 & 35.80 & 52.97 & 68.76 & 77.85 & 79.76 \\ \hline
\end{tabular}
\end{table}

\begin{table}[h]
\centering
\caption{Fraction of total anomalies discovered for \textit{Weather} dataset. We varied the fraction of replaced trees for both fixed and KL-Divergence based approaches. KL-Divergence based approach with 95\% threshold shows stable results compared with other parameter settings (we marked this row \textbf{bold}). We also \underline{underlined} the best results obtained for different label feedback budget.}
\label{tab:param_sensitivity_weather}
\begin{tabular}{clrrrrr}
\cline{3-7}
                       &        & \multicolumn{5}{c}{Feedback received} \\ \hline
\begin{tabular}[c]{@{}c@{}}Tree Replacement\\ technique\end{tabular} & Amount          & 0             & 250            & 500            & 750            & 1000           \\ \hline
\multirow{8}{*}{Fixed} & 5\%    & 0.15  & 11.22 & 20.12 & 24.16 & 26.28 \\ 
                       & 10\%   & 0.15  & 11.33 & 20.43 & 25.44 & 27.91 \\ 
                       & 15\%   & 0.15  & 11.45 & 20.78 & 25.46 & 27.80 \\ 
                       & 20\%   & 0.15  & 11.60 & {\ul 21.23} & 26.05 & {\ul 29.13} \\ 
                       & 25\%   & 0.15  & 11.40 & 20.95 & {\ul 26.22} & 28.25 \\ 
                       & 30\%   & 0.15  & 11.75 & 20.82 & 25.44 & 27.90 \\ 
                       & 50\%   & 0.15  & 11.20 & 19.71 & 23.60 & 25.76 \\ 
                       & 80\%   & 0.15  & 11.11 & 17.39 & 20.14 & 20.63 \\ \hline
\multirow{3}{*}{KL}    & 92.5\% & 0.15  & 11.01 & 19.05 & 22.88 & 25.08 \\ 
                                                                     & \textbf{95.0\%} & \textbf{0.15} & \textbf{11.42} & \textbf{20.30} & \textbf{25.38} & \textbf{28.08} \\ 
                       & 97.5\% & 0.15  & {\ul 11.81} & 21.13 & 25.81 & 28.32 \\ \hline
\end{tabular}
\end{table}
\clearpage
\newpage
\section*{Appendix B. Comparison with streaming anomaly detectors}
\label{appendix:streaming-baselines-compare}
We performed a comparison with recent streaming based anomaly detection methods including RRCF \cite{guha:2016} and the classical baseline \cite{liu:08}. Table \ref{tab:streaming-comparison-electricty}, \ref{tab:streaming-comparison-covtype}, and \ref{tab:streaming-comparison-weather} demonstrate the better anomaly discovery performance of our HiLAD-Stream approach for all three datasets, namely, \textit{Electrcity}, \textit{Covtype}, and \textit{Weather} respectively.

\begin{table}[h]
\centering
\caption{Comparison of fraction of total anomalies discovered for the \textit{Electricity} dataset.}
\label{tab:streaming-comparison-electricty}
\begin{tabular}{crrrrrrr}
\cline{2-8}
             & \multicolumn{7}{c}{Feedback received}              \\ \hline
Algorithms   & 0    & 250   & 500   & 750   & 1000  & 1250 & 1500 \\ \hline
IFor         &  0.07   &    7.51   &    15.15   & 20.59      &  24.59     &  27.51    &   28.99   \\
RRCF         &  0.00    &  3.35     & 6.12      &  8.6     &    9.98   &    12.68  &    15.30  \\ \hline

\texttt{HiLAD-Stream}(KL-95\%) & \textbf{0.07} & \textbf{12.39} & \textbf{21.33} & \textbf{28.88} & \textbf{32.83} & \textbf{34.58} & \textbf{35.33}\\ \hline
\end{tabular}
\end{table}

\begin{table}[h]
\centering
\caption{Comparison of fraction of total anomalies discovered for the \textit{Covtype} dataset. }
\label{tab:streaming-comparison-covtype}
\begin{tabular}{crrrrrrr}
\cline{2-8}
             & \multicolumn{7}{c}{Feedback received}              \\ \hline
Algorithms   & 0    & 500   & 1000  & 1500  & 2000  & 2500 & 3000 \\ \hline
IFor         &  0.04 &  12.11 &  17.88 &  23.99 &  31.43 &  36.48 &  40.90     \\
RRCF         & 0.00  & 0.33 & 0.76      & 1.20      & 1.67      & 2.03      & 2.62      \\ \hline
\texttt{HiLAD-Stream}(KL-95\%) & \textbf{0.04} & \textbf{17.91} & \textbf{35.85} & \textbf{53.01} &  \textbf{69.17} & \textbf{78.42} & \textbf{80.44} \\ \hline
\end{tabular}
\end{table}

\begin{table}[h]
\centering
\caption{Comparison of fraction of total anomalies discovered for the \textit{Weather} dataset. }
\label{tab:streaming-comparison-weather}
\begin{tabular}{crrrrrrr}
\cline{2-8}
             & \multicolumn{7}{c}{Feedback received}              \\ \hline
Algorithms   & 0    & 250   & 500   & 750   & 1000  \\ \hline
IFor         & 0.15 &   1.33 &   4.10 &   9.10 &  14.31      \\
RRCF         & 0.15     & 2.4  & 6.10   & 8.54      & 11.59      \\ \hline
\texttt{HiLAD-Stream}(KL-95\%) & \textbf{0.15} & \textbf{11.42} & \textbf{20.30} & \textbf{25.38} & \textbf{28.08} &  &  \\ \hline
\end{tabular}
\end{table}

\clearpage

\vskip 0.2in
\bibliography{active_anomaly_ensembles}
\bibliographystyle{theapa}
\end{document}